\documentclass[aos,preprint,12pt]{imsart}

\setattribute{title}{case}{}
\setattribute{title}{size}{\Large\bfseries\mathversion{bold}}

\RequirePackage[OT1]{fontenc}
\RequirePackage{amsthm,amsmath,natbib}
\RequirePackage[colorlinks]{hyperref}
\usepackage{caption}
\usepackage{makecell}
\usepackage{array}
\usepackage{mathrsfs}
\usepackage{amssymb,bbm}
\usepackage{amsfonts}
\usepackage{amsmath,amssymb}
\usepackage{graphicx}
\usepackage{subfigure}
\usepackage{amscd}
\usepackage{color}
\usepackage{booktabs}
\usepackage{latexsym}
\usepackage{array,bm}
\usepackage{epstopdf}
\usepackage{enumerate}
\usepackage{soul}
\usepackage{ulem}
\usepackage[flushleft]{threeparttable}
\usepackage{algorithm}
\usepackage{algorithmic}
\usepackage{xr}
\usepackage{rotating}
\usepackage{multirow}
\usepackage{ragged2e}
\usepackage{longtable}
\usepackage{threeparttablex}
\usepackage{needspace}

\usepackage[capitalize,noabbrev]{cleveref}

\newcommand{\VaR}{\mathrm{VaR}}
\newcommand{\ES}{\mathrm{ES}}

\newcommand{\argmin}{\mathop{\mathrm{argmin}}}
\newcommand{\argmax}{\mathop{\mathrm{argmax}}}

\newcommand{\Enc}{\mathrm{Encoder}}
\newcommand{\Dec}{\mathrm{Decoder}}
\newcommand{\Ret}{\mathrm{Retriever}}
\newcommand{\InfEnc}{\mathrm{InfEncoder}}
\newcommand{\InfDec}{\mathrm{InfDecoder}}

\newcommand{\bbR}{\mathbb{R}}

\newcommand{\bb}{\bm{b}}
\newcommand{\bc}{\bm{c}}

\newcommand{\bg}{\bm{g}}
\newcommand{\bh}{\bm{h}}
\newcommand{\bk}{\bm{k}}
\newcommand{\bo}{\bm{o}}
\newcommand{\bq}{\bm{q}}
\newcommand{\br}{\bm{r}}

\newcommand{\bu}{\bm{u}}
\newcommand{\bv}{\bm{v}}

\newcommand{\bx}{\bm{x}}
\newcommand{\by}{\bm{y}}
\newcommand{\bz}{\bm{z}}

\newcommand{\bC}{\bm{C}}

\newcommand{\bE}{\bm{E}}
\newcommand{\bG}{\bm{G}}
\newcommand{\bH}{\bm{H}}

\newcommand{\bK}{\bm{K}}
\newcommand{\bM}{\bm{M}}
\newcommand{\bQ}{\bm{Q}}

\newcommand{\bS}{\bm{S}}
\newcommand{\bT}{\bm{T}}
\newcommand{\bU}{\bm{U}}
\newcommand{\bV}{\bm{V}}
\newcommand{\bW}{\bm{W}}
\newcommand{\bX}{\bm{X}}
\newcommand{\bY}{\bm{Y}}
\newcommand{\bZ}{\bm{Z}}
\newcommand{\vF}{\bm{F}}

\newcommand{\balpha}{\bm{\alpha}}
\newcommand{\bbeta}{\bm{\beta}}
\newcommand{\bgamma}{\bm{\gamma}}
\newcommand{\bdelta}{\bm{\delta}}

\newcommand{\bzeta}{\bm{\zeta}}
\newcommand{\btheta}{\bm{\theta}}

\newcommand{\bphi}{\bm{\phi}}
\newcommand{\bpsi}{\bm{\psi}}
\newcommand{\bomega}{\bm{\omega}}

\newcommand{\vf}{\bm{f}}

\startlocaldefs
\theoremstyle{plain}

\newcommand{\vecc}{\mathrm{vec}}

\usepackage{booktabs}

\def\boxit#1{\vbox{\hrule\hbox{\vrule\kern6pt
          \vbox{\kern6pt#1\kern6pt}\kern6pt\vrule}\hrule}}

\endlocaldefs

\renewcommand{\arraystretch}{1.5}

\begin{document}

\begin{frontmatter}

\title{ReSGA: A Large Tail Risk Model for Learning Value-at-Risk and Expected Shortfall}

{\thankstext{t2}{A user-oriented companion website for this paper, including stock-level tail risk forecasts, documentation, and links to the data and source code, is available at \url{https://tailrisk-resga.github.io}.}}

\thankstext{t3}{Correspondence to Zhoufan Zhu (Email: tylerzzf@xmu.edu.cn)}

\begin{aug}

  \author[a]{\fnms{Yichi} \snm{Zhang}
  \ead[label=e1]{yichizhang60@connect.hku.hk}},
  \author[a]{\fnms{Ke} \snm{Zhu}
  \ead[label=e2]{mazhuke@hku.hk}}
  \and
  \author[b,c]{\fnms{Zhoufan} \snm{Zhu}
  \ead[label=e3]{tylerzzf@xmu.edu.cn}} \thanksref{t3}

    \affiliation[a]{Department of Statistics and Actuarial Science, The University of Hong Kong, Hong Kong}

    \affiliation[b]{Wang Yanan Institute for Studies in Economics, Xiamen University, China}
    
    \affiliation[c]{Department of Finance, School of Economics, Xiamen University, China}


\end{aug}

\vspace{3mm}

\begin{abstract}

Learning Value-at-Risk (VaR) and Expected Shortfall (ES) is important for managing financial risks effectively. Existing approaches with limited parameters are vulnerable to model misspecification in the era of big data. To address this limitation, we propose a large tail risk model, the retrieval-enhanced self-grouping autoencoder (ReSGA), which is designed with millions of parameters to exploit the rich cross-sectional dependence and long-term temporal dynamics of assets using their characteristics. Applied to monthly US equity returns from 1926 to 2023 with 153 firm characteristics, ReSGA outperforms twelve econometric and machine learning competitors in terms of out-of-sample loss and statistical backtesting. In addition, its forecast advantages can translate into significant economic gains from long-short decile portfolios that are constructed by a new size-enhanced left-side momentum strategy. To clarify the role of complexity, we further conduct a systematic scaling analysis and demonstrate that improvements in joint VaR-ES forecasting are primarily driven by data complexity rather than model complexity. Finally, our analyses of group-importance and transfer-learning exhibit the interpretability and cross-market generalizability of ReSGA.

\end{abstract}


\end{frontmatter}

\newpage

\section{Introduction}

Value-at-Risk (VaR) and Expected Shortfall (ES) are central measures of financial tail risk in both academic research and regulatory practice. Considering the loss of a portfolio or investment, VaR characterizes only a quantile of the distribution of this loss, while ES summarizes expected losses beyond that quantile and therefore captures the severity of extreme downside outcomes. Moreover, ES is a coherent risk measure, satisfying properties like subadditivity, monotonicity, positive homogeneity, and translation invariance \citep{artzner1999}. In contrast, VaR is generally not subadditive, implying that portfolio risk measured by VaR may exceed the sum of individual risks. These advantages make ES more consistent with the diversification principles of portfolio theory and capture the tail risk in a more comprehensive manner, explaining its growing role in modern risk management \citep{BCBS2019FRTB}\footnote{For more studies on VaR and ES, one can refer to \cite{Acerbi2002on}, \cite{Rockafellar2002Conditional},  \cite{AcerbiSzekely2014}, \cite{du2017backtesting}, \cite{Li2023PELVE}, and \cite{Chronopoulos2024Forecasting}.}.

Although conceptually attractive, ES remains challenging to forecast in practice. A fundamental obstacle is that ES on its own is not elicitable, whereas VaR is \citep{gneiting2011}. In other words, there exists no loss function for which ES is the unique minimizer of expected loss. This lack of direct elicitability partly explains why the development of ES forecasting methods has lagged behind that of VaR. A major breakthrough is provided by \cite{fissler2016}, which introduce a class of consistent scoring rules, known as Fissler-Ziegel (FZ) loss functions, to utilize the joint elicitability of VaR and ES. This breakthrough leads to a growing literature on joint VaR-ES forecasting. For example, \cite{patton2019} employ FZ loss functions in semi-parametric settings based on the generalized autoregressive score (GAS) model and the generalized autoregressive conditional heteroskedasticity (GARCH) model; \cite{taylor2019} exploits the link between the asymmetric Laplace likelihood and the FZ loss for joint VaR-ES estimation; \cite{merlo2021} further extend the Laplace-based approach to multivariate regression settings. Until now, most existing work on joint VaR-ES forecasting remains confined to parametric or semi-parametric models. In such frameworks, the number of parameters is kept modest to ensure interpretability, and each asset is often modeled separately with its own set of parameters. While tractable, their performance depends critically on correct model specification, where the true data-generating process for tail risks may be far more complex than a GARCH or GAS model can capture. 

Fortunately, the era of big data brings new opportunities for tail risk modeling. Risk forecasters now have access to decades of financial data, covering tens of thousands of assets, along with hundreds of firm characteristics. This abundance of information naturally raises a key question: Can we develop a large tail risk model, analogous to large language models (e.g., ChatGPT, Gemini, and Claude) in natural language processing, that learns from rich and diverse data to predict VaR and ES more accurately? In other words, \textit{instead of relying on asset-specific (semi-)parametric models, could a single high-capacity universal model trained on massive data serve as a general engine for joint VaR-ES forecasting?} 

To shed light on large tail risk models, we first develop a unified learning framework for joint VaR-ES forecasting. This framework assumes a universal function indexed by unknown parameters, mapping from predictors (e.g., firm characteristics) to two latent but unconstrained risk scores. Then, it transforms these two risk scores into valid VaR and ES forecasts through a fixed one-to-one mapping. This data-driven framework enables linear regression models, machine learning methods, and deep learning architectures to be treated in the same manner, with parameters estimated by minimizing the FZ loss. 

Under this framework, we propose a new large tail risk model, the \textit{retrieval-enhanced self-grouping autoencoder} (ReSGA), designed to fully exploit complex financial information for joint VaR-ES forecasting. ReSGA first learns latent group structures across assets, allowing tail-risk information to be shared among assets with similar financial behavior. It then incorporates a retrieval mechanism that enables each asset to draw on relevant long-term historical ``experiences'' from both its own past and other assets within the same group, thereby enriching usable temporal information. By leveraging cross-sectional dependence and long-term temporal dynamics in a flexible data-driven manner, ReSGA exploits spatial-temporal information more effectively than existing approaches. 

Empirically, we study the monthly VaR and ES of United States (US) equity returns using a comprehensive dataset provided by \cite{jensen2023}, covering more than 40,000 stocks from January 1926 to December 2023. Each stock-month observation is associated with 153 firm characteristics that are widely used in the asset pricing literature. We evaluate the out-of-sample VaR and ES forecasting performance of ReSGA and its 12 competing models through the average loss. From this analysis, we find that ReSGA consistently attains the lowest out-of-sample average loss across all stock universes considered. Besides the loss analysis, we further assess the statistical performance of all models via several statistical hypothesis tests, including the Diebold--Mariano (DM; \citealp{diebold1995}) test, the model confidence set (MCS; \citealp{Hansen2011model}) test, the conditional coverage (CC; \citealp{Christoffersen1998EvaluatingIF}) test, and the auxiliary expected shortfall regression (AESR; \citealp{bayer2022}) test. The out-of-sample testing results also provide consistent evidence in favor of ReSGA. 

Beyond statistical assessments, we are interested in whether the tail risk forecasts generated by ReSGA can lead to economically meaningful gains. Following \cite{AtilganYigit2020}, we construct a left-side momentum strategy by sorting stocks into deciles based on the predicted VaR or ES. The resulting VaR- and ES-sorted portfolios exhibit sharp nonlinearities concentrated in the extreme deciles: Stocks with the highest predicted tail risk (i.e., the lowest VaR or ES) deliver substantially lower returns and markedly worse downside risk, while a long-short strategy that buys the lowest-risk decile and sells the highest-risk decile, yields economically sizable Sharpe ratios. These patterns indicate that ReSGA’s tail risk forecasts capture cross-sectional variation in downside risk, which is also priced in the cross-section. Building on this insight, we propose a new size-enhanced left-side momentum signal that combines predicted ES with firm size. This signal is motivated by the ``too big to fail'' phenomenon in the US market: Tail risk in large firms is more likely to reflect compensated systematic risk, whereas similar tail risk in small firms often stems from idiosyncratic fragility. 
Based on this signal, we propose long-short portfolios for each considered model and find that all proposed non-econometric portfolios exhibit significant alphas relative to the Fama--French five-factor model. Hence, it indicates that our size-enhanced left-side momentum signal could broaden the mean-variance frontier. More importantly, our portfolio performances confirm that ReSGA delivers the best economic performance among all competing models, with the highest Sharpe ratio for the long-short decile portfolio and a clear monotonic return pattern across deciles. All of the above findings show that the forecasting advantage from ReSGA is not only statistically relevant but also economically meaningful.

Taken together, our statistical and economic findings above highlight that models can deliver better tail risk forecasts by exploiting richer information in the input. We view the informativeness of model input as one aspect of data complexity. This phenomenon connects to a broader debate on the virtue of complexity. In artificial intelligence, a growing body of evidence shows that the out-of-sample predictive performance of models often improves systematically with increases in either data complexity (with respect to \textit{in-sample data availability}) or model complexity (with respect to \textit{model parameter size}); see, for example, \cite{kaplan2020scalinglawsneurallanguage}. This empirical regularity, known as the scaling law, underpins the success of large language models in the last five years. In finance, there has been an increasing focus on the investigation of model complexity, while the examination of data complexity remains very rare. \cite{Bryan2024virture} provide early theoretical and empirical support for the virtue of model complexity in asset pricing, demonstrating that simple models with few parameters can severely understate return predictability relative to more complex alternatives. See more empirical evidence for this view in \cite{APTDidisheim2024} and \cite{Artificial2025Kelly}. However, at the same time, other studies also question whether such gains from model complexity are economically meaningful or robust in asset pricing applications \citep{Berk2023CommentVirtueComplexity,Seemingly2025Stefan,Buncic2025Complexity,CarteaJinShi2025Complexity}. In terms of risk forecasting, evidence on the role of model complexity is far more limited; \cite{li2025} show that machine learning models outperform linear models in realized volatility forecasting by capturing nonlinear temporal dynamics. However, to date, the role of model complexity and data complexity in tail risk forecasting remains largely unexplored.

To fill the gap, we further investigate the scaling performance of ReSGA and its competitors by varying model parameter size or in-sample data size to assess the role of model complexity or data complexity, respectively. First, our experiment results show that, holding the in-sample data fixed, simply increasing parameter size alone does not yield consistent performance improvements: Larger models do not reliably outperform smaller ones, and the best results in terms of out-of-sample average loss and portfolio Sharpe ratio typically occur at intermediate scales (around or below the in-sample sample size). In particular, models that exploit spatial–temporal inputs, such as ReSGA, benefit more from additional parameters and consistently outperform simpler architectures once sufficient parameters are available. Second, our complementary experiments vary the available in-sample data size while keeping the model architectures fixed. The corresponding results demonstrate that, in terms of out-of-sample average loss and portfolio Sharpe ratio, the performance of each model exhibits a clear improvement trend with the size of the in-sample data used. Together, the aforementioned scaling results indicate that our out-of-sample performance gains are primarily driven by richer data rather than by parameter growth alone. Hence, they provide limited support for a general virtue of model complexity, instead emphasizing the central role of data complexity in tail risk forecasting.

Lastly, we illustrate the usefulness of ReSGA through a group-importance analysis and a transfer-learning analysis. In the group-importance analysis, we explore economic model interpretability by investigating which types of firm characteristics drive tail risk predictability in the ReSGA model. It turns out that the predictive power for VaR and ES is concentrated in a small set of economic themes (or groups): Value, Low Risk, Momentum, Quality, and Short-Term Reversal. Additionally, the importance of these groups varies systematically across firm size and over time, reflecting shifts in the underlying economic sources of tail risk. In the transfer-learning analysis, we deploy the ReSGA model trained only on US equity data, without any re-estimation or fine-tuning, to five major international equity markets: China, Japan, the United Kingdom (UK), Australia and Canada. Despite substantial cross-market heterogeneity, ReSGA consistently presents reliable out-of-sample VaR-ES forecasting performance in every market, indicating that its predictive advantage is not US-specific and remains robust under market shifts. Hence, ReSGA, as a universal large tail risk model, has great generalizability in learning VaR and ES. At the portfolio level, however, return patterns differ substantially across countries. In China and Japan, higher size-enhanced left-side momentum is associated with higher returns, while in the UK, Australia, and Canada, large return dispersion concentrates in the extreme deciles. These findings suggest that tail-risk pricing varies across markets, likely reflecting differences in market structure, investor behavior, and risk premia.

The remaining paper proceeds as follows. \cref{sec:methodology} introduces a unified VaR-ES learning framework and the ReSGA model. \cref{sec:empirical} presents the empirical results for the US equity market, while \cref{sec:transfer} examines the transferability of the ReSGA model trained on US data to other markets. \cref{sec:conclusion} concludes. Technical details are provided in Appendix \ref{sec:resga_arch} and other appendices in the supplementary materials.

\section{Methodology}\label{sec:methodology}

Let $r_{i,t}$ denote the return of asset $i$ at time $t$, where $t = 1, \ldots, T$ and $i = 1, \ldots, N_t$, with $N_t$ representing the number of available assets at time $t$. The variables of interest in this paper are the conditional $\tau$-th VaR and ES of $r_{i,t}$, defined respectively as
\begin{align*}
    \VaR_{i,t} =  F_{i,t}^{-1}(\tau) \,\,\, \text{and} \,\,\, \ES_{i,t} = E \left( r_{i,t} \mid r_{i,t} \le \VaR_{i,t}, \mathcal{F}_{t-1} \right),
\end{align*}
where $\tau\in(0,1)$ is the quantile level, $F_{i,t}(\cdot)$ is the conditional distribution of $r_{i,t}$ given $\mathcal{F}_{t-1}$, and $\mathcal{F}_{t-1}$ is the information set containing all available information up to time $t-1$.

In the following subsections, \cref{sec:loss} introduces a unified learning framework for $\VaR_{i,t}$ and $\ES_{i,t}$. \cref{sec:resga} presents our new ReSGA model. \cref{sec:competitor} describes a set of representative benchmark tail risk models that are used for comparison in our empirical study. 

\subsection{A Unified Learning Framework}\label{sec:loss}

Following the seminal work of \cite{koenker1978regression}, a common way to learn VaR is based on the check loss function. However, as shown in \cite{gneiting2011}, ES alone is not elicitable, meaning that it cannot be the unique minimizer of an expected loss function. \cite{fissler2016} solve this issue by showing that VaR and ES are jointly elicitable: $(\VaR, \ES)$ is the unique minimizer of the general FZ loss function. Specifically, to jointly learn the VaR and ES of a univariate random variable $Y$, \cite{fissler2016} define the general FZ loss function (also termed as the scoring rule) as follows:
\begin{align}\label{general_FZ}
    \begin{split}
        \ell(Y, (v, e))
        &= \big(\bm{1}\{ Y \le v\} - \tau\big)g_1(v) - \bm{1}\{Y \le v\} g_1(Y) 
        \\
        &\quad + g_2(e)\left[e-v+\frac{1}{\tau}\bm{1}\{ Y \le v\} (v - Y)\right] -  \tilde{g}_2(e),
    \end{split}
\end{align}
where $\tau\in(0, 1)$ is the quantile level, $\bm{1}\{\cdot\}$ is the indicator function, and $g_1,g_2:\mathbb{R} \to \mathbb{R}$ are two pre-determined functions, with $g_2$ being strictly increasing and convex, and $\tilde{g}_2$ being its primitive. In particular, when $g_1(x) = 0$ and $g_2(x) = -1/x$, the general FZ loss function in (\ref{general_FZ}) becomes the commonly used degree-0 FZ loss function:
\begin{align}\label{eq:FZ0}
\ell_{\mathrm{FZ0}}(Y, (v, e))
= -\frac{1}{\tau e}\,\bm{1}\{Y \le v\}(v - Y)
+ \frac{v}{e} + \log(-e) - 1,
\end{align}
where $(v, e)\in \Gamma$ for an admissible region $\Gamma=\{(v, e): e < v < 0\}$. Using the degree-0 FZ loss function, a joint learning of VaR and ES can be achieved by noting 
\begin{align}\label{es_optim}
    (\VaR \mbox{ of }Y,\, \ES \mbox{ of } Y)=\argmin_{(v,e)\in\Gamma} \, \mathbb{E} \left[\ell_{\mathrm{FZ0}}(Y, (v, e)) \right].
\end{align}

As shown in \eqref{es_optim}, the learning of $\VaR_{i,t}$ and $\ES_{i,t}$ under $\ell_{\mathrm{FZ0}}$ needs to meet a key admissible condition: $\ES_{i,t} < \VaR_{i,t} < 0$. To achieve this goal, we propose a new unified learning framework for $\VaR_{i,t}$ and $\ES_{i,t}$. Specifically, let $\by_{i,t} = (y_{i,t,1}, y_{i,t,2})'\in\mathbb{R}^2$ represent a vector of two latent risk scores $y_{i,t,1}$ and $y_{i,t,2}$. Then, we define a Softplus-based one-to-one mapping $p$: $\by_{i,t} \rightarrow (\VaR_{i,t}, \ES_{i,t})$,
satisfying
\begin{align}\label{activate}
    \begin{split}
        \VaR_{i, t} & = -\mathrm{Softplus}(y_{i,t,1}), \\
        \ES_{i, t} & = - \big[\mathrm{Softplus}(y_{i,t,1}) + \mathrm{Softplus}(y_{i,t,2})\big],
    \end{split}
\end{align}
where $\mathrm{Softplus}(\cdot)$ is a function \citep{Softplus} defined as
\begin{align*}
    \mathrm{Softplus}(x) = \log (1 + e^x) \in \mathbb{R}^{+}, \,\,\, \text{for} \,\,\, x \in \mathbb{R}.
\end{align*}
Since both $\mathrm{Softplus}(y_{i,t,1})$ and $\mathrm{Softplus}(y_{i,t,2})$ are strictly positive, the mapping in \eqref{activate} guarantees that $\ES_{i,t} < \VaR_{i,t} < 0$, regardless of the values of $y_{i,t,1}$ and $y_{i,t,2}$.
Consequently, we are able to learn $\VaR_{i,t}$ and $\ES_{i,t}$ through a model for $\by_{i,t}$, the form of which can be specified without imposing restrictions on $y_{i,t,1}$ and $y_{i,t,2}$. 

Following the above idea, our learning framework only needs to specify a model for $\by_{i,t}$ indexed by unknown parameters $\btheta$. Therefore, unless otherwise stated, all tail risk models in this paper are designed for $\by_{i,t}$. By writing  $\by_{i,t} \equiv \by_{i,t}(\btheta)$, we have $(\VaR_{i,t}, \ES_{i,t})=p(\by_{i,t}(\btheta))$ according to (\ref{activate}), so we can make use of result (\ref{es_optim}) to estimate $\btheta$ by minimizing the empirical degree-0 FZ loss:
\begin{align}\label{eq:train_objective}
    \widehat{\btheta}
    & = \argmin_{\btheta} \sum_{t=1}^{T}
    \sum_{i=1}^{N_t}
    \ell_{\mathrm{FZ0}}\!\big(r_{i,t}, \big( \VaR_{i,t}, \ES_{i,t} \big) \big) = \argmin_{\btheta}
    \sum_{t=1}^{T} \sum_{i=1}^{N_t}
    \ell_{\mathrm{FZ0}}\!\big(r_{i,t}, p(\by_{i,t}(\btheta))\big).
\end{align}
In practice, due to the massive data volume, we use the adaptive moment estimation (Adam) algorithm in \cite{Kingma2015AdamAM} to solve optimization problem \eqref{eq:train_objective}. Given $\widehat{\btheta}$, $\VaR_{i,t}$ and $\ES_{i,t}$ are then learned by $\widehat{\VaR}_{i,t}$ and $\widehat{\ES}_{i,t}$, respectively, where 
$(\widehat{\VaR}_{i,t}, \widehat{\ES}_{i,t})=p(\by_{i,t}(\widehat{\btheta}))$.

\subsection{The ReSGA Model}\label{sec:resga}

Let $\bY_{t} \in \mathbb{R}^{N_t \times 2}$ denote the matrix of cross-sectional risk scores at time $t$, where its $i$-th row is $\by_{i,t}'$. Under the learning framework in Section \ref{sec:loss}, we design a new ReSGA model for $\bY_{t}$, which aims to capture the nonlinear relationships between asset characteristics and risk scores, while simultaneously accounting for spatial and temporal dependencies of assets. To facilitate the construction of ReSGA, we let $\bX_{i,t-1} \in \mathbb{R}^{S \times P}$ be the feature matrix of asset $i$ up to time $t-1$, with its $s$-th row $\bx_{i,t-1-S+s}'$, where $\bx_{i,t} = (x_{i,t,1}, \ldots, x_{i,t,P})'\in\mathbb{R}^{P}$ contains $P$ characteristics of this asset at time $t$. Here, $P$ is the number of characteristics and $S$ is the number of time lags. Collecting all individual feature matrices, we obtain the tensor of asset characteristics $\mathcal{X}_{t-1} = [\bX_{1,t-1}, \dots, \bX_{N_t,t-1}] \in \mathbb{R}^{N_t \times S \times P}$.

Using $\mathcal{X}_{t-1}$ as the input, ReSGA, parameterized by $\btheta=(\bphi',\bpsi')'$, learns $\bY_{t}$ through three core modules, \textit{encoder}, \textit{retriever}, and \textit{decoder}, which are defined as follows:
\begin{align}\label{resga_structure}
\begin{split}
    (\mathcal{H}_t, \mathbb{M}_t) &= \Enc(\mathcal{X}_{t-1}; \bphi), 
    \\
    \bZ_t &= \Ret(\mathcal{H}_{t}; \mathbb{M}_t),
    \\
    \bY_t &= \Dec \left(\mathcal{H}_t, \bZ_t; \mathbb{M}_t, \bpsi \right),
\end{split}
\end{align}
where $\Enc(\cdot; \bphi)$, $\Ret(\cdot; \mathbb{M}_t)$, and $\Dec(\cdot, \cdot; \mathbb{M}_t, \bpsi)$ denote the functional forms of encoder, retriever, and decoder, respectively (see their detailed architectures in Sections \ref{sec:resga_enc}--\ref{sec:resga_dec}).  
 Generally speaking, 
\begin{itemize}
    \item $\Enc(\cdot; \bphi)$, parameterized by $\bphi$, extracts a temporal feature $\mathcal{H}_t\in \mathbb{R}^{N_t \times S \times D}$ for all assets across lags in an autoregressive manner, along with a set $\mathbb{M}_t$ that captures latent group structures among assets according to their similarities in characteristics, where the dimension $D$ serves as a user-specific hyperparameter controlling model complexity;
    \item $\Ret(\cdot;\mathbb{M}_t)$ leverages the learned group structure $\mathbb{M}_t$ to retrieve long-term historical information not only from each asset’s own temporal history but also from other assets within the same group, yielding the retrieval feature $\bZ_t \in \mathbb{R}^{N_t \times D}$;
    \item $\Dec(\cdot, \cdot; \mathbb{M}_t, \bpsi)$, parameterized by $\bpsi$, hierarchically aggregates the group-, retrieval-, and asset-level information to generate the risk score $\bY_t$. 
\end{itemize}

It is worth noting that $\mathbb{M}_t$ depends on the specific choice of $\mathcal{X}_{t-1}$ and loss function. In this study, under the degree-0 FZ loss in (\ref{eq:train_objective}), $\mathbb{M}_t$ reveals the dynamic topological relationships among assets for tail risk analysis, guided by the similarities in asset characteristics. When ReSGA is applied to other studies, different inputs and loss functions can lead to different interpretations for $\mathbb{M}_t$.

\subsection{Competing Models}\label{sec:competitor}

To assess the performance of ReSGA, we consider its 12 competing models. These competitors are classified into four categories: \textit{Point-wise}, \textit{Temporal}, \textit{Spatial–temporal} and \textit{Econometric} models, depending on whether they exploit temporal or cross-sectional information and whether they employ firm characteristics. Models in the first three categories are based on our learning framework in Section \ref{sec:loss}, while models in the last category learn VaR and ES using existing econometric approaches. For more details about these 12 competing models, one can refer to Appendix \ref{sec:models} in the supplementary materials. 

\paragraph{Point-wise Models}

The point-wise models aim to capture the relationship between asset characteristics and tail risks, but ignore both temporal and cross-sectional dependencies. They directly map the most recent asset characteristics to risk scores as follows:
\begin{align}\label{point-wise}
    \by_{i,t} = f(\bx_{i,t-1}; \btheta),
\end{align}
where $\bx_{i,t-1} \in\mathbb{R}^{P}$ is the vector of $P$ characteristics for asset $i$ at time $t-1$, and $f(\cdot;\btheta)$ is a parametric function indexed by $\btheta$. Particularly, we focus on two models in (\ref{point-wise}): (i) the classical linear regression model, labeled as ``Linear''; (ii) a three-hidden-layer neural network model proposed by \cite{gu2020empirical}, labeled as ``NN''. 

\paragraph{Temporal Models}

Unlike point-wise models, the temporal models explicitly account for the dynamic evolution of asset characteristics over time by assuming
\begin{align}\label{temporal_model}
    \by_{i,t} = g(\bX_{i,t-1}; \btheta),
\end{align}
where $\bX_{i, t - 1}\in\mathbb{R}^{S\times P}$ is the feature matrix of asset $i$ at time $t-1$, and $g(\cdot;\btheta)$ is a parametric function indexed by $\btheta$. Note that the hyperparameter $S$ specifies the look-back time period of characteristics. When $S=1$, $\bX_{i,t-1}'$ reduces to $\bx_{i,t-1}$, which is the vector of the most recent characteristics (i.e., the input to point-wise models in (\ref{point-wise})). Compared with point-wise models, the temporal models in (\ref{temporal_model}) exploit not only the predictive power of the latest characteristics, but also the historical evolution of characteristics, allowing them to capture dynamic patterns that may enhance predictive performance.

The temporal models require specific designs to effectively capture temporal dynamics. In this paper, we employ several representative temporal deep learning models in \eqref{temporal_model}, including:
\begin{enumerate}
    \item[(i)] two extended models: a lag-augmented neural network model that flattens $\bX_{i,t-1}$ into a vector as input (labeled as ``LANN''), and a decomposition-based neural network model proposed by \cite{zeng2023transformers} that separately builds trend and seasonal components (labeled as ``DLinear'');
    \item[(ii)] two recurrent neural network models: long short-term memory network model from \cite{hochreiter1997long} (labeled as ``LSTM'') and gated recurrent unit network model from \cite{GRU} (labeled as ``GRU'');
    \item[(iii)] three transformer-type neural network models: the ``Informer'' model proposed by \cite{zhou2021informer}, which leverages an attention-based encoder and decoder to capture temporal dependencies, along with its encoder-only and decoder-only variants (labeled as ``EInformer'' and ``DInformer'', respectively).
\end{enumerate}

\paragraph{Spatial-temporal Models}

For the spatial-temporal models, they simultaneously capture both the temporal evolution of asset characteristics and the cross-sectional interactions among assets using advanced deep learning architectures. Specifically, they are defined as
\begin{align}\label{spatial_temporal_model}
    \bY_{t} = m(\mathcal{X}_{t-1}; \btheta),
\end{align}
where $\mathcal{X}_{t-1} \in \mathbb{R}^{N_t \times S \times P}$ is the tensor of asset characteristics, and $m(\cdot;\btheta)$ is a parametric function indexed by $\btheta$. Compared to the temporal models, which handle assets separately, the spatial–temporal models allow each asset’s risk scores to depend not only on its own historical information but also on that of other assets. By explicitly modeling these inter-asset dependencies, the spatial–temporal models are capable of capturing common shocks, contagion effects, and network spillovers, which are particularly relevant in financial markets. This spatial information sharing is expected to lead to improved robustness and predictive accuracy, especially during periods of heightened market co-movement or systemic stress.

Given the rich information contained in $\mathcal{X}_{t-1}$, the spatial–temporal models must be carefully designed to efficiently exploit both temporal and cross-sectional dependencies. Besides the proposed ReSGA model, we consider another spatial–temporal model in \eqref{spatial_temporal_model}: the self-grouping autoencoder model from \cite{GRAND} (labeled as ``SGA''). In short, the SGA model can be viewed as a simplified version of ReSGA. Both models share a common encoder–decoder architecture that learns time-varying group structures across assets. However, SGA removes all retrieval-related components and focuses solely on contemporaneous spatial dependencies. As a result, it could perform well when the temporal length $S$ is relatively short. In contrast, ReSGA extends this framework by incorporating a retrieval mechanism to capture long-term spatial–temporal interactions, making it suitable for longer sequences and more complex systemic risk patterns. 

\paragraph{Econometric Models}

In the econometrics literature, \cite{patton2019} apply two widely used semi-parametric models, GAS and GARCH, to study the dynamics of VaR and ES. These models do not rely on machine learning or deep learning techniques to utilize the information of asset characteristics, yet they need to impose the constraint of $\ES_{i,t}<\VaR_{i,t}<0$ when implementing optimization in model estimation. Including them allows us to assess whether more complex models deliver meaningful improvements over well-established econometric approaches. Notably, unlike the aforementioned non-econometric models, which employ a unified learning framework with a single parameter set shared across assets, the GAS and GARCH models are estimated separately for each asset to learn VaR and ES directly.

\vspace{2mm}

\cref{tab:model_characteristics} summarizes the features of all considered tail risk models, such as nonlinearity from deep learning, temporal dynamics from lagged characteristics, and spatial information sharing within cross-sectional input. In addition, it also distinguishes models by whether they support long-history inputs (i.e., capturing long memory effect). By clarifying the incremental modeling capabilities across point-wise, temporal,  spatial–temporal, and econometric models, this table clearly illustrates how ReSGA integrates all these features within a single model to exploit available information efficiently. 

\begin{table}[!ht]
\centering
\caption{Summary of model features.}
\label{tab:model_characteristics}
\begin{threeparttable}
\setlength{\tabcolsep}{6.2mm}
\begin{tabular}{lccccc}
\toprule
  Model & Universal  & Nonlinear  & Temporal  & Spatial  & Long Memory
\\
\midrule
Linear              & \checkmark & --          & --          & --          & --          \\
NN                  & \checkmark & \checkmark  & --          & --          & --          \\
LANN                & \checkmark & \checkmark  & \checkmark  & --          & \checkmark  \\
DLinear             & \checkmark & \checkmark  & \checkmark  & --          & \checkmark  \\
LSTM                & \checkmark & \checkmark  & \checkmark  & --          & --          \\
GRU                 & \checkmark & \checkmark  & \checkmark  & --          & --          \\
Informer            & \checkmark & \checkmark  & \checkmark  & --          & \checkmark  \\
EInformer           & \checkmark & \checkmark  & \checkmark  & --          & \checkmark  \\
DInformer           & \checkmark & \checkmark  & \checkmark  & --          & \checkmark  \\
SGA                 & \checkmark & \checkmark  & \checkmark  & \checkmark  & --          \\
ReSGA               & \checkmark & \checkmark  & \checkmark  & \checkmark  & \checkmark  \\
GARCH               & --         & --          & \checkmark  & --          & --          \\
GAS                 & --         & --          & \checkmark  & --          & --          \\
\bottomrule
\end{tabular}
\footnotesize
\textit{Notes.}  
``Universal'' indicates that the model is applied to all assets. ``Nonlinear'' showcases that the model is nonlinear. ``Temporal'' means that the model processes sequential information from multiple time lags. ``Spatial'' signifies that the model makes use of information shared across assets. ``Long Memory'' expresses that the model is able to utilize long-term historical information.
\end{threeparttable}
\end{table}

\section{Empirical Study of US Equity}\label{sec:empirical}

\subsection{Data}\label{sec:data}

This section analyzes a comprehensive US equity dataset compiled by \cite{jensen2023}. This dataset covers more than 40,000 stocks, providing a broad and representative cross-section of the US equity market for large-scale risk forecasting. Meanwhile, it spans the period from January 1926 to December 2023 at a monthly frequency, offering nearly a century of observations across different market regimes, including economic expansions and major crisis episodes. In this dataset, each stock in every month is associated with 153 firm characteristics\footnote{See \url{https://jkpfactors.s3.amazonaws.com/documents/Documentation.pdf} for more details.}. As in \cite{gu2020empirical}, we 
rank-normalize all characteristics cross-sectionally to mitigate the impact of outliers, and then impute missing characteristic values using contemporaneous cross-sectional medians. After this preprocessing treatment, we let $r_{i,t}$ denote the excess return of stock $i$ in month $t$, and construct the characteristics tensor $\mathcal{X}_{t-1}$ from all 153 characteristics. 

Throughout the analysis, we apply the ReSGA model to learn $\VaR_{i,t}$ and $\ES_{i,t}$ at the quantile level $\tau=0.05$, with a comparison to the 12 competing models outlined in Section \ref{sec:competitor}. To evaluate out-of-sample forecasting performance for all models, we follow \cite{gu2020empirical} to adopt an expanding-window process, with the last ten years of the sample (2014--2023) reserved for out-of-sample evaluation. Specifically, each model is trained on data from 1926 to 1995, validated on data from 1996 to 2013, and tested on data from 2014; this process then shifts forward by one year (trained on 1926--1996, validated on 1997--2014, and tested on 2015), and it is repeated until reaching the end of the out-of-sample period\footnote{As an exception, two econometric models, GAS and GARCH, have no need for validation, so they are estimated using all training and validation data.}. For the detailed implementation of the above training, validation, and testing process, one can refer to \cref{sec:implementation}. 

\subsection{Statistical Performance Evaluation}

We first evaluate the out-of-sample forecasting performance of ReSGA and its 12 competitors from a statistical perspective. As in \cite{patton2019}, we consider the following out-of-sample average loss as a natural assessment criterion:
\begin{align}\label{oos_loss}
    \ell_{\text{oos}} = \frac{1}{\sum_{t \in \text{oos}} |\mathbb{S}_t|}\sum_{t \in \text{oos}} \sum_{i \in \mathbb{S}_t} \ell_{\mathrm{FZ0}}\left(r_{i,t}, \left(\widehat{\VaR}_{i,t}, \widehat{\ES}_{i,t}\right) \right),
\end{align}
where ``oos'' denotes the out-of-sample period, $\mathbb{S}_t$ denotes a group of stocks at time $t$, with $|\mathbb{S}_t|$ being its cardinality, $\ell_{\mathrm{FZ0}}$ is the degree-0 FZ loss function defined in \eqref{eq:FZ0}, and $\widehat{\VaR}_{i,t}$ and $\widehat{\ES}_{i,t}$ are forecasts of $\VaR_{i,t}$ and $\ES_{i,t}$ from each learned model. 

\begin{table}[!ht]
\centering
\caption{Out-of-sample average loss $\ell_{\text{oos}}$ across different models and stock groups.}
\label{tab:oos_loss_by_size}
\begin{threeparttable}
\setlength{\tabcolsep}{5.8mm}
\begin{tabular}{lcccccc}
\toprule
 & All& Mega & Large & Small & Micro & Nano \\
\midrule

\multicolumn{7}{l}{\textit{Point-wise models}} \\[2pt]
\quad Linear     & $3.3442$ & $2.8431$ & $3.0908$ & $3.3305$ & $3.4642$ & $3.7616$ \\
\quad NN         & $3.3322$ & $2.8770$ & $3.0958$ & $3.3055$ & $3.4380$ & $3.7561$ \\[3pt]

\midrule
\multicolumn{7}{l}{\textit{Temporal models}} \\[2pt]
\quad LANN    & $3.3440$ & $2.8797$ & $3.0944$ & $3.3055$ & $3.4679$ & $3.8151$ \\
\quad DLinear    & $3.3166$ & $2.8385$ & $3.0703$ & $3.2919$ & $3.4508$ & $3.7404$ \\
\quad LSTM       & $3.3063$ & $2.8368$ & $3.0670$ & $3.2838$ & $3.4368$ & $3.7184$ \\
\quad GRU        & $3.3022$ & $2.8326$ & $3.0601$ & $3.2807$ & $3.4357$ & $3.7082$ \\
\quad Informer   & $3.2985$ & $2.8217$ & $3.0532$ & $3.2779$ & $3.4334$ & $3.7098$ \\
\quad EInformer  & $3.3276$ & $2.8387$ & $3.0732$ & $3.3148$ & $3.4648$ & $3.7395$ \\
\quad DInformer  & $3.3134$ & $2.8258$ & $3.0578$ & $3.2925$ & $3.4535$ & $3.7346$ \\[3pt]

\midrule
\multicolumn{7}{l}{\textit{Spatial-temporal models}} \\[2pt]
\quad SGA        & $3.2868$ & $2.8215$ & $3.0527$ & $3.2647$ & $3.4118$ & $3.7017$ \\
\quad ReSGA      & $\pmb{3.2793}$ & $\pmb{2.8081}$ & $\pmb{3.0417}$ & $\pmb{3.2635}$ & $\pmb{3.4036}$ & $\pmb{3.6932}$ \\[3pt]

\midrule
\multicolumn{7}{l}{\textit{Econometric models}} \\[2pt]
\quad GAS       & $4.3461$ & $3.0214$ & $3.1899$ & $3.7713$ & $5.5837$ & $5.0057$ \\
\quad GARCH     & $4.7122$ & $2.9189$ & $3.2377$ & $3.8226$ & $6.6247$ & $5.0849$ \\ [3pt]

\bottomrule
\end{tabular}

\vspace{3pt}
\footnotesize

\textit{Notes.}
This table reports out-of-sample average loss $\ell_{\text{oos}}$ for six stock groups, with $\mathbb{S}_t$ formed by all stocks and five size-based subgroups (mega, large, small, micro, and nano) of stocks. These subgroups are defined by the percentiles of firms’ market capitalization among all stocks at the end of each month: Mega stocks are above the 80th percentile, large stocks are between the 50th and 80th percentiles, small stocks are between the 20th and 50th percentiles, micro stocks are between the 1st and 20th percentiles, and nano stocks fall below the 1st percentile. The lowest value for each column is highlighted in boldface. 
\end{threeparttable}
\end{table}

\cref{tab:oos_loss_by_size} reports the results of $\ell_{\text{oos}}$ across models and different stock groups. From this table, we can draw the following interesting findings:
\begin{itemize}
    \item ReSGA achieves the lowest loss across the full stock group, as well as within five size-based stock groups: mega, large, small, micro, and nano. This consistent dominance indicates that ReSGA provides a broadly effective engine for VaR and ES forecasting, with its combination of learned cross-sectional group structure and retrieval-based temporal augmentation.
    
    \item Two spatial-temporal models tend to outperform seven temporal models, which in turn generally outperform two point-wise models. A notable exception is LANN, which performs worse than NN in most cases despite having richer inputs. By flattening the $S \times P$ characteristics matrix  into a single vector, LANN discards temporal ordering and therefore fails to model sequential dependence directly. As a result, it fails to capture dynamic patterns such as persistence or volatility clustering, illustrating that exploiting temporal information requires appropriate model architecture. Nevertheless, the overall ranking among all eleven universal models provides strong empirical support for the virtue of data complexity: Improvements from point-wise to temporal models reflect the value of incorporating temporal information, while further gains from temporal to spatial-temporal models highlight the importance of cross-sectional information. 

    \item All eleven universal models substantially outperform two econometric models: GAS and GARCH. A natural explanation is that GAS and GARCH are primarily designed to capture return dynamics and largely overlook rich firm-level information, which is crucial for VaR and ES forecasting. 
    Note that GAS and GARCH are estimated separately for each asset, whereas all universal models are trained on the pooled samples across assets. Hence, the above advantage of universal models over econometric models also points to a clear data scaling effect, highlighting the advantage of larger effective sample sizes and richer information content.
    
    \item Within the universal model family, moving beyond linear specifications yields clear gains, since nonlinear models (e.g., NN, LSTM, SGA, and ReSGA) deliver lower losses than the linear model in most cases. This finding suggests that nonlinearities and interaction effects among firm characteristics play an important role in tail risk forecasting.

    \item In terms of size-based stock groups, a clear monotonic pattern emerges: Larger firms systematically exhibit lower losses. This pattern is expected, as larger firms have greater liquidity, more standard published financial statements, and more stable trading environments, all of which contribute to their better predictability of tail risk. 
\end{itemize}

In addition to the out-of-sample average loss, we further evaluate the out-of-sample performance of all models using four statistical hypothesis tests: the DM in \cite{diebold1995}, MCS in \cite{Hansen2011model}, CC in \cite{Christoffersen1998EvaluatingIF}, and AESR in \cite{bayer2022}. See their detailed descriptions in \cref{sec:testing} of the supplementary materials. In terms of the values of $\ell_{\text{oos}}$ with $\mathbb{S}_t$ being the full stock group, the DM test aims to compare the prediction performance of any two models, and the MCS test searches for a batch of models that perform significantly better than the others. The CC and AESR tests provide statistical evidence of the validity of VaR and ES forecasts, respectively. Notably, we do not adopt the e-backtesting approach of \cite{wang2025backtest} to examine the validity of ES forecasts, since that framework is mainly designed for sequential monitoring and non-fixed-sample sizes, whereas our empirical exercise is based on fixed-sample out-of-sample evaluation.

\cref{tab:dm_test} reports the pairwise DM test statistics for all considered models. From this table, we have two clear findings. First, except for SGA, ReSGA dominates all other models with significantly lower values of $\ell_{\text{oos}}$. Second, all universal models deliver statistically significant improvements over econometric models. 

Next, starting from the full set of models, our MCS test iteratively eliminates inferior models at the 90\% confidence level, until only those statistically indistinguishable models remain to form the final model set. It turns out that only SGA and ReSGA are retained in the final model set. This outcome indicates that no other forecasting models perform as well as SGA and ReSGA in a statistical sense, underscoring the importance of spatial-temporal modeling for tail risk forecasting.

Moreover, we apply the CC and AESR tests to check the validity of VaR and ES forecasts for each stock at the significance level $\alpha \in \{0.01, 0.05, 0.10\}$, respectively, and then denote the proportion of stocks having valid VaR and ES forecasts as the pass rate. \cref{tab:cc_esr_test} reports pass rates with respect to VaR and ES forecasts across all models. From this table, we find that, except for ES at the level $\alpha=0.01$, the two spatial-temporal models, SGA and ReSGA have the largest values of pass rates. In particular, ReSGA shows a substantially larger pass rate than SGA for the prediction of ES at the level $\alpha=0.05$ or $0.10$.

Overall, the above statistical performance evaluations consistently demonstrate the state-of-the-art performance of ReSGA in tail risk forecasting, which in turn implies the virtue of data complexity.

\begin{sidewaystable}[p]
\centering
\caption{Pairwise DM test statistics.}
\label{tab:dm_test}
\begin{threeparttable}
\setlength{\tabcolsep}{2pt}
\begin{tabular}{lccccccccccccc}
\toprule
 & GAS & GARCH & Linear & NN & LANN & DLinear & LSTM & GRU & Informer & EInformer & DInformer & SGA & ReSGA \\
\midrule
GAS & -- &  &  &  &  &  &  &  &  &  &  &  &  \\
GARCH & 1.04 & -- &  &  &  &  &  &  &  &  &  &  &  \\
Linear & -3.35$^{***}$ & -2.39$^{**}$ & -- &  &  &  &  &  &  &  &  &  &  \\
NN & -3.39$^{***}$ & -2.41$^{**}$ & -0.85 & -- &  &  &  &  &  &  &  &  &  \\
LANN & -3.30$^{***}$ & -2.34$^{**}$ & 1.02 & 8.37$^{***}$ & -- &  &  &  &  &  &  &  &  \\
DLinear & -3.42$^{***}$ & -2.39$^{**}$ & -1.73$^{*}$ & -0.11 & -2.86$^{***}$ & -- &  &  &  &  &  &  &  \\
LSTM & -3.43$^{***}$ & -2.40$^{**}$ & -2.21$^{**}$ & -1.32 & -5.24$^{***}$ & -2.11$^{**}$ & -- &  &  &  &  &  &  \\
GRU & -3.45$^{***}$ & -2.41$^{**}$ & -2.46$^{**}$ & -1.80$^{*}$ & -5.65$^{***}$ & -2.89$^{***}$ & -2.23$^{**}$ & -- &  &  &  &  &  \\
Informer & -3.44$^{***}$ & -2.41$^{**}$ & -3.80$^{***}$ & -1.62 & -4.46$^{***}$ & -3.13$^{***}$ & -1.31 & -0.56 & -- &  &  &  &  \\
EInformer & -3.36$^{***}$ & -2.37$^{**}$ & -0.23 & 0.42 & -0.81 & 0.76 & 1.22 & 1.45 & 2.27$^{**}$ & -- &  &  &  \\
DInformer & -3.41$^{***}$ & -2.39$^{**}$ & -2.66$^{***}$ & -0.21 & -1.83$^{*}$ & -0.32 & 0.58 & 0.91 & 1.62 & -1.39 & -- &  &  \\
SGA & -3.51$^{***}$ & -2.43$^{**}$ & -4.62$^{***}$ & -3.56$^{***}$ & -8.01$^{***}$ & -4.40$^{***}$ & -3.05$^{***}$ & -2.17$^{**}$ & -1.40 & -2.25$^{**}$ & -2.13$^{**}$ & -- &  \\
ReSGA & -3.51$^{***}$ & -2.44$^{**}$ & -4.69$^{***}$ & -4.49$^{***}$ & -7.63$^{***}$ & -4.82$^{***}$ & -4.13$^{***}$ & -3.26$^{***}$ & -2.33$^{**}$ & -2.60$^{***}$ & -2.51$^{**}$ & -1.12 & -- \\
\bottomrule
\end{tabular}

\vspace{3pt}
\footnotesize
\textit{Notes.}
A lower-triangular matrix that reports the pairwise DM test statistic. Each entry shows the value of DM statistic by comparing the model in the row against the one in the column. Its negative (positive) value indicates that the row model yields lower (higher) loss, where the symbols ${***}$, ${**}$, and ${*}$ denote the rejection of the null hypothesis of equal prediction performance at the 1\%, 5\%, and 10\% levels, respectively.
\end{threeparttable}
\end{sidewaystable}

\begin{table}[!h]
\centering
\caption{Pass rates of stocks with valid VaR (or ES) predictions across different models.}
\label{tab:cc_esr_test}
\begin{threeparttable}
\setlength{\tabcolsep}{18pt}
\begin{tabular}{lccccccc}
\toprule
& \multicolumn{3}{c}{VaR} & & \multicolumn{3}{c}{ES} \\
\cmidrule{2-4} \cmidrule{6-8}
 & 0.01 & 0.05 & 0.10 & & 0.01 & 0.05 & 0.10 \\
\midrule

\multicolumn{4}{l}{\textit{Point-wise Models}} & \multicolumn{3}{l}{}\\[2pt]
\quad Linear      & 95.14 & 87.52 & 82.02 & & 78.90 & 70.13 & 64.73 \\
\quad NN         & 95.92 & 88.65 & 83.47 & & 67.92 & 60.04 & 54.75 \\[3pt]

\midrule
\multicolumn{4}{l}{\textit{Temporal Models}} & \multicolumn{3}{l}{}\\[2pt]
\quad LANN         & 95.08 & 87.17 & 82.26 & & 56.14 & 49.59 & 44.89 \\
\quad DLinear      & 96.41 & 89.95 & 84.88 & & 85.28 & 78.07 & 73.18 \\
\quad LSTM         & 97.28 & 91.54 & 86.53 & & 83.65 & 76.82 & 71.90 \\
\quad GRU          & 97.39 & 91.85 & 86.96 & & 83.93 & 77.66 & 72.56 \\
\quad Informer     & 97.04 & 90.89 & 86.04 & & 87.12 & 81.02 & 76.91 \\
\quad EInformer    & 96.23 & 89.43 & 84.24 & & 88.75 & 82.27 & 77.13 \\
\quad DInformer    & 96.20 & 89.70 & 84.72 & & 86.81 & 80.42 & 76.41 \\[3pt]

\midrule
\multicolumn{4}{l}{\textit{Spatial-temporal Models}} & \multicolumn{3}{l}{}\\[2pt]
\quad SGA         & \pmb{98.07} & 92.82 & 88.27 & & 87.63 & 80.48 & 75.56 \\
\quad ReSGA   & 97.86 & \pmb{92.84} & \pmb{88.33} & &
88.63 & \pmb{82.86} & \pmb{79.17} \\[3pt]

\midrule
\multicolumn{4}{l}{\textit{Econometric Models}} & \multicolumn{3}{l}{}\\[2pt]
\quad GAS         & 93.59 & 87.71 & 82.76 & & \pmb{88.84} & 82.81 & 78.03 \\
\quad GARCH       & 92.42 & 84.43 & 79.77 & & 79.72 & 71.99 & 66.86 \\

\bottomrule
\end{tabular}

\vspace{4pt}
\footnotesize
\textit{Notes.} Pass rate is the proportion of stocks that have valid VaR (or ES) predictions, assessed using the CC (or AESR) test at the significance level $\alpha \in \{0.01, 0.05, 0.10\}$. For each column, the largest pass rate is highlighted in boldface.
\end{threeparttable}
\end{table}

\subsection{Economic Performance Evaluation} 

To explore the economic implications of the VaR and ES, it is important to evaluate all tail risk models from an economic perspective. 
Following \citet{AtilganYigit2020}, we first assess economic gains using decile portfolios sorted on the predicted VaR or ES. This trading strategy is referred to as ``left-side momentum''.  It guides us to monthly rebalance ten decile portfolios (P1--P10), after sorting all stocks in descending order according to their predicted VaR or ES at the end of each month. Note that our predicted VaR and ES are negative in this paper, so P1 corresponds to the lowest-risk portfolio and P10 represents the highest-risk portfolio.

To keep the presentation concise, \cref{tab:portfolios_VaR_ES} reports portfolio results only for ReSGA, as the performance for other models is qualitatively similar. As shown in this table, under VaR-based sorting, average portfolio returns remain relatively stable up to P8 ($1.090\%$), but drop sharply to $0.619\%$ in P9 and turn negative in P10 ($-0.284\%$), while the Sharpe ratio falls to $-0.085$ in P10. This decline in returns is accompanied by a marked increase in downside risk: Maximum drawdown worsens from $0.499$ in P8 to $0.730$ in P9 and $0.807$ in P10. A similar pattern appears under ES-based sorting, with average returns decreasing from $1.196\%$ in P8 to $0.512\%$ in P9 and $-0.024\%$ in P10, and maximum drawdown deepening to $0.710$ in P9 and $0.810$ in P10.

The above findings indicate P9 and P10, which include stocks with high predicted tail risk, exhibit a substantially worse performance in terms of return, Sharpe ratio, and drawdown. \cite{AtilganYigit2020} attribute this pattern to behavioral underreaction to bad news and the persistence of left-tail risks, which concentrate losses in the most risk-exposed portfolios. Our findings mirror this stylized fact under both VaR- and ES-based sorting, supporting the view that tail-risk forecasts from the ReSGA can capture economically meaningful variation in downside risk across the cross-section.

\begin{table}[!h]
\centering
\caption{Out-of-sample performances of decile portfolios using the signal of VaR or ES.}
\label{tab:portfolios_VaR_ES}
\begin{threeparttable}
\setlength{\tabcolsep}{6.2pt}
\begin{tabular}{clccccccccccc}
\toprule
Signal & Metric
& P1 & P2 & P3 & P4 & P5 & P6 & P7 & P8 & P9 & P10 & H--L \\
\midrule
\multirow{3}{*}{VaR} & Avg
& 0.947 & 0.956 & 0.991 & 0.829 & 0.961 & 1.324 & 1.013 & 1.090 & 0.619 & -0.284 & 1.231 \\
& MDD
& 0.199 & 0.209 & 0.253 & 0.292 & 0.260 & 0.321 & 0.582 & 0.499 & 0.730 & 0.807 & 0.681 \\
& SR
& 0.909 & 0.788 & 0.708 & 0.570 & 0.585 & 0.724 & 0.431 & 0.431 & 0.217 & -0.085 & 0.412 \\
\midrule
\multirow{3}{*}{ES} & Avg
& 0.900 & 1.001 & 0.940 & 0.813 & 1.031 & 1.297 & 0.895 & 1.196 & 0.512 & -0.024 & 0.924 \\
& MDD
& 0.207 & 0.201 & 0.235 & 0.303 & 0.268 & 0.400 & 0.561 & 0.515 & 0.710 & 0.810 & 0.769 \\
& SR
& 0.826 & 0.818 & 0.697 & 0.515 & 0.663 & 0.694 & 0.389 & 0.487 & 0.185 & -0.007 & 0.298 \\
\bottomrule
\end{tabular}

\vspace{6pt}
\footnotesize
\textit{Notes.}
This table reports the performance of value-weighted decile portfolios constructed by predicted VaR and ES from the ReSGA model as sorting signal. At each month, stocks are sorted in descending order of the corresponding risk signal and assigned to ten portfolios (P1--P10). Portfolio P1 (High) contains stocks with the lowest tail risk, while P10 (Low) contains those with the highest tail risk. All portfolios are value-weighted using market equity and rebalanced monthly. Avg denotes the average portfolio return in percentage, MDD denotes the maximum drawdown, and SR denotes the annualized Sharpe ratio.
H--L (i.e., High--Low) is a long-short portfolio that buys P1 and sells P10.
\end{threeparttable}
\end{table}

Beyond the above left-side momentum, we further design a new size-enhanced left-side momentum strategy, which utilizes the trading signal motivated by the ``too big to fail'' phenomenon in the US stock market. Our key trading idea is to exploit the different economic implications of tail risk across firms of different sizes. Specifically, we assign strong buying signal to large-cap stocks with small values of ES, reflecting that downside risk in large firms often attracts market attention, policy support, or investor demand, and it thus can limit extreme losses and help preserve long-term value. In contrast, we assign strong selling signal to small-cap stocks with significantly negative ES, which often reflects idiosyncratic fragility, limited liquidity, or elevated default risk, rather than priced systematic risk.

Motivated by this idea, we construct the following trading signal:
\begin{align}\label{Cap_ES}
    \alpha_{i,t}
    = \bigl(\mathrm{Cap}_{i,t-1} - \overline{\mathrm{Cap}}_{t-1}\bigr)
    \times \bigl[1 - \exp(\widehat{\ES}_{i,t})\bigr],
\end{align}
where $\mathrm{Cap}_{i,t-1}$ denotes the log market capitalization of asset $i$ at time $t-1$, and $\overline{\mathrm{Cap}}_{t-1}$ is the corresponding cross-sectional mean. Here, the term $[1 - \exp(\widehat{\ES}_{i,t})]$ lies between $0$ and $1$, with its value decreasing as $\widehat{\ES}_{i,t}$ increases, serving as a proxy for tail-loss severity; the term $(\mathrm{Cap}_{i,t-1} - \overline{\mathrm{Cap}}_{t-1})$ tends to be positive for large-cap stocks and negative for small-cap stocks, acting as a classifier for firm size. Combining these two terms, the signal $\alpha_{i,t}$ in (\ref{Cap_ES}) assigns high values to large-cap stocks with severe predicted tail risk, and low values to small-cap stocks with similar tail risk.

To assess the economic performance of our size-enhanced left-side momentum strategy, we adopt the signal $\alpha_{i,t}$ to propose monthly rebalanced decile portfolios as before, with $\widehat{\ES}_{i,t}$ computed from different models. Moreover, we regress the monthly returns of the resulting H--L portfolios on the Fama–French five factors (\citealp{fama2015five}) to examine whether there is an abnormal return (alpha) unexplained by standard factors. \cref{tab:portfolio_MCES} reports the results of this analysis, and it delivers the following findings: 

\begin{itemize}
    \item The ReSGA-based strategy delivers the highest value of monthly average return, annualized Sharpe ratio and alpha for the H--L portfolio, outperforming all competing strategies. This result indicates that the best ES forecasting accuracy from ReSGA can translate into the strongest economic gains when combined with firm size information.
    
    \item Regardless of the choice of models, decile portfolios exhibit a clear and robust monotonic pattern: Moving from P1 to P10, average portfolio returns and annualized Sharpe ratios decline steadily and become strongly negative in the lowest deciles (P9 and P10). This pattern confirms that the signal $\alpha_{i,t}$ can effectively rank stocks that have economically meaningful size-enhanced tail risk exposure.
    
    \item For all strategies based on the universal models, the $p$-values of the estimated alphas are below $0.05$, whereas the corresponding $p$-values for the GAS- and GARCH-based strategies are $0.051$ and $0.128$, respectively. This contrast indicates that the econometric models exhibit relatively weak predictive power, highlighting the advantage of our proposed unified learning framework.
\end{itemize}

Taken together, \cref{tab:portfolios_VaR_ES,tab:portfolio_MCES} show that the superior advantage of ReSGA in learning tail risks can translate into economically meaningful gains.

\begin{table}[!htbp]
\centering
\caption{Out-of-sample performance of decile portfolios using the signal of $\alpha_{i,t}$ across different models.}
\label{tab:portfolio_MCES}
\setlength{\tabcolsep}{4pt}
\renewcommand{\arraystretch}{1.1}

\begin{tabular}{llcc ccccc ccccc}
\toprule
 & Metric & P1 & P2 & P3 & P4 & P5 & P6 & P7 & P8 & P9 & P10 & H--L & Alpha\\
\midrule

\multicolumn{3}{l}{\textit{Point-wise Models}} & \multicolumn{11}{l}{}\\[2pt]

\multirow{2}{*}{\quad Linear} & Avg & 0.931 & 0.804 & 0.761 & 0.830 & 0.534 & 0.587 & 0.469 & 0.461 & -0.193 & -0.887 & 1.818 & 1.477 \\
 & SR & 0.732 & 0.597 & 0.500 & 0.510 & 0.306 & 0.328 & 0.264 & 0.217 & -0.076 & -0.274 & 0.673 & (0.030) \\
\multirow{2}{*}{\quad NN} & Avg  & 0.926 & 0.764 & 0.741 & 0.820 & 0.535 & 0.538 & 0.651 & 0.314 & -0.223 & -0.901 & 1.827 & 1.510 \\
 & SR & 0.720 & 0.582 & 0.491 & 0.494 & 0.307 & 0.307 & 0.357 & 0.148 & -0.087 & -0.272 & 0.659 & (0.035) \\
\midrule

\multicolumn{3}{l}{\textit{Temporal Models}} & \multicolumn{11}{l}{}\\[2pt]

\multirow{2}{*}{\quad LANN} & Avg  & 0.934 & 0.809 & 0.825 & 0.821 & 0.657 & 0.646 & 0.575 & 0.427 & -0.174 & -1.039 & 1.973 & 1.625 \\
 & SR & 0.731 & 0.612 & 0.551 & 0.500 & 0.380 & 0.356 & 0.307 & 0.196 & -0.066 & -0.317 & 0.723 & (0.019) \\
\multirow{2}{*}{\quad DLinear} & Avg  & 0.928 & 0.767 & 0.785 & 0.827 & 0.646 & 0.643 & 0.591 & 0.503 & -0.234 & -0.947 & 1.876 & 1.516 \\
 & SR & 0.725 & 0.590 & 0.527 & 0.513 & 0.372 & 0.355 & 0.314 & 0.231 & -0.088 & -0.292 & 0.691 & (0.032) \\
\multirow{2}{*}{\quad LSTM} & Avg  & 0.926 & 0.830 & 0.807 & 0.843 & 0.623 & 0.645 & 0.640 & 0.380 & -0.146 & -1.108 & 2.034 & 1.720 \\
 & SR & 0.722 & 0.643 & 0.552 & 0.525 & 0.362 & 0.354 & 0.343 & 0.175 & -0.055 & -0.338 & 0.744 & (0.010) \\
\multirow{2}{*}{\quad GRU} & Avg & 0.916 & 0.855 & 0.836 & 0.831 & 0.679 & 0.656 & 0.586 & 0.447 & -0.212 & -0.944 & 1.860 & 1.536 \\
 & SR & 0.713 & 0.658 & 0.575 & 0.519 & 0.393 & 0.362 & 0.312 & 0.205 & -0.082 & -0.286 & 0.671 & (0.030) \\
\multirow{2}{*}{\quad Informer} & Avg  & 0.908 & 0.816 & 0.796 & 0.835 & 0.609 & 0.632 & 0.615 & 0.464 & -0.133 & -1.095 & 2.002 & 1.680 \\
 & SR & 0.709 & 0.623 & 0.531 & 0.510 & 0.355 & 0.349 & 0.328 & 0.212 & -0.050 & -0.329 & 0.719 & (0.013) \\
\multirow{2}{*}{\quad EInformer} & Avg  & 0.938 & 0.758 & 0.815 & 0.801 & 0.643 & 0.632 & 0.605 & 0.434 & -0.146 & -0.911 & 1.850 & 1.501 \\
 & SR & 0.737 & 0.582 & 0.555 & 0.494 & 0.373 & 0.350 & 0.321 & 0.199 & -0.055 & -0.279 & 0.678 & (0.029) \\
\multirow{2}{*}{\quad DInformer} & Avg  & 0.913 & 0.807 & 0.776 & 0.841 & 0.625 & 0.646 & 0.548 & 0.512 & -0.149 & -0.949 & 1.862 & 1.519 \\
 & SR & 0.714 & 0.604 & 0.509 & 0.508 & 0.363 & 0.356 & 0.292 & 0.236 & -0.056 & -0.294 & 0.689 & (0.023) \\
\midrule

\multicolumn{3}{l}{\textit{Spatial-temporal Models}} & \multicolumn{11}{l}{}\\[2pt]

\multirow{2}{*}{\quad SGA} & Avg  & 0.918 & 0.800 & 0.830 & 0.797 & 0.693 & 0.679 & 0.554 & 0.346 & -0.049 & -1.251 & 2.169 & 1.812 \\
 & SR & 0.719 & 0.612 & 0.566 & 0.500 & 0.403 & 0.380 & 0.290 & 0.160 & -0.018 & -0.378 & 0.780 & (0.013) \\
\multirow{2}{*}{\quad ReSGA} & Avg  & 0.905 & 0.806 & 0.868 & 0.819 & 0.697 & 0.702 & 0.552 & 0.336 & -0.069 & -1.308 & \pmb{2.213} & \pmb{1.842} \\
 & SR & 0.710 & 0.613 & 0.585 & 0.510 & 0.401 & 0.391 & 0.291 & 0.153 & -0.026 & -0.392 & \pmb{0.787} & (0.013) \\
\midrule

\multicolumn{3}{l}{\textit{Econometric Models}} & \multicolumn{11}{l}{}\\[2pt]

\multirow{2}{*}{\quad GAS} & Avg  & 0.917 & 0.831 & 0.767 & 0.785 & 0.678 & 0.636 & 0.693 & 0.513 & 0.097 & -0.705 & 1.621 & 1.217 \\
 & SR & 0.703 & 0.655 & 0.587 & 0.519 & 0.394 & 0.348 & 0.357 & 0.238 & 0.038 & -0.237 & 0.672 & (0.051) \\
\multirow{2}{*}{\quad GARCH} & Avg  & 0.899 & 0.809 & 0.837 & 0.806 & 0.714 & 0.712 & 0.556 & 0.539 & 0.037 & -0.496 & 1.395 & 1.012 \\
 & SR & 0.695 & 0.630 & 0.629 & 0.546 & 0.418 & 0.387 & 0.285 & 0.252 & 0.015 & -0.161 & 0.562 & (0.128) \\
\bottomrule
\end{tabular}

\vspace{6pt}
\begin{minipage}{\linewidth}
\footnotesize
\textit{Notes.} This table reports the performance of value-weighted decile portfolios constructed using $\alpha_{i,t}$ in \eqref{Cap_ES} as sorting signal, where $\widehat{\ES}_{i,t}$ is computed from each model. The column ``Alpha’’ reports the estimated alpha (intercept) and the corresponding $p$-value of the $t$-statistic (in parentheses) in the Fama–French five-factor regression. For the H--L portfolios, the largest values of Avg, SR, and Alpha are highlighted in boldface. Other descriptions are consistent with those in \cref{tab:portfolios_VaR_ES}.
\end{minipage}
\end{table}

\subsection{Scaling Performance Evaluation}\label{sec:scaling}

To examine whether there is virtue of model complexity in learning VaR and ES, we begin by evaluating scaling performance with respect to parameter size for all considered machine-learning-based models (i.e., all universal models excluding Linear). Specifically, we adopt a within-architecture scaling strategy: For each model, we vary the number of model parameters by adjusting width/depth-related hyperparameters, while keeping all other mechanisms unchanged\footnote{For example, in the NN model, parameter size is controlled by the hidden-state dimension and the number of hidden layers. See \cref{sec:fc} for more details.}. This approach produces multiple variants of each model with substantially different numbers of learnable parameters. The connection between hyperparameter choices and parameter size for each model is reported in \cref{tab:hyperparam_settings} of the supplementary materials.

\cref{tab:param_scaling_compact} details the out-of-sample average loss of different models when their parameter size varies over several orders of magnitude. It includes exact loss values, while its accompanying figure visualizes the corresponding scaling patterns\footnote{For each model, we fit a polynomial function linking loss to log parameter size, with the polynomial order selected by the Akaike information criterion \citep{AIC}.}. Combining both pieces of evidence, we obtain three main interesting findings:
\begin{itemize}

    \item ReSGA consistently outperforms all competitors once the parameter size reaches $10^5$ or above, with SGA ranking second, whereas  Informer achieves the lowest loss at smaller scales. This pattern is evident from the bold entries in the table and the crossing behavior in the figure: ReSGA and SGA improve sharply from small to intermediate scales, while Informer deteriorates once its parameter size becomes large. These findings indicate that models with richer spatial-temporal inputs, such as SGA and ReSGA, need sufficient capacity in parameters to exploit useful information from inputs, while temporal models can perform well with relatively few parameters. In contrast, the point-wise model of NN benefits little from expanded capacity, likely because its input information is inherently limited.

    \item There is no systematic evidence for a general virtue of model complexity. Except for LSTM, the parameter-richest specification (around $10^7$ parameters) does not deliver the best performance within each model family. The figure makes this non-monotonicity especially visible: Most loss curves flatten or turn upward after intermediate scales. Even for LSTM, where loss performance improves with scale, the gains are modest. Meanwhile, LSTM with around $10^7$ parameters still underperforms ReSGA with $10^5$ parameters. This absence of monotonic and substantial improvement suggests that simply increasing model size alone does not guarantee better tail-risk forecasts.

    \item As shown by \citet{kaplan2020scalinglawsneurallanguage}, in natural language processing, optimal performance is often achieved when model size is of the same order as the in-sample data size. This principle is only partially supported in our financial task. To be specific, ReSGA and SGA achieve their lowest out-of-sample losses at around $10^6$ parameters, which is comparable to the effective training sample size (approximately $2\times 10^6$); however, this alignment is not universal, since other models show no clear match between optimal parameter size and sample size. These observations point to the heterogeneity in the virtue of data complexity: Spatial-temporal models could exploit richer inputs, allowing for more capacity in parameters to be used effectively, whereas temporal and point-wise models rely on less information and consequently obtain fewer benefits from increased parameterization.
    
\end{itemize}

\begin{table}[!ht]
\centering
\caption{Out-of-sample average loss across different parameter sizes.}
\label{tab:param_scaling_compact}
\begin{threeparttable}

\setlength{\tabcolsep}{22pt}

\begin{tabular}{lccccc}
\toprule
Model & $\approx10^3$ & $\approx10^4$ & $\approx10^5$ & $\approx10^6$ & $\approx10^7$ \\
\midrule
NN
& $3.3322$
& $\underline{3.3128}$
& $3.3236$
& $3.3568$ 
& $3.6057$ \\

LANN
& --
& $3.3440$
& $\underline{3.3923}$
& $3.4627$ 
& $3.7543$ \\

DLinear
& $3.3180$
& $3.3166$
& $\underline{3.3153}$
& $3.3174$ 
& $3.3153$ \\

LSTM
& $3.3184$
& $3.3063$
& $3.3083$
& $3.3088$ 
& $\underline{3.3024}$ \\

GRU
& $3.3067$
& $3.3065$
& $3.3022$
& $\underline{3.2997}$ 
& $3.3003$ \\

Informer
& $\pmb{3.2999}$
& $\pmb{\underline{3.2985}}$
& $3.3361$
& $3.3812$ 
& $3.4128$ \\

EInformer
& $\underline{3.3136}$
& $3.3163$
& $3.3187$
& $3.3276$ 
& $3.3806$ \\

DInformer
& $3.3134$
& $\underline{3.3097}$
& $3.3276$
& $3.3310$ 
& $3.3428$ \\

SGA
& $3.3319$
& $3.3234$
& $3.2934$
& $\underline{3.2868}$
& $3.2965$ \\

ReSGA
& $3.3417$
& $3.3250$
& $\pmb{3.2915}$
& $\pmb{\underline{3.2793}}$
& $\pmb{3.2879}$ \\

\bottomrule
\end{tabular}

\vspace{0.5em}

\includegraphics[width=0.95\textwidth]{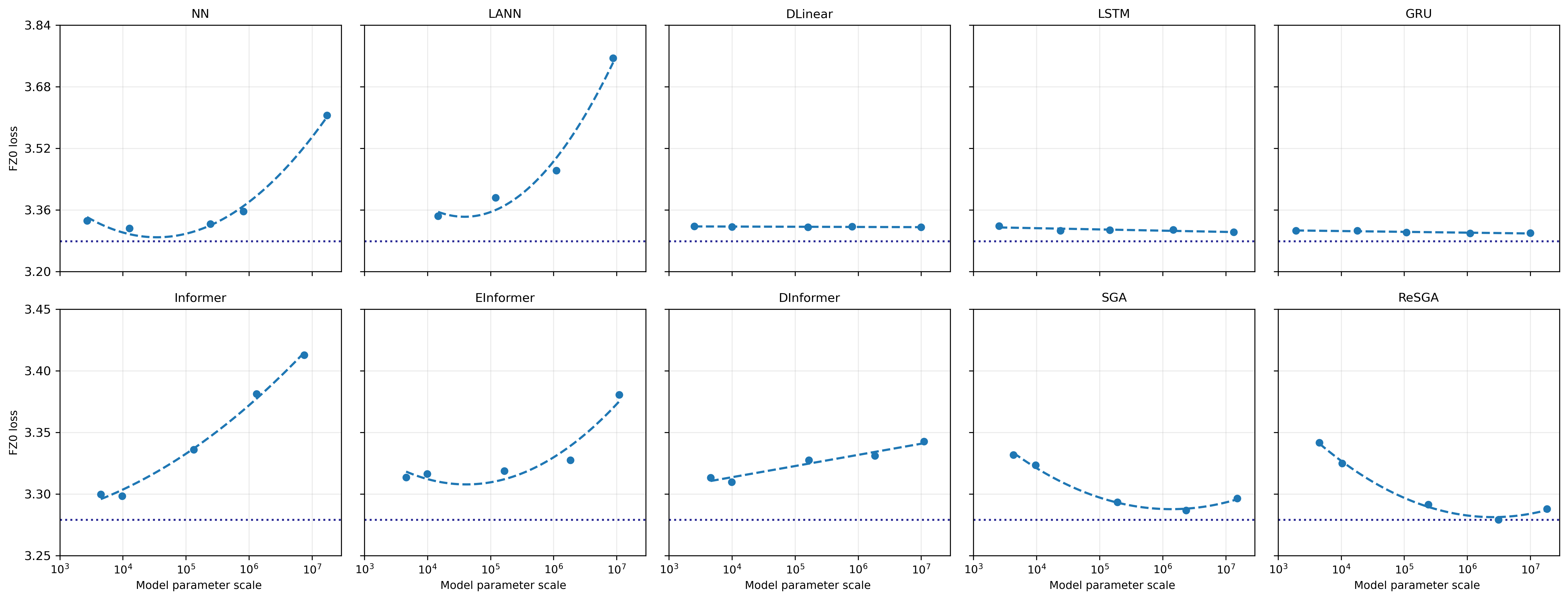}

\vspace{0.5em}

\begin{tablenotes}[flushleft]
\footnotesize
\item \textit{Notes.} The table (upper panel) reports out-of-sample average loss $\ell_{\text{oos}}$ for different models with varying parameter sizes (approximately $10^3$ to $10^7$ parameters), with $\mathbb{S}_t$ formed by all stocks. Underlined entries indicate the lowest loss in each row, and the lowest loss in each column is highlighted in boldface. ``--'' indicates no feasible hyperparameters for constructing such parameter size. The accompanying figure (bottom panel) visualizes the corresponding scaling behavior of average loss across different parameter scales. Here, the dotted line stands for the best loss ($=3.2793$).
\end{tablenotes}

\end{threeparttable}
\end{table}

Meanwhile, \cref{tab:hl_param_scaling_new} reports the out-of-sample Sharpe ratios of H--L portfolios as parameter size varies, with an accompanying figure to visualize the trend of Sharpe ratio across parameter scales. From \cref{tab:hl_param_scaling_new}, we find that Sharpe ratio fails to exhibit a clear monotonic pattern as the parameter size increases. First, the best-performing parameter size differs across models: Most models peak at intermediate scales, while LSTM and Informer attain their highest Sharpe ratios at the largest scale. Second, the leading model also changes across parameter scales. These mixed patterns indicate that increasing model complexity alone does not reliably improve portfolio-level gains.

\begin{table}[!ht]
\centering
\caption{Out-of-sample H--L portfolio Sharpe ratios across different parameter sizes.}
\label{tab:hl_param_scaling_new}
\begin{threeparttable}

\setlength{\tabcolsep}{24pt}

\begin{tabular}{lccccc}
\toprule
Model & $\approx10^3$ & $\approx10^4$ & $\approx10^5$ & $\approx10^6$ & $\approx10^7$ \\
\midrule

NN     & 0.659 & 0.659 & 0.660 & \underline{0.736} & 0.718 \\
LANN   & --    & 0.723 & \underline{$\pmb{0.794}$} & 0.703 & 0.695 \\
DLinear& 0.666 & \underline{0.691} & 0.666 & 0.648 & 0.650 \\
LSTM   & 0.670 & $\pmb{0.744}$ & 0.671 & 0.710 & \underline{$\pmb{0.756}$} \\
GRU    & 0.714 & 0.700 & 0.671 & \underline{0.740} & 0.697 \\
Informer & $\pmb{0.717}$ & 0.719 & 0.702 & 0.725 & \underline{0.729} \\
EInformer & 0.639 & 0.694 & \underline{0.713} & 0.678 & 0.712 \\
DInformer & 0.689 & \underline{0.711} & 0.647 & 0.686 & 0.658 \\
SGA       & 0.633 & 0.658 & 0.710 & \underline{0.780} & 0.755 \\
ReSGA     & 0.622 & 0.673 & 0.765 & \underline{$\pmb{0.787}$} & 0.749 \\

\bottomrule
\end{tabular}

\vspace{0.5em}

\includegraphics[width=0.95\textwidth]{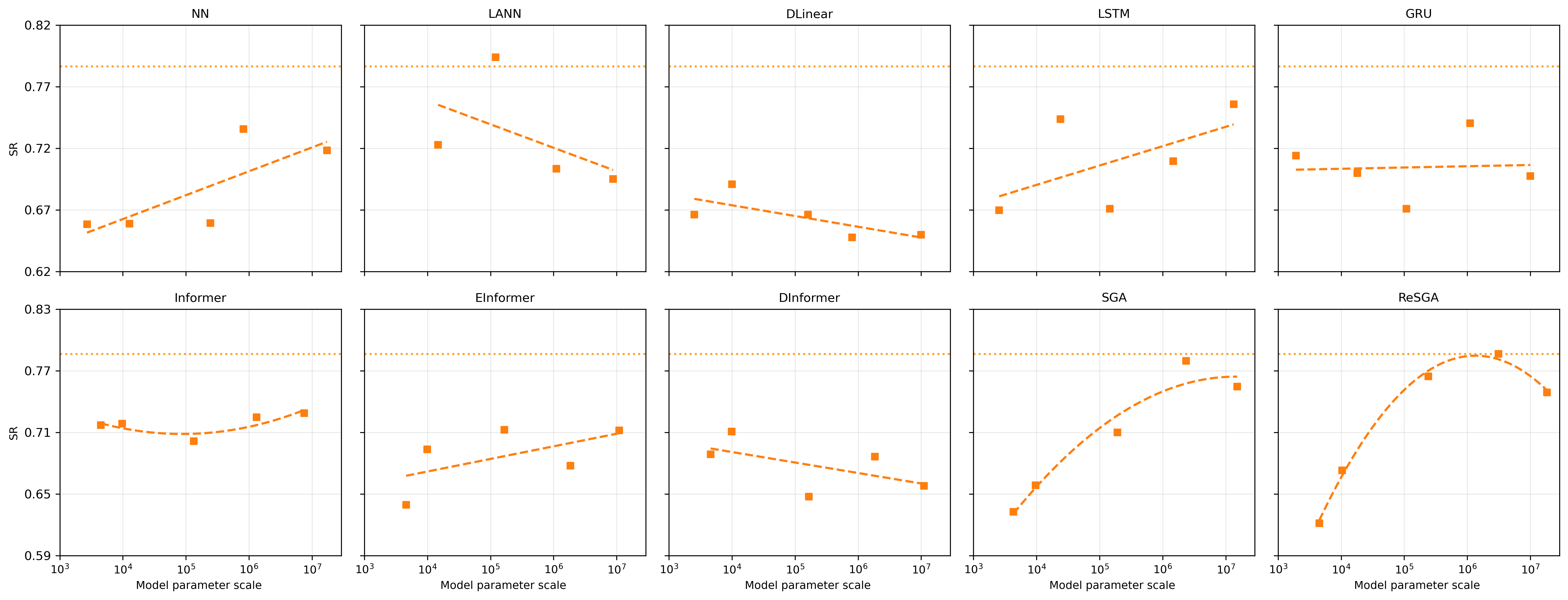}

\vspace{0.5em}

\begin{tablenotes}[flushleft]
\footnotesize

\item \textit{Notes.} The table (upper panel) reports the annualized Sharpe ratio (SR) of H--L portfolios under different parameter scales. 
Underlined entries indicate the highest SR in each row, and the highest SR in each column is highlighted in boldface. ``--'' indicates that the corresponding result is unavailable. The accompanying figure (bottom panel) plots SRs to visualize the scaling pattern across different parameter scales from the economic perspective. Here, the dotted line stands for the best SR ($=0.794$).
\end{tablenotes}
\end{threeparttable}
\end{table}

To further examine the virtue of data complexity in learning VaR and ES, we next evaluate scaling performance with respect to in-sample data size. Specifically, we hold each model fixed and vary the fraction of available in-sample data used for training and validation, from $1\%$ to $100\%$ of the full in-sample dataset. For each fraction, we draw the in-sample data uniformly at random from the full in-sample dataset, and  keep the hyperparameter configuration fixed at the values selected using the full in-sample dataset. This design holds model complexity fixed, and it isolates the role of in-sample data size that is one key dimension of data complexity.

\cref{tab:data_scaling_fz0} reports the out-of-sample average loss across different in-sample data sizes, while its accompanying figure visualizes how these losses evolve as more in-sample data become available. From this table and its accompanying figure, we can reach the following findings: 
\begin{itemize}
    \item All models benefit from more in-sample data (from $1\%$ to $100\%$). With the exception of EInformer, each model attains its best performance when the full in-sample dataset is available. Meanwhile, the loss curves in the figure mostly present monotonic pattern and decline as the sample fraction increases.

    \item The best-performing model changes with the amount of in-sample data. When in-sample data are scarce (only $1\%$, $5\%$, or $10\%$), temporal models such as GRU and Informer outperform the other models. Once the fraction of used in-sample data reaches $25\%$ or above, the spatial-temporal models (SGA and especially ReSGA) become dominant, indicating that these models require sufficient data to fully exploit cross-sectional sharing and long-term temporal dependencies.
    
    \item Point-wise model, NN, exhibits the largest improvement as the amount of in-sample data grows, consistent with its reliance on pooled cross-sectional information to compensate for weaker inductive structure. Nevertheless, even with the full in-sample dataset, NN remains clearly outperformed by the best temporal and spatial-temporal models. Thus, more data help broadly, with the greatest statistical gains when the model can use the richest in-sample information effectively.
\end{itemize}

\begin{table}[!th]
\centering
\caption{Out-of-sample average loss across different in-sample data sizes.}
\label{tab:data_scaling_fz0}
\begin{threeparttable}

\setlength{\tabcolsep}{17.5pt}
\begin{tabular}{lcccccc}
\toprule
Model & 1\% & 5\% & 10\% & 25\% & 50\% & 100\% \\
\midrule
NN        & 4.4749 & 4.4075 & 4.3214 & 3.6949 & 3.3621 & \underline{3.3322}  \\
LANN      & 4.4241 & 4.3898 & 4.3771 & 4.0613 & 3.4774 & \underline{3.3440}  \\
DLinear   & 3.3679 & 3.3363 & 3.3248 & 3.3201 & 3.3175 & \underline{3.3166}  \\
LSTM      & 3.3506 & 3.3227 & 3.3147 & 3.3102 & 3.3090 & \underline{3.3063}  \\
GRU       & $\pmb{3.3387}$ & $\pmb{3.3172}$ & 3.3117 & 3.3093 & 3.3068 & \underline{3.3022}  \\
Informer  & 3.3585 & 3.3262 & $\pmb{3.3094}$ & 3.2996 & 3.2995 & \underline{3.2985}  \\
EInformer & 3.3470 & 3.3363 & 3.3350 & 3.3329 & \underline{3.3269} & 3.3276  \\
DInformer & 3.6000 & 3.3531 & 3.3259 & 3.3184 & 3.3164 & \underline{3.3134}  \\
SGA       & 3.4851 & 3.3680 & 3.3246 & 3.2995 & $\pmb{3.2907}$ & \underline{3.2868}  \\
ReSGA     & 3.4938 & 3.3734 & 3.3378 & $\pmb{3.2933}$ & 3.2936 & $\underline{\pmb{3.2793}}$  \\
\bottomrule
\end{tabular}

\vspace{0.5em}

\includegraphics[width=0.95\textwidth]{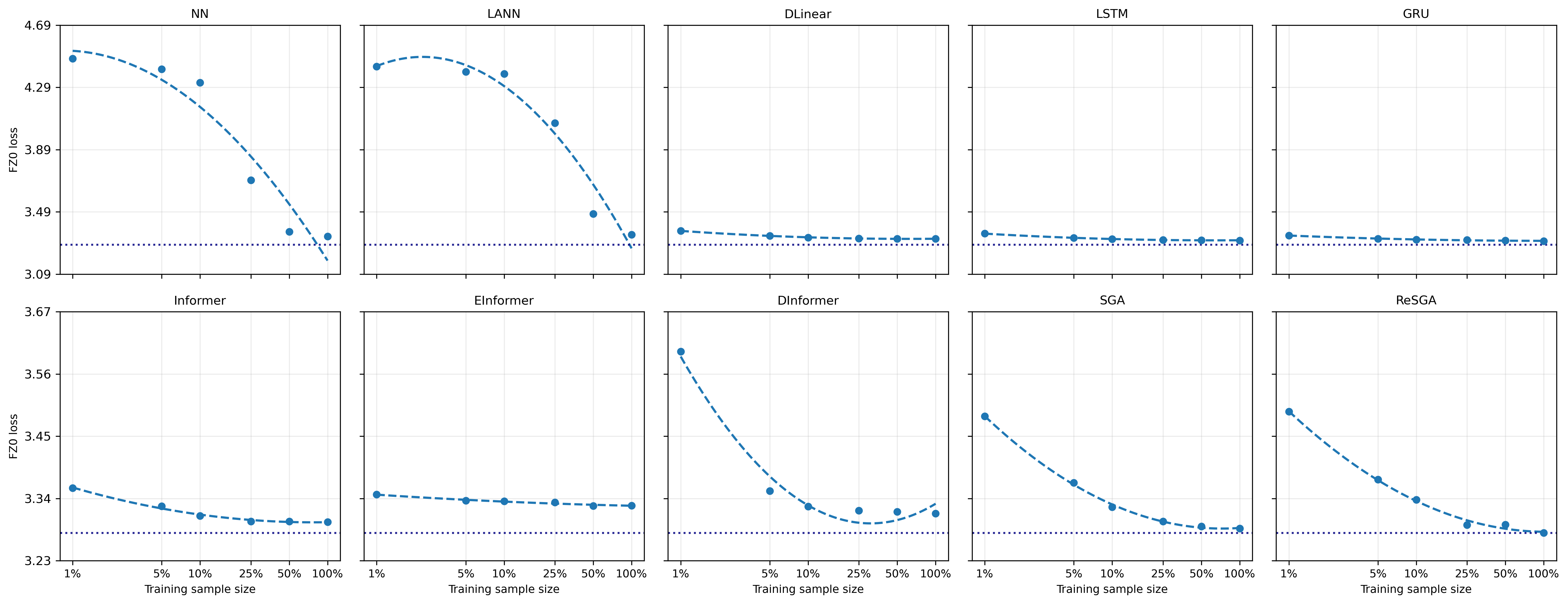}

\vspace{0.5em}
\begin{tablenotes}[flushleft]
\footnotesize
\item \textit{Notes.} The table (upper panel) reports the out-of-sample average loss of each model, as the in-sample data size varies across $\{1\%, 5\%, 10\%, 25\%, 50\%, 100\%\}$ of the full in-sample dataset. The accompanying figure (bottom panel) visualizes the corresponding scaling behavior of average loss across different in-sample data sizes. Other descriptions are consistent with those in \cref{tab:param_scaling_compact}.
\end{tablenotes}
\end{threeparttable}
\end{table}

Beyond evaluation from loss, \cref{tab:hl_sample_size_scaling_new} reports the out-of-sample Sharpe ratios of H--L portfolios when the in-sample data size varies, while its accompanying figure visualizes the scaling patterns of Sharpe ratio. Being broadly consistent with the loss results in \cref{tab:data_scaling_fz0}, this table delivers the following notable results from the economic perspective:  
\begin{itemize}
    \item ReSGA delivers the highest Sharpe ratio across different models once the in-sample data size reaches at least $25\%$ of the full dataset.
    \item The fitted curves show that portfolio performance generally improves as more in-sample data become available, although the pattern is not strictly monotonic for every model.
\end{itemize}
Taken together, the above results show that a larger in-sample dataset can lead to stronger economic performance, especially for models that can effectively exploit rich information.

\begin{table}[!th]
\centering
\caption{Out-of-sample H--L portfolio Sharpe ratios across different in-sample sizes.}
\label{tab:hl_sample_size_scaling_new}
\begin{threeparttable}
\setlength{\tabcolsep}{18pt}

\begin{tabular}{lcccccc}
\toprule
Model & 1\% & 5\% & 10\% & 25\% & 50\% & 100\% \\
\midrule

NN       & 0.324 & 0.404 & 0.525 & 0.676 & \underline{0.732} & 0.659 \\
LANN     & 0.328 & 0.320 & 0.346 & 0.494 & 0.666 & \underline{0.723} \\
DLinear  & 0.588 & 0.653 & 0.630 & 0.655 & 0.677 & \underline{0.691} \\
LSTM     & 0.608 & 0.666 & 0.666 & 0.662 & 0.712 & \underline{0.744} \\
GRU      & 0.607 & \underline{$\pmb{0.692}$} & 0.670 & 0.661 & 0.664 & 0.671 \\
Informer & $\pmb{0.656}$ & 0.665 & $\pmb{0.675}$ & 0.690 & 0.687 & \underline{0.719} \\
EInformer& 0.640 & 0.624 & 0.615 & 0.652 & \underline{0.685} & 0.678 \\
DInformer& 0.523 & 0.659 & 0.665 & 0.683 & 0.673 & \underline{0.689} \\
SGA      & 0.617 & 0.629 & 0.628 & 0.685 & 0.763 & \underline{0.780} \\
ReSGA    & 0.618 & 0.632 & 0.663 & $\pmb{0.705}$ & $\pmb{0.782}$ & \underline{$\pmb{0.787}$} \\

\bottomrule
\end{tabular}

\vspace{0.5em}

\includegraphics[width=0.95\textwidth]{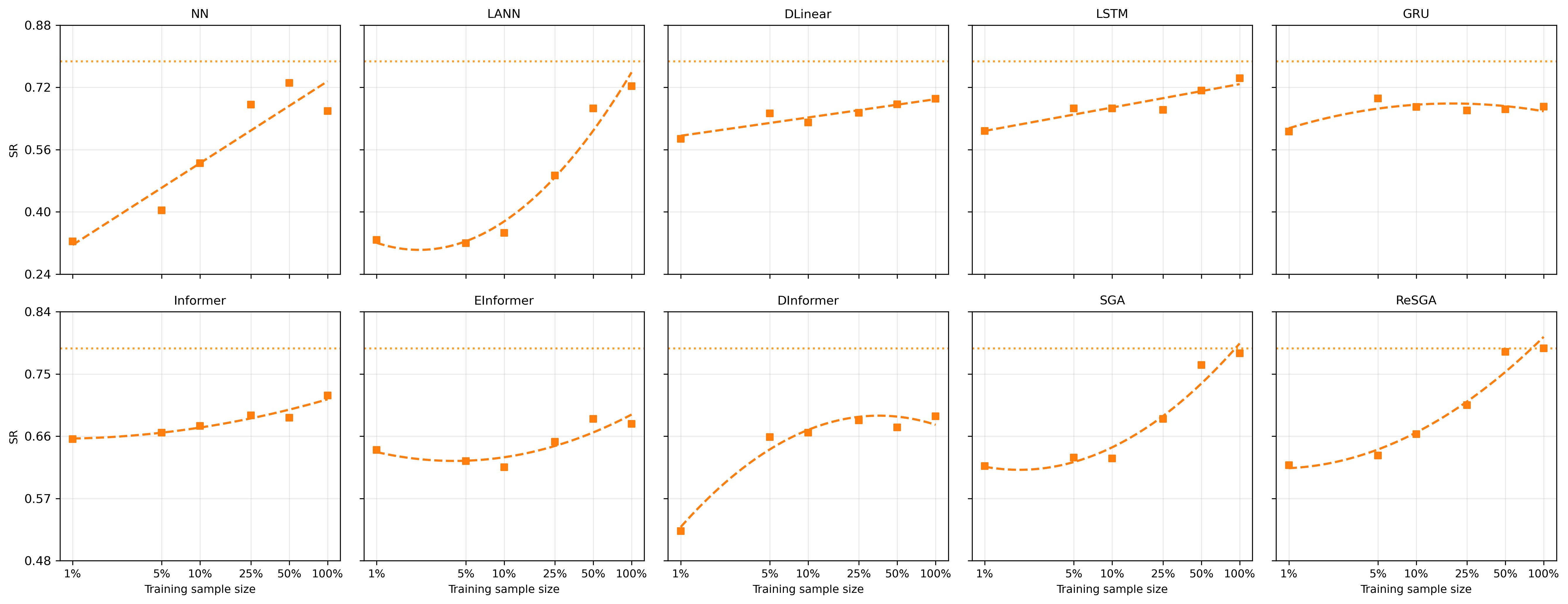}

\vspace{0.5em}

\begin{tablenotes}[flushleft]
\footnotesize
\item \textit{Notes.} The table (upper panel)  reports the annualized Sharpe ratio (SR) of H--L portfolios under different in-sample data sizes. 
The accompanying figure (bottom panel) plots SRs to visualize the scaling pattern across different in-sample data sizes from the economic perspective. Other descriptions are consistent with those in \cref{tab:hl_param_scaling_new}.
\end{tablenotes}
\end{threeparttable}
\end{table}

Overall, our analysis results provide little support for the virtue of model complexity. Instead, they highlight the importance of data complexity: Performance gains arise mainly from models that can exploit richer information with the use of a larger in-sample data size, rather than a larger parameter size alone. While the available sample size is always limited in finance, this motivates the next scaling dimension: expanding the effective information content of the input with appropriate model specification and enough parameters, which appears to be a more effective path to improving tail risk forecasting accuracy than simply increasing model size.

\subsection{Group Importance}

A further empirical question concerns which firm characteristics are most important for forecasting tail risk. Due to the superior performance of ReSGA from both statistical and economic perspectives, we utilize this model to tackle the above question. Assessing importance at the level of individual characteristics, however, is challenging due to the well-known dilution problem: When many characteristics are highly correlated, their marginal contributions become spread across related variables, obscuring the true economic drivers of predictive performance \citep{li2025}. To address this issue, we assess importance at the group level rather than at the level of individual characteristics. Specifically, we adopt the 13-category taxonomy of \cite{jensen2023}, which organizes the 153 firm characteristics into economically interpretable groups, including Low Risk, Value, Quality, Low Leverage, Momentum, Size, Profit Growth, Short-Term Reversal, Seasonality, Investment, Profitability, Debt Issuance, and Accruals. As argued by \cite{li2025}, grouping characteristics reduces redundancy among highly correlated signals and provides a clearer decomposition of predictive content. This approach allows us to evaluate the incremental value of entire economic themes, rather than attributing importance to individual characteristics that may proxy for similar information.

We quantify group importance via a drop-group procedure. To be specific, for each group (denoted as $\mathbb{G}$), we follow \cite{gu2020empirical} to construct perturbed testing samples in which all characteristics in group $\mathbb{G}$ are set to zero, while all remaining characteristics are left unchanged. By passing these perturbed samples through the trained model to generate new forecasts, we recompute the out-of-sample average loss in \eqref{oos_loss} on these new forecasts, with the set $\mathbb{S}_t$ containing all examined stocks. Then, we measure the importance of group $\mathbb{G}$ by the increase in loss of perturbed samples relative to that obtained using the original testing samples. After computing importance for all 13 groups, we set negative values to zero to avoid spurious attribution and normalize the remaining values to sum to one.

\begin{figure}[!h]
    \centering
    \includegraphics[width=1.0\textwidth]{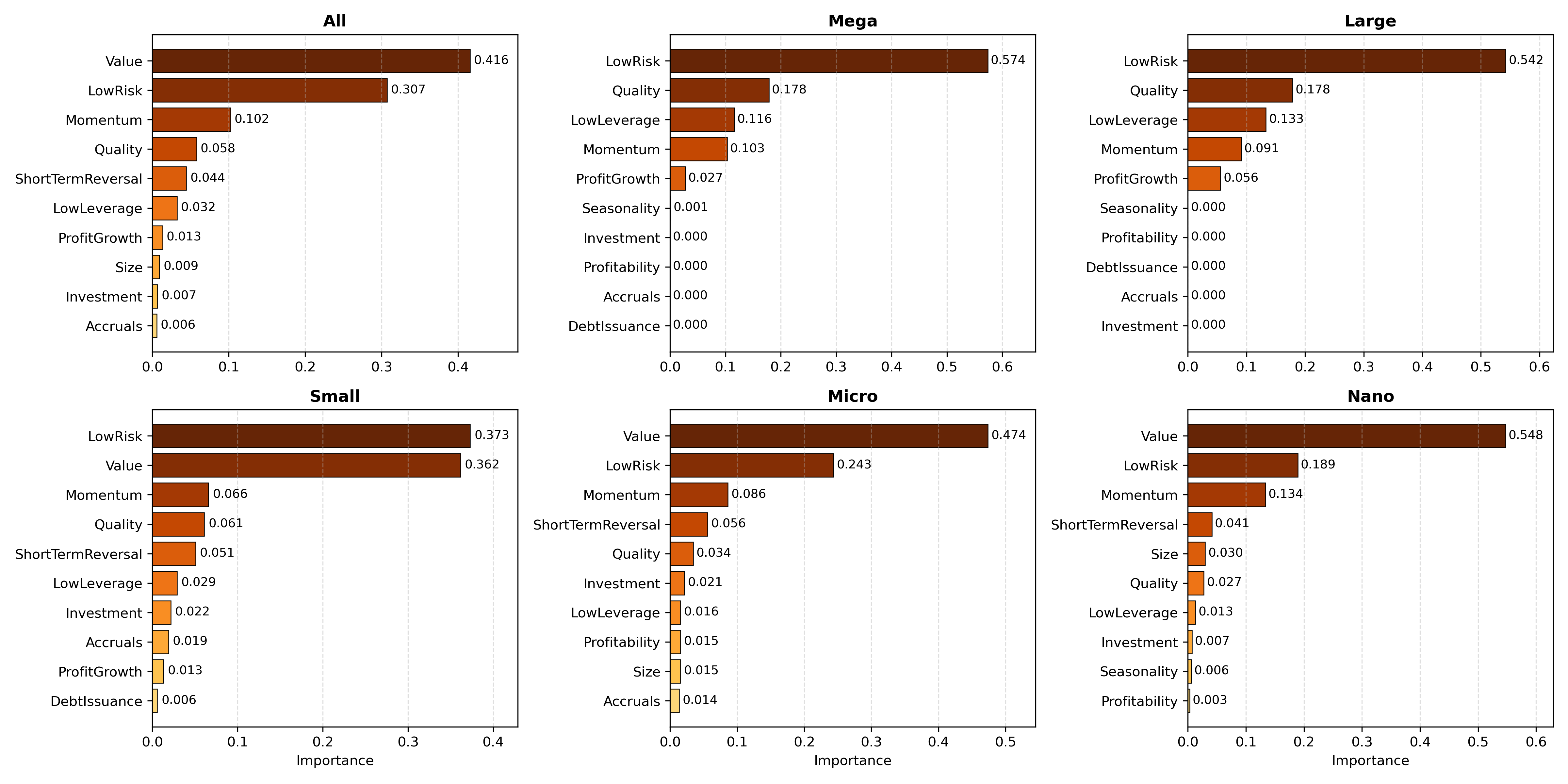}
    \caption{Group importance from ReSGA during the out-of-sample period.}
    \label{fig:overall_group_importance}
\end{figure}

\cref{fig:overall_group_importance} exhibits the ten most important groups from ReSGA over the out-of-sample period (2014.01--2023.12). From this figure, two clear and economically intuitive patterns emerge.
\begin{itemize}
    \item The top five groups (Value, Low Risk, Momentum, Quality, and Short-Term Reversal) together account for more than 90\% of importance for the full dataset. This concentration indicates that tail risk predictability is driven by a small set of groups rather than being evenly distributed across all characteristics. Similar findings can be found in the asset pricing literature (\citealp{gu2020empirical, Gu2021AutoencoderAP}; \citealp{yang2024asset}). Note that the concentration is even stronger for mega- and large-cap stocks, with top five groups receiving over $99\%$ importance, whereas the top five groups in small-, micro-, and nano-stocks contribute $91\%$, $89\%$, and $94\%$, respectively.
    
    \item The ranking of the leading groups shows both stability and systematic variation across different levels of market capitalization. Generally speaking, Low Risk and Momentum are consistently important regardless of the level of market capitalization. This is intuitive, as volatility- and beta-related characteristics in Low Risk are closely linked to tail risk, while Momentum captures persistent return dynamics that affect tail outcomes. Apart from these two common drivers, for mega- and large-cap stocks, Quality and Low Leverage are more influential, suggesting that tail risk variation among large firms is closely tied to balance-sheet strength and financial stability; in contrast, for small-, micro-, and nano-cap stocks, Value and Short-Term Reversal become more crucial, where valuation performance and short-horizon price reversals are expected to closely relate to distress risk, illiquidity, and left-tail outcomes of those stocks.
\end{itemize}

\begin{figure}[!ht]
    \centering
    \includegraphics[width=\textwidth]{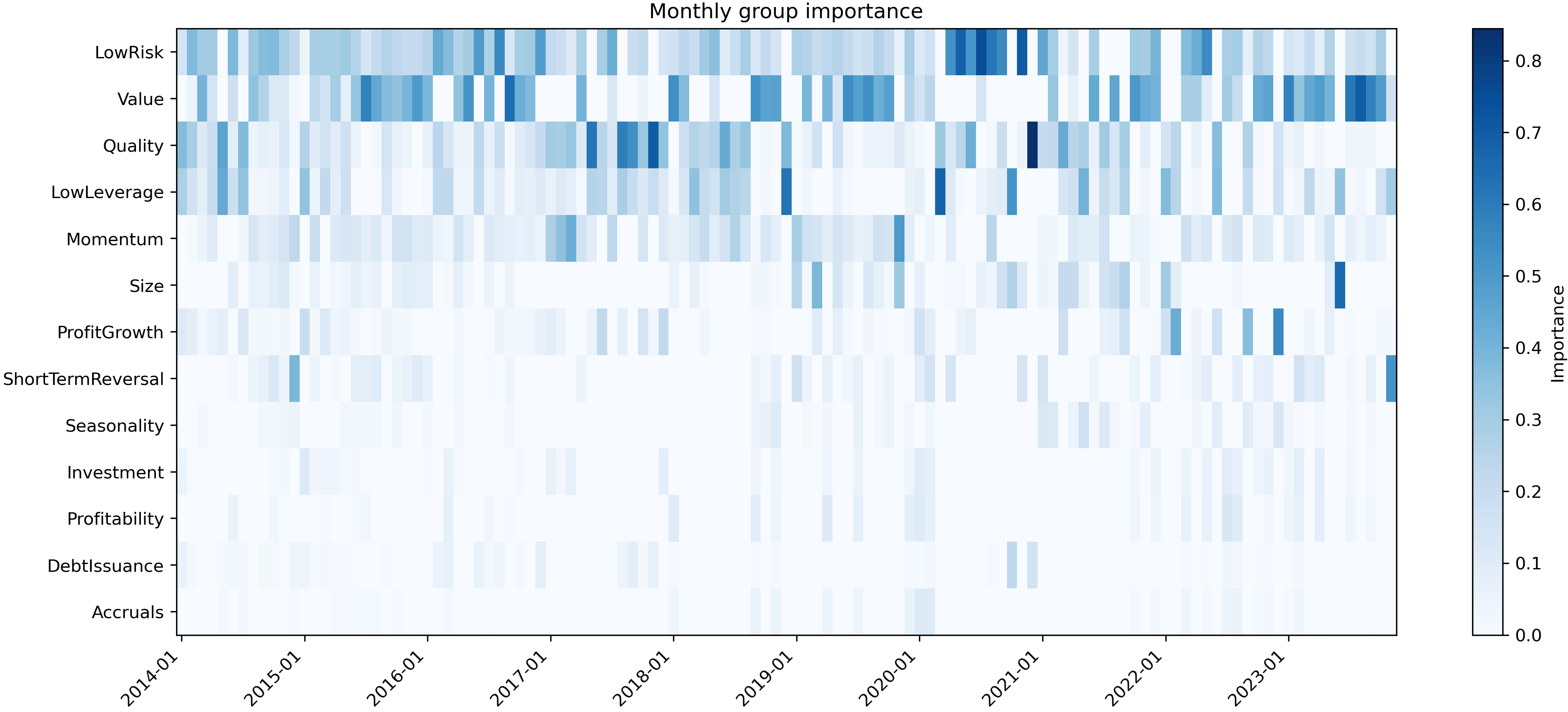}
    \caption{Group importance from ReSGA across months during the out-of-sample period.}
    \label{fig:monthly_group_importance}
\end{figure}

Moreover, \cref{fig:monthly_group_importance} traces the month-by-month evolution of group importance for ReSGA, where we compute group importance using the increase of loss from each month. Its heatmap reinforces that Low Risk remains important almost throughout the entire out-of-sample period, consistent with its role documented from \cref{fig:overall_group_importance}. As one may expect, the importance of Low Risk was especially pronounced from April to September 2020, since the COVID-19 shock had severely disrupted global financial markets. In contrast, Value and Quality display a visible ``complementary'' pattern over time, where the elevated period of Quality importance often coincides with muted Value, and vice versa. This finding suggests that the dominant predictors of tail risk shift between valuation-driven repricing forces and balance-sheet resilience/financial strength. Additionally, Quality and Low Leverage exhibit similar dynamics in certain months, a similarity that is expected due to their overlapping economic content related to profitability, safety, and leverage constraints.

\section{Transfer Learning}\label{sec:transfer}

Fundamentally, our universal models are driven by a principle that tail risk is governed by a mapping structure from asset characteristics to risk scores. As long as relevant asset characteristics are available, the model, by construction, can be applied to different assets without re-estimation. This contrasts sharply with econometric models, such as GARCH and GAS, which are inherently asset-specific. Hence, it raises a natural question: Can a universal model, trained on the largest and most information-rich US equity market, generalize to other equity markets without re-estimation?

To address this question, we conduct a transfer learning exercise. To be specific, we train each universal model exclusively on US equity dataset as in Section \ref{sec:empirical} and then apply it, without any re-training or fine-tuning, to five major international equity markets: China, Japan, UK, Australia, and Canada. This exercise allows us to assess whether the universal models capture fundamental relationships between firm characteristics and future tail risk across different institutional, regulatory, and liquidity environments.

\cref{tab:oos_loss_across_countries} reports the out-of-sample average loss across universal models and international equity markets. From this table, we have the following noteworthy findings: 

\begin{itemize}
    \item Despite substantial cross-market heterogeneity, ReSGA delivers the lowest out-of-sample loss in Japan, UK, and Australia, and remains highly competitive in China (ranked fourth) and Canada (ranked second). This strong and stable performance indicates that the predictive structure learned by ReSGA from US data generalizes well across international equity markets, even without re-estimation.
    
    \item In contrast, SGA performs poorly in most non-US markets, except for Canada. In particular, it is often dominated by simpler temporal models. This pattern suggests that cross-sectional dependence learned in the source market does not necessarily transfer well to other markets, potentially diluting asset-level tail-risk signals and introducing noise. The relatively strong performance of SGA in Canada is plausibly attributable to the close economic integration and high similarity between the US and Canadian equity markets, which makes cross-sectional structures learned from US data more transferable in this case. Note that ReSGA mitigates the above limitation of SGA by augmenting cross-sectional grouping with a retrieval mechanism that extends the effective temporal memory. The ability to exploit long historical information appears to compensate for potential misspecification in the cross-sectional structure, resulting in more robust transfer performance across markets.
    
    \item Temporal models, such as GRU, LSTM, and DLinear, also exhibit relatively stable out-of-sample performance across markets and often rank among the top predictors. This robustness further supports the view that temporal dynamics constitute a more transferable source of predictive power than cross-sectional dependence when models are deployed across markets without re-estimation.
\end{itemize}

\begin{table}[!h]
\centering
\caption{Out-of-sample average loss across different universal models and equity markets.}
\label{tab:oos_loss_across_countries}
\begin{threeparttable}
\setlength{\tabcolsep}{21pt}
\begin{tabular}{lccccc}
\toprule
& China 
& Japan 
& UK 
& Australia
& Canada \\
\midrule

\multicolumn{6}{l}{\textit{Point-wise Models}} \\[2pt]
\quad Linear    & $3.3011$ & $3.0623$ & $3.4380$ & $3.7542$ & $3.5735$ \\
\quad NN        & $3.3285$ & $3.0274$ & $3.4260$ & $3.6783$ & $3.4846$ \\[3pt]

\midrule
\multicolumn{6}{l}{\textit{Temporal Models}} \\[2pt]
\quad LANN      & $3.2924$ & $2.9953$ & $3.4582$ & $3.7393$ & $3.5002$ \\
\quad DLinear   & $3.2652$ & $2.9796$ & $3.4470$ & $3.7843$ & $3.5455$ \\
\quad LSTM      & $3.2677$ & $2.9521$ & $3.4262$ & $3.6733$ & $3.5375$\\
\quad GRU       & $\pmb{3.2547}$   & $2.9408$ & $3.4352$ & $3.6962$ & $3.5364$ \\
\quad Informer  & $3.2943$ & $2.9587$ & $3.4267$ & $3.7586$ & $3.5118$ \\
\quad EInformer & $3.3146$ & $2.9810$ & $3.4680$ & $3.8530$ & $3.5785$ \\
\quad DInformer & $3.2661$ & $2.9799$ & $3.4394$ & $3.7640$ & $3.5347$ \\[3pt]

\midrule
\multicolumn{6}{l}{\textit{Spatial-temporal Models}} \\[2pt]
\quad SGA       & $3.6346$ & $3.1192$ & $3.5729$ & $3.8702$  & $\pmb{3.4071}$\\
\quad ReSGA     & $3.2666$ & $\pmb{2.8884}$ & $\pmb{3.3992}$ & $\pmb{3.6333}$ & $3.4100$\\[3pt]

\bottomrule
\end{tabular}

\vspace{3pt}
\footnotesize
\textit{Notes.} 
This table reports out-of-sample average loss, which is obtained by applying each universal model trained exclusively on the US equity market to other five global markets. Lower values indicate better tail risk forecasting performance, with the lowest value for each market being highlighted in boldface. 
\end{threeparttable}
\end{table}

Besides the loss analysis, \cref{tab:mcs_modelwise} further displays the models in the final model set from the MCS test for the transfer learning across markets. For reference, we also include the earlier results for the US market, where only SGA and ReSGA remain in the final model set. The cross-market evidence from \cref{tab:mcs_modelwise} delivers a clear message: ReSGA is the only model selected in all final model sets. This result indicates that the advantage of ReSGA is not confined to the US market, but generalizes robustly across international equity universes. Moreover, the CC and AESR testing results in \cref{tab:transfer_var_es} provide further evidence of the effectiveness of ReSGA in transfer learning: Across the five target markets, ReSGA delivers consistently strong pass rates for both VaR and ES forecasts and ranks among the top performers in nearly all cases. 

\begin{table}[!ht]
\centering
\caption{Selected models in the final model set constructed by the MCS test across markets.}
\label{tab:mcs_modelwise}
\begin{threeparttable}
\setlength{\tabcolsep}{20pt}
\begin{tabular}{lcccccc}
\toprule
Model
& US
& China
& Japan
& UK
& Australia
& Canada \\
\midrule
Linear              & --         & \checkmark & --         & --               & --          & -- \\
NN                  & --         & --         & --         & --               & --          & -- \\
LANN                & --         & --         & --         & --               & --          & -- \\
DLinear             & --         & \checkmark & --         & --               & --          & -- \\
LSTM                & --         & \checkmark & --         & \checkmark       & --          & -- \\
GRU                 & --         & \checkmark & \checkmark & --               & --          & -- \\
Informer            & --         & --         & --         & --               & --          & -- \\
EInformer           & --         & --         & --         & --               & --          & -- \\
DInformer           & --         & \checkmark & --         & --               & --          & -- \\
SGA                 & \checkmark & --         & --         & --               & --          & \checkmark  \\
ReSGA               & \checkmark & \checkmark & \checkmark & \checkmark       &  \checkmark & \checkmark \\
\bottomrule
\end{tabular}

\vspace{4pt}
\footnotesize
\textit{Notes.}
A model with check mark ($\checkmark$) remains in the final model set, which is constructed by the MCS test at the $90\%$ confidence level for the corresponding market.
\end{threeparttable}
\end{table}

\begin{table}[!ht]
\centering
\caption{Pass rates of stocks with valid VaR (or ES) predictions across different models and markets.}
\label{tab:transfer_var_es}
\begin{threeparttable}
\setlength{\tabcolsep}{5pt}
\begin{tabular}{lcccccclccccc}
\toprule
 & \multicolumn{5}{c}{VaR} & & \multicolumn{5}{c}{ES} \\
 \cmidrule{2-6} \cmidrule{8-12}
 & China & Japan & UK & Australia & Canada & & China & Japan & UK & Australia & Canada \\
\midrule
\multicolumn{6}{l}{\textit{Point-wise Models}} & \multicolumn{6}{l}{}\\[2pt]

\quad Linear 
& 92.65 & 61.40 & 81.55 & 48.85 & 55.55 
& & 75.37 & 20.88 & 82.71 & 91.90 & 85.51 \\

\quad NN
& 94.90 & 67.98 & 86.65 & 63.98 & 69.52 
& & 50.59 & 23.74 & 66.25 & 78.03 & 74.80 \\

\addlinespace[3pt]
\midrule
\multicolumn{6}{l}{\textit{Temporal Models}} & \multicolumn{6}{l}{}\\[2pt]

\quad LANN
& 95.60 & 75.44 & 84.46 & 62.11 & 71.41 
& & 37.31 & 24.68 & 58.78 & 66.67 & 66.62 \\

\quad DLinear 
& 93.77 & 71.68 & 84.17 & 57.56 & 67.80 
& & 81.42 & 40.69 & 85.77 & 89.25 & 82.51 \\

\quad LSTM 
& 95.38 & 85.22 & 89.20 & 75.07 & 74.01 
& & 75.36 & 54.28 & 85.58 & 89.25 & 77.04 \\

\quad GRU 
& 95.38 & 87.35 & 85.87 & 67.65 & 72.71 
& & 77.19 & 72.61 & 88.72 & 91.18 & 81.71 \\

\quad Informer 
& 93.55 & 77.46 & 86.56 & 61.52 & 70.11 
& & 82.47 & 48.53 & 88.17 & 91.27 & 83.44 \\

\quad EInformer 
& 94.13 & 73.34 & 81.78 & 51.94 & 65.34 
& & 79.52 & 41.01 & 87.52 & 88.08 & 85.45 \\

\quad DInformer 
& 93.16 & 70.27 & 85.49 & 58.03 & 68.59 
& & \pmb{82.98} & 39.21 & 88.45 & 91.10 & 84.38 \\

\addlinespace[3pt]
\midrule
\multicolumn{6}{l}{\textit{Spatial-temporal Models}} & \multicolumn{6}{l}{}\\[2pt]

\quad SGA 
& 51.65 & 69.26 & 75.42 & 50.61 & \pmb{88.59} 
& & 75.98 & \pmb{87.62} & \pmb{91.40} & 90.85 & 86.92 \\

\quad ReSGA
& \pmb{95.62} & \pmb{89.97} & \pmb{91.63} & \pmb{75.42} & 88.45 
& & 67.39 & 70.55 & 87.99 & \pmb{95.05} & \pmb{89.72} \\

\addlinespace[3pt]
\bottomrule
\end{tabular}

\vspace{4pt}
\footnotesize
\textit{Notes.} This table reports pass rate of stocks in each market with valid VaR (or ES) predictions, assessed using the CC (or AESR) test at the significance level $\alpha=0.05$. For each column, the largest pass rate is highlighted in boldface.
\end{threeparttable}
\end{table}

Lastly, we examine the economic gains of ReSGA in the transfer learning exercise. \cref{tab:resga_country_portfolios} reports cross-market portfolio performance based on the ES-driven trading signal defined in \eqref{Cap_ES}. Its results show that portfolio return patterns differ substantially between US and non-US markets. In China and Japan, the decile portfolios exhibit a reversed pattern, with higher predicted signal associated with higher returns, leading to negative H--L portfolio performance. In the UK, Australia, and Canada, the results are more heterogeneous. Although monotonicity is weaker, the extreme deciles display substantial return dispersion, especially in the lowest-signal portfolios, which drive large negative H--L returns. These findings suggest that the pricing of tail risk varies across markets, likely reflecting differences in market structure, investor behavior, and risk premia.

Taken together, the transfer learning results reinforce our central conclusions: ReSGA functions as a general engine for tail risk forecasting.

\begin{table}[!ht]
\centering
\caption{Cross-market decile portfolio performance from ReSGA.}
\label{tab:resga_country_portfolios}
\begin{threeparttable}
\setlength{\tabcolsep}{4.5pt}
\renewcommand{\arraystretch}{1.0}

\begin{tabular}{llccccccccccc}
\toprule
Market & Metric & P1 & P2 & P3 & P4 & P5 & P6 & P7 & P8 & P9 & P10 & H--L \\
\midrule
\multirow{2}{*}{US} & Avg & 0.905 & 0.806 & 0.868 & 0.819 & 0.697 & 0.702 & 0.552 & 0.336 & -0.069 & -1.308 & 2.213 \\
 & SR & 0.710 & 0.613 & 0.585 & 0.510 & 0.401 & 0.391 & 0.291 & 0.153 & -0.026 & -0.392 & 0.787 \\
\addlinespace[2pt]
\multirow{2}{*}{China} & Avg & 0.431 & 0.526 & 0.576 & 0.607 & 0.729 & 0.769 & 1.022 & 1.225 & 1.360 & 1.848 & -1.417 \\
 & SR & 0.216 & 0.250 & 0.265 & 0.264 & 0.296 & 0.315 & 0.401 & 0.465 & 0.523 & 0.654 & -0.619 \\
\addlinespace[2pt]
\multirow{2}{*}{Japan} & Avg & 0.451 & 0.380 & 0.527 & 0.527 & 0.615 & 0.525 & 0.563 & 0.644 & 0.620 & 1.146 & -0.696 \\
 & SR & 0.403 & 0.358 & 0.478 & 0.468 & 0.525 & 0.443 & 0.438 & 0.462 & 0.384 & 0.610 & -0.543 \\
\addlinespace[2pt]
\multirow{2}{*}{UK} & Avg & 0.240 & 0.325 & 0.367 & 0.328 & 0.046 & -0.005 & -0.296 & -0.568 & -0.196 & 7.021 & -6.781 \\
 & SR & 0.175 & 0.199 & 0.221 & 0.204 & 0.029 & -0.003 & -0.171 & -0.292 & -0.081 & 0.602 & -0.586 \\
\addlinespace[2pt]
\multirow{2}{*}{Australia} & Avg & 0.545 & 0.528 & 0.501 & 0.299 & 0.210 & 0.537 & 0.531 & 1.541 & 1.676 & 5.324 & -4.778 \\
 & SR & 0.319 & 0.295 & 0.265 & 0.160 & 0.096 & 0.227 & 0.196 & 0.454 & 0.524 & 1.406 & -1.553 \\
\addlinespace[2pt]
\multirow{2}{*}{Canada} & Avg & 0.547 & 0.531 & 0.610 & 0.749 & 0.564 & 0.690 & 0.654 & 0.569 & 2.135 & 3.795 & -3.248 \\
 & SR & 0.367 & 0.388 & 0.382 & 0.405 & 0.299 & 0.329 & 0.312 & 0.248 & 0.652 & 1.111 & -1.191 \\
\bottomrule
\end{tabular}

\vspace{4pt}
\begin{minipage}{\linewidth}
\footnotesize
\textit{Notes.} This table reports the performance of value-weighted decile portfolios (across markets), constructed using $\alpha_{i,t}$ in \eqref{Cap_ES} as sorting signal, where $\widehat{\ES}_{i,t}$ is computed from ReSGA. Here, ReSGA is only trained on the US data and applied to other markets without re-estimation. 
At each month, stocks in every market are sorted in descending order of $\alpha_{i,t}$ and assigned to ten portfolios (P1--P10). Other descriptions are consistent with those in Table \ref{tab:portfolios_VaR_ES}.
\end{minipage}
\end{threeparttable}
\end{table}

\section{Conclusion}\label{sec:conclusion}

This paper studies tail risk forecasting, with a particular focus on VaR and ES, in the era of big data. Methodologically, we develop a unified joint VaR-ES learning framework supervised by the FZ loss. Based on this framework, we propose a large tail risk model, ReSGA, which captures the nonlinear relationships between asset characteristics and VaR-ES by accounting for spatial and temporal dependencies among assets. Empirically, using nearly a century of US equity data covering over 40,000 stocks and 153 firm characteristics, we find that ReSGA consistently delivers the best out-of-sample forecasting performance among a broad set of competing models. These forecasting gains are economically meaningful: Trading strategies based on ReSGA forecasts exhibit pronounced left-tail momentum and deliver superior long-short decile portfolio performances using a newly proposed size-enhanced left-side momentum signal.
 
Moreover, we provide systematic evidence on the scaling behavior of ReSGA and other machine-learning-based universal models. Our results offer little support for a general virtue of model complexity measured by parameter size alone. Instead, they favor the virtue of data complexity in learning tail risk, as improvements in forecasting VaR and ES can be achieved through larger in-sample datasets or more informative model inputs. In other words, the success of ReSGA is not only due to its access to rich available information but also to its ability to effectively extract, share, and utilize this information. Finally, we conduct group-importance analysis and transfer-learning experiments to illustrate the interpretability and cross-market generalizability of ReSGA. 


\bibliographystyle{imsart-nameyear}
\bibliography{Ref}

\appendix

\renewcommand{\thetable}{\thesection\arabic{table}}
\renewcommand{\thefigure}{\thesection\arabic{figure}}
\renewcommand{\theequation}{\thesection\arabic{equation}}
\renewcommand{\thealgorithm}{\thesection\arabic{algorithm}}
\renewcommand{\thedefinition}{\thesection\arabic{definition}}

\setcounter{section}{0}
\setcounter{equation}{0}
\setcounter{figure}{0}

\section{The Architecture of ReSGA}\label{sec:resga_arch}

\subsection{Encoder}\label{sec:resga_enc}

The encoder of ReSGA first applies the widely used recurrent neural network, GRU \citep{GRU}, to extract the temporal feature $\mathcal{H}_{t}$. The GRU is designed to capture temporal dynamics among asset characteristics through its recurrent hidden states. Let $\underline{\bx}_{i, t-1}^s = \bx_{i, t-1-S+s}$ denote the input characteristic vector of asset $i$ at time lag $s$. Then, the updates within the GRU are formulated as
\begin{align}\label{gru}
\begin{split}
    \text{(update gate)} \quad \bz_{i,t}^s &= \mathrm{sigmoid}\left(\bW_{z} \underline{\bx}_{i,t-1}^s + \bU_{z} \bh_{i,t}^{s-1} + \bb_{z}\right), \\
    \text{(reset gate)} \quad \br_{i,t}^s &= \mathrm{sigmoid}\left(\bW_{r} \underline{\bx}_{i,t-1}^s + \bU_{r} \bh_{i,t}^{s-1} + \bb_{r}\right), \\
    \text{(candidate hidden state)} \quad \tilde{\bh}_{i,t}^s &= \tanh \left(\bW_{h} \underline{\bx}_{i,t-1}^s + \bU_{h}\left(\br_{i,t}^s \odot \bh_{i,t}^{s-1}\right) + \bb_{h} \right), \\
    \text{(hidden state)} \quad \bh_{i,t}^s &= \bz_{i,t}^s \odot \bh_{i,t}^{\,s-1} + \left(1 - \bz_{i,t}^s\right) \odot \tilde{\bh}_{i,t}^s,
\end{split}
\end{align}
where $\bW_* \in \mathbb{R}^{D \times P}$, $\bU_* \in \mathbb{R}^{D \times D}$, and $\bb_* \in \mathbb{R}^D$ for $* \in \{z,r,h\}$ are weight matrices and bias vectors shared across time lags. The hidden-state dimension $D$ is a tunable hyperparameter that controls model complexity, and $\odot$ denotes element-wise multiplication. The element-wise activation functions $\mathrm{sigmoid}$ and $\tanh$ are defined as $\mathrm{sigmoid}(x) = \frac{1}{1 + e^{-x}}$ and $\tanh(x) = \frac{e^x - e^{-x}}{e^x + e^{-x}}$ for $x\in \bbR$.

In \eqref{gru}, the hidden state $\bh_{i,t}^{s} \in \mathbb{R}^D$ serves as the key feature, summarizing the accumulated temporal information up to lag $s$, while $\tilde{\bh}_{i,t}^s$ represents the candidate hidden state, used to store newly inputted information $\bx_{i, t-1}^s$ at lag $s$. The update gate $\bz_{i,t}^{s} \in (0,1)^D$ controls how much of the past information from $\bh_{i,t}^{s - 1}$ should be kept, whereas the reset gate $\br_{i,t}^{s} \in (0,1)^D$ determines the degree to which the previous hidden state $\bh_{i,t}^{s - 1}$ contributes to the candidate state $\tilde{\bh}_{i,t}^{s} \in (-1,1)^D$. Together, these gating mechanisms enable the GRU to adaptively balance memory retention and new information integration, effectively capturing complex temporal dependencies in an autoregressive manner.

For each asset $i = 1, \dots, N_t$ and lag $s = 1, \dots, S$, we obtain the hidden states $\{\bh_{i,t}^1, \dots, \bh_{i,t}^S\}$ from the GRU, and then we stack them into the temporal feature across all assets and lags to construct
\begin{align}\label{temporal_feature}
    \mathcal{H}_t = [\bH_{1,t}, \dots, \bH_{N_t,t}] \in \mathbb{R}^{N_t \times S \times D} \,\,\, \text{with} \,\,\, \bH_{i,t} = [\bh_{i,t}^1, \dots, \bh_{i,t}^S] \in \mathbb{R}^{S \times D}.
\end{align}

In addition to $\mathcal{H}_t$, the encoder of ReSGA uses the most recent hidden states $\{\bh_{1,t}^S, \dots, \bh_{N_t,t}^S\}$ to infer the latent group structure among assets. Specifically, it constructs a set $\mathbb{M}_t$ that captures inter-asset relationships based on the similarity of temporal features. To fulfill this, we first treat each asset as a potential group center, with its node feature defined as
\begin{align*}
    \bu_{k,t} = \bh_{k,t}^S, \,\,\, \text{for} \,\,\, k = 1, \dots, N_t.
\end{align*}
Then, we measure the similarity between asset $i$ and center $k$ by the cosine similarity:
\begin{align*}
    \gamma_{i,k,t} = \text{Cosine} \left( \bu_{k,t}, \bh_{i,t}^S\right)
    = \frac{\bu_{k,t}' \bh_{i,t}^S}{\|\bu_{k,t}\| \, \|\bh_{i,t}^S\|}, 
    \,\,\, \text{for} \,\,\, i, k = 1, \dots, N_t,
\end{align*}
where $\|\cdot\|$ denotes the $L_2$ norm. To mitigate the impact of weak or noisy similarities, we retain only the $K^\ast$ strongest connections for each asset $i$:
\begin{align}\label{topK}
    \gamma^\ast_{i,k,t} =
    \begin{cases}
        1, & \text{if } k = i, \\
        \gamma_{i,k,t}, & \text{if } |\gamma_{i,k,t}| > \text{Top}(\mathbb{A}_{i,t}, K^\ast), \\
        0, & \text{otherwise},
    \end{cases}
\end{align}
where $K^\ast$ is a hyperparameter controlling the number of retained connections, $\mathbb{A}_{i,t} = \{|\gamma_{i,k,t}| : k \neq i\}$, and $\text{Top}(\mathbb{A}, b)$ denotes the $b$-th largest value in the set $\mathbb{A}$.

After thresholding, we form each group $\mathbb{M}_{k,t}= \{ i : \gamma^\ast_{i,k,t} \neq 0 \}$ by gathering all assets connected to center $k$.
Meanwhile, we discard those groups containing only their own center, as single-member groups carry no sharing information. Consequently, we 
reach the group structure:
\begin{align}\label{group_structure}
    \mathbb{M}_t = \{\mathbb{M}_{k,t} : k = k_1, \dots, k_{K_t} \},
\end{align}
where $k_1, \dots, k_{K_t}$ denote the indices of all identified group centers.

Our data-driven way to discover group structure $\mathbb{M}_{t}$ matches many empirical practices. In financial markets, assets such as stocks naturally form groups based on their shared characteristics and economic linkages. Typically, stocks within the same group $\mathbb{M}_{k,t}$ tend to co-move due to common factors such as industrial background, financial statement, or ownership background. For instance, technology companies like NVIDIA, Apple, Microsoft, and AMD tend to cluster into the same group, as their prices are jointly affected by factors such as semiconductor supply chains, innovation cycles, and market sentiment toward the tech sector. These co-movements underscore the importance of modeling group-specific relationships to capture heterogeneity across different groups. For more empirical elaborations on the group structures in stock return forecasting, one can refer to \citet{Zhu2024Graph} and references therein. 

Combining \eqref{temporal_feature} and \eqref{group_structure}, we can equivalently write the workflow of encoder in a brief form
\begin{align}\label{resga_encoder}
    (\mathcal{H}_t, \mathbb{M}_t) = \Enc(\mathcal{X}_{t-1}; \bphi),
\end{align}
where $\bphi$ includes all trainable parameters in \eqref{gru}. 

\subsection{Retriever}\label{sec:resga_ret}

Although the temporal feature $\bh_{i,t}^s$ in \eqref{gru} summarizes the accumulated temporal information up to lag $s$ (i.e. time $t-S+s$), it inevitably ``forgets'' long-term dependencies due to the effects of the update and reset gates. Moreover, our encoder only extracts the temporal feature for each asset independently, without incorporating useful temporal information from other related assets. For instance, during the 2022 technology sell-off, the price movement of NVIDIA exhibited a pattern strikingly similar to that of Apple during the 2008 financial crisis: Both experienced rapid growth followed by sharp corrections driven by macroeconomic tightening. Retrieving such long-ago but structurally similar patterns across assets can provide valuable insights that purely temporal models may overlook. 

To address these limitations, we introduce the retriever of ReSGA, which captures structurally similar patterns from the historical trajectories of related assets. Specifically, we partition the $S$ lags into $L^\ast$ non-overlapping segments, forming
\begin{align*}
    \bH_{i,t} = [\bh_{i,t}^1, \dots, \bh_{i,t}^{S/L^\ast}, \dots, \bh_{i,t}^{2S/L^\ast}, \dots, \bh_{i,t}^{(L^\ast-1)S/L^\ast}, \dots, \bh_{i,t}^{S}],
\end{align*}
where $\bh_{i,t}^{lS/L^\ast}$ for $l = 1, \dots, L^\ast$ represents the historical context up to time $lS/L^\ast$, and $L^\ast$ is a user-specific hyperparameter satisfying that $S$ is divisible by $L^\ast$. The temporal features from the $L^\ast$ segments thus act as reference checkpoints, each summarizing distinct stages of past information for subsequent retrieval and comparison. 

Next, we retrieve information from its related assets defined by the group structure $\mathbb{M}_t$. For each asset $i$, the related assets are collected into a set
\begin{align}\label{center_set}
    \mathbb{N}_{i,t} = \{k : i \in \mathbb{M}_{k,t}\},
\end{align}
where $\mathbb{M}_{k,t}$ is the group centered at asset $k$. By construction, $\mathbb{N}_{i,t}$ contains the $K^{\ast}$ most similar assets to $i$ identified by the encoder, indicating which assets’ historical trajectories should be searched in the retrieval stage.

We now define the retrieval feature for asset $i$ at time $t$ by aggregating information from a set of related assets. For each related asset $j \in \mathbb{N}_{i,t}$, its contribution to the retrieval feature of asset $i$ is given by
\begin{align}\label{related_R}
    \bz_{i,j,t} = \sum_{l = 1}^{L^\ast - 1} \alpha_{i,j,l,t}\, \bh_{j,t}^{(lS/L^\ast) + 1} \in \mathbb{R}^{D},
\end{align}
where $\alpha_{i,j,l,t}$ is the normalized attention weight defined as
\begin{align}\label{attn_weight}
    \alpha_{i,j,l,t} = 
    \frac{\exp(\beta_{i,j,l,t})}{
    \sum_{\ell = 1}^{L^\ast - 1} \exp(\beta_{i,j,\ell,t})},
    \,\,\,
    \text{with} \,\,\,
    \beta_{i,j,l,t} = \mathrm{Cosine}\!\left(\bh_{i,t}^S,\, \bh_{j,t}^{lS/L^\ast}\right).
\end{align}

Here, the attention weight $\alpha_{i,j,l,t}$ quantifies the similarity between the current temporal feature of asset $i$, $\bh_{i,t}^S$, and the checkpoint feature of asset $j$, $\bh_{j,t}^{lS/L^\ast}$. This similarity measures how much the present feature of asset $i$ resembles the past feature of asset $j$ at segment $l$. Importantly, while the attention weights are based on these checkpoint features, $\bz_{i,j,t}$ uses the subsequent feature $\bh_{j,t}^{(lS/L^\ast) + 1}$ that immediately follows each checkpoint\footnote{This is why only the first $L^\ast-1$ checkpoints are used, as $\bh_{j,t}^{S + 1}$ is not available at time $t$.}. The rationale is that if asset $i$’s current feature mirrors asset $j$’s historical feature at time $lS/L^\ast$, then asset $j$’s next state after that point provides valuable predictive information for asset $i$’s future evolution. In this way, the retriever aligns historical features across assets and integrates their forward states, allowing the model to infer potential future behavior from similar past behaviors.

Subsequently, we aggregate the contributions from all related assets to form the retrieval feature.  
To fulfill this, we compute the average similarity between each related asset $j$ and the target asset $i$ across all checkpoints:
\begin{align*}
    \bar{\beta}_{i,j,t} = \frac{1}{L^\ast-1}\sum_{l=1}^{L^\ast - 1} \beta_{i,j,l,t},
\end{align*}
which captures the overall relevance of asset $j$ to asset $i$.
Then, following \eqref{related_R} and \eqref{attn_weight}, we calculate the relative contribution of each related asset and aggregate it into the retrieval feature for each asset $i$:
\begin{align*}
    \bz_{i,t} =  \sum_{j\in\mathbb{N}_{i,t}} \omega_{i,j,t} \bz_{i,j,t} \in \mathbb{R}^{D}, \,\,\, \text{with} \,\,\, \omega_{i,j,t} = 
    \frac{\exp(\bar{\beta}_{i,j,t})}{
    \sum_{m \in \mathbb{N}_{i,t}}\exp(\bar{\beta}_{i,m,t})}.
\end{align*}

\begin{figure}[!ht]
    \centering
    \includegraphics[width=0.7\textwidth,height=12cm]{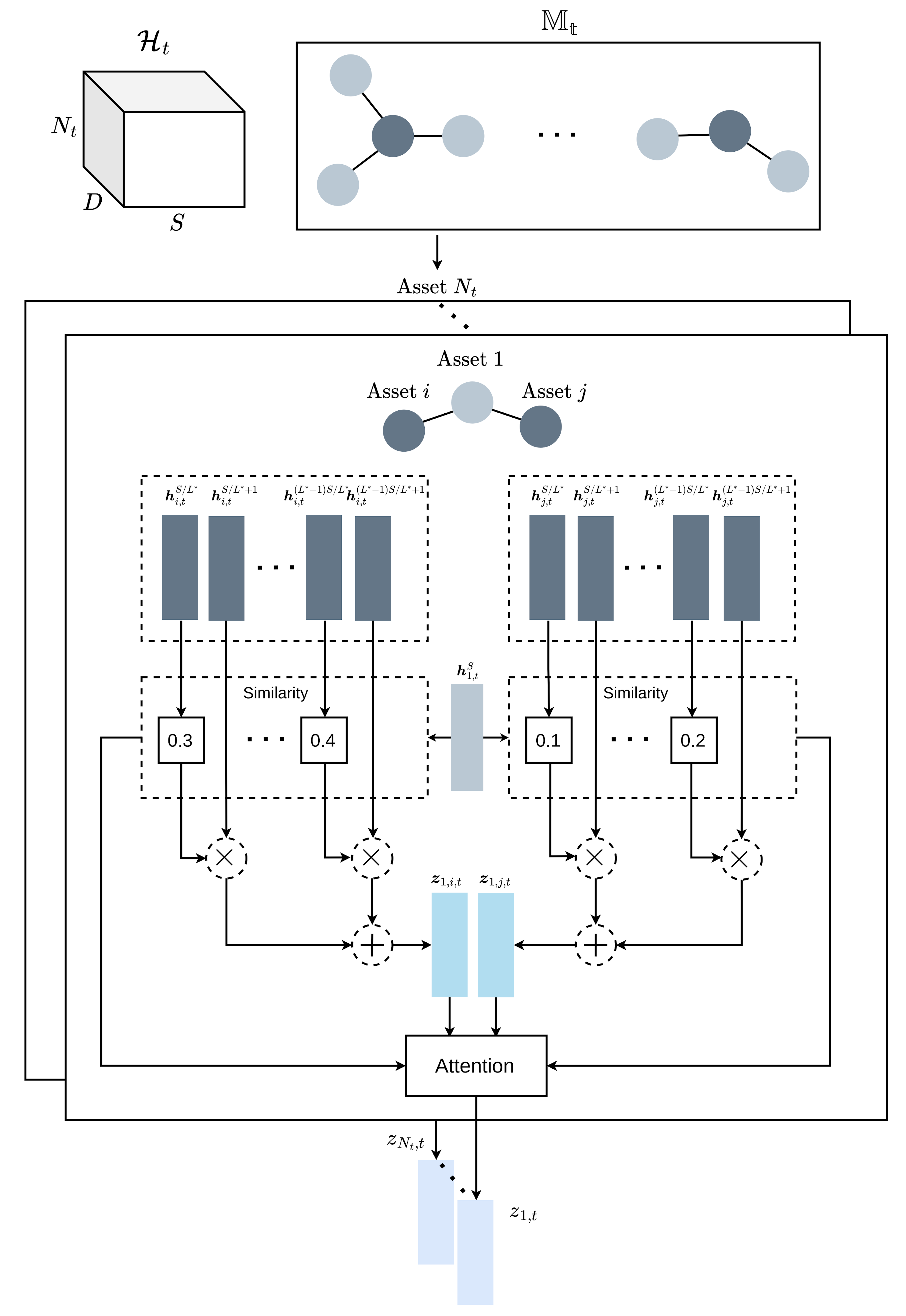}
    \caption{Retriever component of ReSGA.}
    \label{fig:retriever_module}
    \caption*{\footnotesize \textit{Notes.} This figure presents the retriever component used in ReSGA. The retriever operates conditionally on the group structure $\mathbb{M}_t$ and $\mathcal{H}_t$ obtained by the encoder, and compute the retrieval feature $\bZ_t$, which is further fed into the decoder.}
\end{figure}

Finally, we stack all $\bz_{i,t}$ across assets to obtain the retrieval feature:
\begin{align}\label{resga_retriever}
    \bZ_{t} = \Ret(\mathcal{H}_t; \mathbb{M}_t) = [\bz_{1,t}, \dots, \bz_{N_t,t}] \in \mathbb{R}^{N_t \times D}.
\end{align}
By design, $\bZ_{t}$ aggregates information through a two-stage attention mechanism. In the first stage, the model adaptively selects the most informative temporal segments from the history of each related asset, focusing on those that resemble the current behavior of asset $i$. In the second stage, it re-weights the related assets based on their overall relevance, emphasizing those whose past dynamics are most predictive for asset $i$. This retrieval structure allows the retriever to recall and integrate inter-asset temporal patterns that are similar to the present behavior of the target asset, thereby improving its ability to capture and forecast risk dynamics at time $t$. See \cref{fig:retriever_module} for the entire process of retriever.

\subsection{Decoder}\label{sec:resga_dec}

The decoder integrates the temporal features $\mathcal{H}_t$, group structure $\mathbb{M}_t$, and retrieval features $\bZ_t$ to forecast the risk scores $\by_t$. Specifically, it adopts a residual connecting framework (\citealp{Oreshkin2020NBEATSNB}), including four layers:  
(i) a group layer capturing shared group-level co-behavior,  
(ii) a retrieval layer utilizing the retrieval features,   
(iii) an individual layer modeling stock-specific behavior, and
(iv) a fully connected layer producing the final risk score forecasts.

\paragraph{Group Layer}

As the retriever captures spatial–temporal dependencies for each asset by linking its current temporal feature to the historical features of related assets, the group layer instead focuses on modeling contemporaneous interactions among assets within the same group. In other words, while the retriever aligns the present state of asset $i$ with the past trajectories of its peers to recover long-term temporal relations, the group layer exploits the most recent temporal features across assets to capture shared short-term co-movements.

Specifically, the group layer aims to capture common influences within each group based on the most recent temporal feature $\{\bh_{1, t}^S, \dots, \bh_{N_t,t}^S\}$ and group structure $\mathbb{M}_t$. For each group $\mathbb{M}_{k,t}$ in $\mathbb{M}_t$, we first design a group-level feature by aggregating member features:
\begin{align*}
    \bG_{k,t} = \mathrm{LeakyReLU}\!\left[\bW_1 \left(\sum_{i \in \mathbb{M}_{k,t}} \gamma_{i,k,t}^{*} \bh^S_{i,t} \right) + \bb_1 \right] \in \mathbb{R}^D, 
\end{align*}
where $\bh^S_{i,t}$ is the temporal feature of asset $i$ from \eqref{gru}, $\bW_1 \in \mathbb{R}^{D \times D}$ and $\bb_1 \in \mathbb{R}^D$ are learnable parameters, and $\mathrm{LeakyReLU}(\cdot)$ is an element-wise activation defined as 
\begin{align*}
    \mathrm{LeakyReLU}(x) =
    \begin{cases}
        x, & x > 0, \\
        0.01 x, & x \leq 0.
    \end{cases}
\end{align*}
This aggregation produces $\bG_{k,t}$ as a concise summary of the short-term shared dynamics within group $\mathbb{M}_{k,t}$.

Next, we propagate the group-level effects to individual assets through a graph attention mechanism \citep{velivckovic2018graph}:
\begin{align*}
    \bg_{i,t} = \sum_{k \in \mathbb{N}_{i,t}} \alpha^\ast_{i,k,t}\bG_{k,t}, 
    \,\,\, \text{with} \,\,\,
    \alpha^\ast_{i,k,t} = \frac{\exp(\beta^\ast_{i,k,t})}{
    \sum_{m \in \mathbb{N}_{i,t}}\exp(\beta^\ast_{i,m,t})},
    \,\,\, 
    \beta^\ast_{i,k,t} = \text{Cosine}(\bh^S_{i,t}, \bG_{k,t}),
\end{align*}
where $\mathbb{N}_{i,t}$ in \eqref{center_set} denotes the set of group centers containing asset $i$.  
Here, $\beta^\ast_{i,k,t}$ measures the similarity between the current temporal feature of asset $i$ and the group-level feature of group $k$, while $\alpha^\ast_{i,k,t}$ provides normalized attention weights to indicate the relative importance of each group to asset $i$. 

In the final step of the graph layer, we feed the group feature $\bg_{i,t}$ into a two-headed fully connected network to produce the forward and backward components:
\begin{align}\label{graph_forecast}
\begin{split}
    \widetilde{\bg}_{i,t} &= \mathrm{LeakyReLU}(\bW_2\bg_{i,t} + \bb_2), \\
    \bh^G_{i,t} &= \mathrm{LeakyReLU}(\bW_3\widetilde{\bg}_{i,t} + \bb_3), \\
    \bo^G_{i,t} &= \mathrm{LeakyReLU}(\bW_4\widetilde{\bg}_{i,t} + \bb_4),
\end{split}
\end{align}
where $\bW_2, \bW_3, \bW_4 \in \mathbb{R}^{D \times D}$ and $\bb_2, \bb_3, \bb_4 \in \mathbb{R}^D$ are learnable parameters. Here, $\bo^G_{i,t}$ captures the group-level information directly used for forecasting $\by_{i,t}$, while $\bh^G_{i,t}$ represents the portion of $\bh^S_{i,t}$ already explained by group co-behavior. This residual design separates the group effect from subsequent retrieval and individual layers, enhancing both model interpretability and learning efficiency \citep{Oreshkin2020NBEATSNB}.

\paragraph{Retrieval Layer} 

After the graph layer, the retrieval layer employs the retrieval feature $\bZ_{t}$ defined in \eqref{resga_retriever}, which captures long-term spatial–temporal dependencies through a two-stage attention mechanism.  
To be specific, this retrieval layer passes $\bZ_{t}$ through a two-headed fully connected network, mirroring the structure used in the group layer:
\begin{align}\label{ret_forecast}
\begin{split}
    \widetilde{\bz}_{i,t} &= \mathrm{LeakyReLU}(\bW_5\bz_{i,t} + \bb_5), \\
    \bh^R_{i,t} &= \mathrm{LeakyReLU}(\bW_6\widetilde{\bz}_{i,t} + \bb_6), \\
    \bo^R_{i,t} &= \mathrm{LeakyReLU}(\bW_7\widetilde{\bz}_{i,t} + \bb_7),
\end{split}
\end{align}
where $\bz_{i,t}$ is the retrieval feature of asset $i$ (i.e., $i$-th row of $\bZ_t$), and $\bW_5, \bW_6, \bW_7 \in \mathbb{R}^{D \times D}$ and $\bb_5, \bb_6, \bb_7 \in \mathbb{R}^D$ are learnable parameters.  
Here, $\bh^R_{i,t}$ serves as the backward component of retrieval layer, representing the portion of $\bh_{i,t}$ explained by retrieved information, while $\bo^R_{i,t}$ denotes the forward component of retrieval layer that contributes directly to forecasting $\by_{i,t}$.  
The parameter sets $(\bW_5,\bW_6,\bW_7)$ and $(\bb_5,\bb_6,\bb_7)$ play analogous roles to $(\bW_2,\bW_3,\bW_4)$ and $(\bb_2,\bb_3,\bb_4)$ in the group layer, respectively.

\paragraph{Individual Layer}

The individual layer models the idiosyncratic part of each asset after accounting for the effects already captured by the group and retrieval layers. Specifically, we define:
\begin{align}\label{ind_forecast}
    \bo^{I}_{i,t} = \mathrm{LeakyReLU}(\bW_8 \bh^I_{i,t} + \bb_8),
    \,\,\,\text{with} \,\,\,
    \bh^I_{i,t} = \bh^S_{i,t} - \bh^{G}_{i,t} - \bh^{R}_{i,t},
\end{align}
where $\bW_8 \in \mathbb{R}^{D \times D}$ and $\bb_8 \in \mathbb{R}^D$ are learnable parameters. The output (or the forward component of individual layer) $\bo^{I}_{i,t}$ captures asset-specific behavior unique to asset $i$, ensuring that the residual temporal information unexplained by group or retrieval effects is explicitly modeled.

\paragraph{Fully Connected Layer}

Finally, the decoder aggregates the outputs from the group, retrieval, and individual layers through additive composition to generate the final forecasts:
\begin{align}\label{resga_decoder}
    \bY_t = [\by_{1,t}, \dots, \by_{N_t,t}] = \Dec(\mathcal{H}_{t}, \bZ_t; \mathbb{M}_t, \bpsi) \in \mathbb{R}^{N_t \times 2},
\end{align}
where each individual forecast is given by
    $\by_{i,t} = \bW_9 \left(\bo^{G}_{i,t} + \bo^{R}_{i,t} + \bo^{I}_{i,t}\right) + \bb_9$,
with $\bW_9 \in \mathbb{R}^{2 \times D}$ and $\bb_9 \in \mathbb{R}^2$ as learnable parameters. The collection $\bpsi = \{(\bW_j, \bb_j)\}_{j=1}^{9}$ represents all trainable parameters within the decoder. 

Intuitively, this hierarchical design in (\ref{resga_decoder}) allows the decoder to jointly capture group-level co-behavior, long-term spatial-temporal dependencies, and idiosyncratic behavior within a unified learning framework.

\subsection{Summary}

\begin{figure}[!h]
    \centering
    \includegraphics[width=0.9\linewidth]{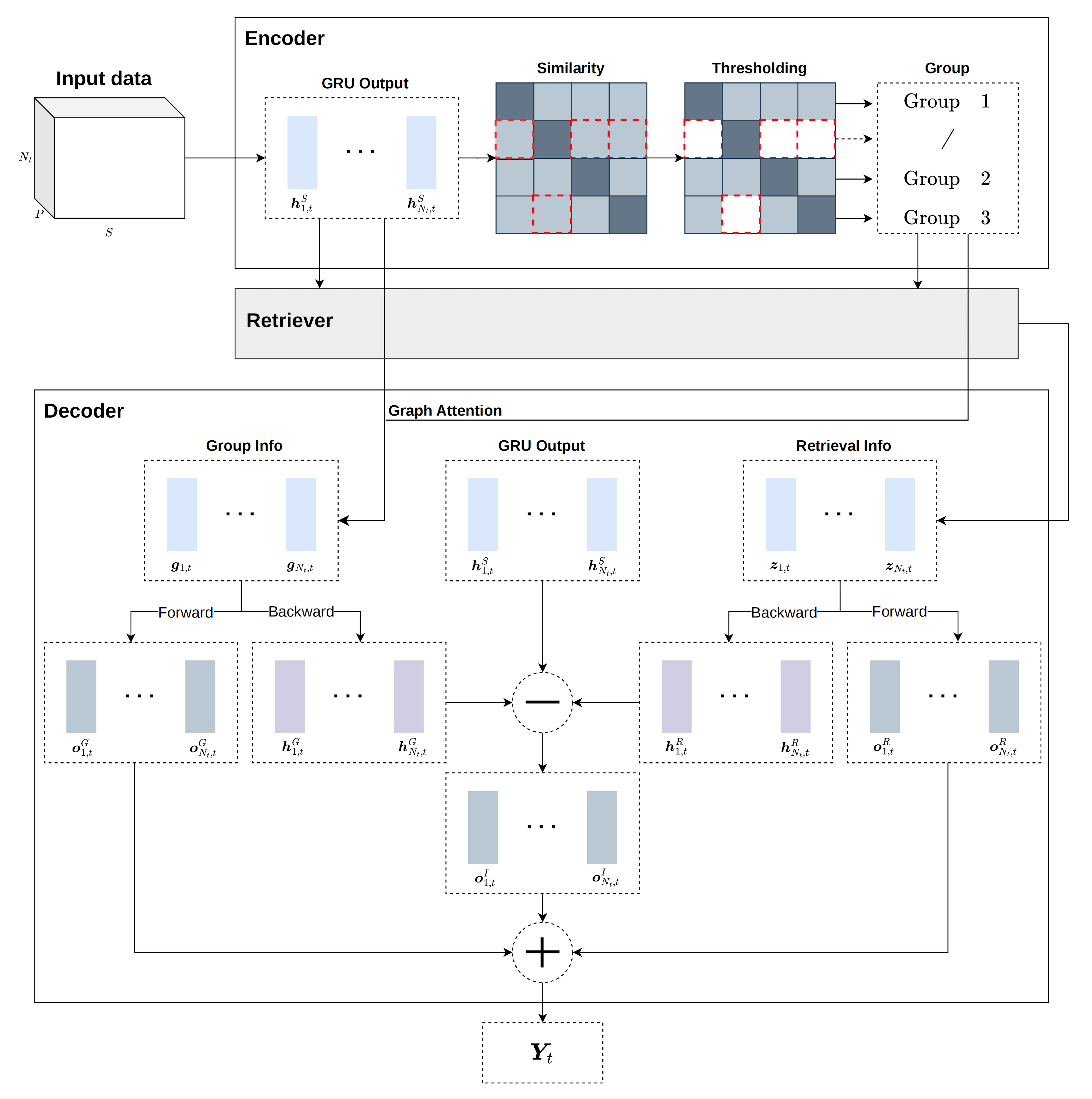}
    \caption{Overall architecture of ReSGA.}
    \label{fig:resga_arch}
    \caption*{\footnotesize \textit{Notes.} This figure summarizes the end-to-end workflow of ReSGA, consisting of an encoder constructing cross-sectional groups, a retriever modeling long-term temporal dependencies, and a decoder combining group, retrieval, and individual information via a residual learning framework to produce tail-risk forecasts. For illustration, we set $N_t = 4$ and $K^\ast = 2$. The internal mechanics of the Retriever block are detailed in Fig \ref{fig:retriever_module}.}
\end{figure}

Accompanied with $\Enc(\cdot; \bphi)$ in \eqref{resga_encoder}, $\Ret(\cdot; \mathbb{M}_t)$ in \eqref{resga_retriever}, and $\Dec(\cdot, \cdot; \mathbb{M}_t, \bpsi)$ in \eqref{resga_decoder}, we get the specification of ReSGA in \eqref{resga_structure}. See Fig \ref{fig:resga_arch} for its entire network architecture.

\section{Model Architectures} \label{sec:models}

\setcounter{equation}{0}

In this appendix, we present the details of all competing models considered in Section \ref{sec:competitor}.

\subsection{Point-Wise Models}\label{sec:point-wise}

We consider two point-wise models in (\ref{point-wise}): Linear and NN.

\subsubsection{Linear}\label{sec:linear_model}

The Linear model assumes $f$ in (\ref{point-wise}) satisfies
\begin{align}\label{linear}
    \by_{i,t} = f(\bx_{i,t - 1}; \btheta) = \balpha + \bbeta' \bx_{i,t - 1} ,
\end{align}
where $\balpha \in \mathbb{R}^{2}$ and $\bbeta \in \mathbb{R}^{P \times 2}$ are the vectors of intercepts and regression coefficients, respectively. Here, $\btheta$ includes all parameters in $\balpha$ and $\bbeta$. To mitigate the potential high-dimensional issue, we apply $\ell_1$-penalization on $\btheta$ during training,  promoting sparsity in the learned parameters.

\subsubsection{NN}\label{sec:fc}

Following \cite{gu2020empirical}, the NN model applies a multi-layer fully connected neural network as the function $f$ in (\ref{point-wise}):
\begin{align}\label{nn3_hidden}
    \bx^{(l)}_{i, t - 1} &= \mathrm{ReLU} \left (\bW^{(l)} \bx^{(l - 1)}_{i, t-1} + \bb^{(l)} \right) \in \mathbb{R}^{D^{(l)}}, \,\,\, \text{for}\,\,\, l = 1, \dots, L, \\
    \label{nn3_linear}
    \by_{i,t} & = f(\bx_{i,t-1}; \btheta) = \bW^{(L + 1)} \bx^{(L)}_{i, t-1} + \bb^{(L + 1)} \in \mathbb{R}^2,
\end{align}
where $\bx_{i,t-1}^{(l)}$ denotes the output from the $l$-th hidden layer,
with $\bx^{(0)}_{i,t-1}=\bx_{i,t-1}$, and $D^{(l)}$ is its dimension. Here, $L$ is the number of hidden layers, the activation function ReLU is defined as
\begin{align*}
    \mathrm{ReLU}(z) = \max(0, z), \,\,\, \text{for}\,\,\, z \in \mathbb{R},
\end{align*}
the weight matrix $\bW^{(l)} \in \mathbb{R}^{D^{(l)} \times D^{(l-1)}}$ and bias vector $\bb^{(l)} \in \mathbb{R}^{D^{(l)}}$, for $l=1, \dots, L+1$, are network parameters, with $D^{(0)} = P$ and $D^{(L+1)} = 2$, and  $\btheta$ contains all the parameters in $\{\bW^{(l)}, \bb^{(l)}: l=1, \dots, L+1\}$.

From \eqref{nn3_hidden}, we can see that the hidden layers recursively apply linear transformations and nonlinear activation to extract predictive features from the raw characteristics, while the final linear projection in \eqref{nn3_linear} is almost the same as the linear model in \eqref{linear}. In particular, the dimensionalities $\{D^{(l)}:l = 1, \dots, L\}$ control the ``width'' of NN, while the number of hidden layers $L$ controls the ``depth'' of NN. Needless to say, they are crucial for the complexity of NN (i.e., the dimensionality of $\btheta$).

According to the geometric pyramid rule proposed by \cite{masters1993practical}, an exponentially decaying sequence of network widths helps balance computational efficiency and nonlinear approximation capability. Specifically, this rule suggests 
\begin{align*}
    D^{(l)} = \left\lfloor \gamma^{l - 1} D \right\rfloor, \,\,\, \text{for}\,\,\, l = 1, \dots, L, 
\end{align*}
where $\lfloor \cdot \rfloor$ is the floor operator, and $\gamma$ is a user-defined hyperparameter. \cite{gu2020empirical} employ $L = 3$, $D = 32$, and $\gamma = 0.5$ for $P = 94$. Following this setup, we also fix $L = 3$ and $\gamma = 0.5$, but vary $D$ to investigate the virtue of model complexity in tail risk forecasting. Additionally, as in \cite{gu2020empirical}, we also apply the batch normalization technique \citep{ioffe2015batch} to control the variability of network inputs across mini-batches. For more technical details on fully connected neural networks, one can refer to \cite{goodfellow2016deep}.

\subsection{Temporal Models}\label{sec:temporal}

We consider seven temporal models in (\ref{temporal_model}): LANN, DLinear, LSTM, GRU, Informer, EInformer, and DInformer.

\subsubsection{LANN}

The LANN model extends the NN model in \cref{sec:fc} by incorporating temporal information through lagged firm characteristics. Architecturally, LANN is identical to the NN, and is defined as
\begin{align*}
    \bx^{(l)}_{i, t - 1} &= \mathrm{ReLU} \left (\bW^{(l)} \bx^{(l - 1)}_{i, t-1} + \bb^{(l)} \right) \in \mathbb{R}^{D^{(l)}}, \quad l = 1, \dots, L, \\
    \by_{i,t} & = f(\bx_{i,t-1}; \btheta) = \bW^{(L + 1)} \bx^{(L)}_{i, t-1} + \bb^{(L + 1)} \in \mathbb{R}^2,
\end{align*}
where the input $\bx^{(0)}_{i,t-1}=\vecc(\bX_{i,t-1})\in\mathbb{R}^{SP}$ flattens the matrix $\bX_{i,t-1}$ into a vector. As a result, the input dimension of the first layer increases from $D^{(0)} = P$ to $D^{(0)} = SP$, and the corresponding weight matrices and bias vectors adjust accordingly. All remaining network architecture and notation follow the NN model in \cref{sec:fc}.

Intuitively, the LANN model treats all lagged characteristics as ``static'' covariates. It does not model temporal dynamics explicitly and therefore ignores potential temporal dependence structures across lags.

\subsubsection{DLinear}

The DLinear model first decomposes the matrix $\bX_{i,t-1}$ into a trend component $\bT_{i,t-1} \in \mathbb{R}^{S \times P}$ and a seasonal component $\bS_{i,t-1} \in \mathbb{R}^{S \times P}$.
For the trend component, the $s$-th row of $\bT_{i,t-1}$ is defined as 
\begin{align}\label{trend_comp}
    \bT_{i,t-1}^s =
  \frac{1}{|\mathcal{N}_K(s)|}\sum_{j\in \mathcal{N}_K(s)} \underline{\bx}_{i,t-1}^j \in \mathbb{R}^P,\,\,\, \text{for}\,\,\, s = 1, \dots, S,
\end{align}
where $\mathcal{N}_K(s) = \left\{ j \in \{1,\dots,S\} : |j-s| \le \lfloor K/2 \rfloor \right\}$ is the local index set for index $s$, and $\underline{\bx}_{i,t-1}^j \equiv \bx_{i,t-1-S+j} \in \mathbb{R}^P$ denotes the $j$-th row of $\bX_{i,t-1}$, which also corresponds to the $j$-th lagged characteristics in the sequence input. Then, for the seasonal component, we have
\begin{align}\label{season_temp}
    \bS_{i,t - 1} = \bX_{i, t-1} - \bT_{i,t-1} \in \mathbb{R}^{S \times P}.
\end{align}
In \eqref{trend_comp} and \eqref{season_temp}, $\bT_{i,t-1}$ aims to capture the slow-moving and persistent temporal patterns, while $\bS_{i,t-1}$, as the residual, retains short-term fluctuations and high-frequency variations.

After decomposing $\bX_{i,t-1}$ into interpretable trend and seasonal components, the DLinear model aggregates these components and projects them into a vector of risk scores:
\begin{align}\label{dlinear_output}
\begin{split}
    \bz_{i,t} &= \bW_{S} \bS_{i,t-1} + \bW_{T} \bT_{i,t-1} \in \mathbb{R}^P,
    \\
    \widetilde{\bz}_{i,t} & = \mathrm{ReLU} \left( \bW_{\mathrm{hid}} \bz_{i,t} + \bb_{\mathrm{hid}} \right) \in \mathbb{R}^D,
    \\
    \by_{i,t} 
    &= g(\bX_{i,t-1};\btheta) 
    = \bW_{\mathrm{out}} \widetilde{\bz}_{i,t}  + \bb_{\mathrm{out}},
\end{split}
\end{align}
where $\bW_T\in\mathbb{R}^{1\times S}$ and $\bW_S\in\mathbb{R}^{1\times S}$ are the weight vectors that aggregate the trend and seasonal components along the temporal dimension, respectively, the weight matrices $\bW_{\mathrm{hid}} \in \mathbb{R}^{D \times P}$ and $\bW_{\mathrm{out}} \in \mathbb{R}^{2 \times D}$, together with the bias vectors $\bb_{\mathrm{hid}} \in \mathbb{R}^D$ and $\bb_{\mathrm{out}} \in \mathbb{R}^2$, define a one-hidden-layer neural network from the aggregated components to the vector of risk scores. 

From \eqref{trend_comp}--\eqref{dlinear_output}, we can see that the DLinear model captures the dynamics of characteristics through trend–seasonal decomposition and  then applies a one-hidden-layer neural network to exploit these components in forecasting. 

\subsubsection{LSTM}  

In the LSTM model, the temporal dynamics of the features are captured through a hidden state vector $\bh^s_{i,t} \in \mathbb{R}^D$ in an autoregressive manner:
\begin{align}
\begin{split}\label{lstm}
    \bm{z}_{i,t}^s &= \mathrm{tanh} (\bW_{z} \underline{\bx}_{i,t-1}^s + \bU_{z} \bm{h}_{i,t}^{s-1} + \bb_z),\\
    \bm{i}_{i,t}^s &= \mathrm{sigmoid} (\bW_{i} \underline{\bx}_{i,t-1}^s + \bU_{i} \bm{h}_{i,t}^{s-1} + \bb_i),\\
    \bm{f}_{i,t}^s &= \mathrm{sigmoid} (\bW_{f} \underline{\bx}_{i,t-1}^s + \bU_{f} \bm{h}_{i,t}^{s-1} + \bb_f),\\
    \bm{c}_{i,t}^s &= \bm{f}_{i,t}^s \odot \bm{c}_{i,t}^{s-1} + \bm{i}_{i,t}^s \odot \bm{z}_{i,t}^s,\\
    \bm{o}_{i,t}^s &= \mathrm{sigmoid} (\bW_{o} \underline{\bx}_{i,t-1}^s + \bU_{o} \bm{h}_{i,t}^{s-1} + \bb_o),\\
    \bm{h}_{i,t}^s &= \bm{o}_{i,t}^s \odot \mathrm{tanh}(\bm{c}_{i,t}^s),
\end{split}
\end{align}
for $s = 1, \dots, S$, where $\underline{\bx}_{i,t-1}^s = \bx_{i,\,t-1-S+s} \in \mathbb{R}^P$ is the $s$-th row of $\bX_{i,t-1}$, $\bW_* \in \mathbb{R}^{D \times P}$, $\bU_* \in \mathbb{R}^{D \times D}$, and $\bb_* \in \mathbb{R}^{D}$ for $* \in \{z,i,f,o\}$ are the weight matrices and bias vectors, respectively. Here, the operator $\odot$ denotes element-wise multiplication, while $\tanh(\cdot)$ and $\mathrm{sigmoid}(\cdot)$ are element-wise activation functions. Following \cite{hochreiter1997long}, the initial states $\bh_{i,\,t-1}^{\,0}$ and $\bc_{i,\,t-1}^{\,0}$ are set to $D$-dimensional zero vectors. 

In \eqref{lstm}, the hidden state $\bh_{i,t}^{s}$ serves as the key representation, summarizing the accumulated information up to step $s$. The cell state $\bc_{i,t}^{s}$ models the internal long-term memory by carrying information across steps. Through the $\mathrm{sigmoid}$ function, the forget gate $\bm{f}_{i,t}^s \in (0,1)^D$, input gate $\bm{i}_{i,t}^s \in (0,1)^D$, and output gate $\bm{o}_{i,t}^s \in (0,1)^D$ regulate how much past information is discarded, how much new information is stored, and how much information is revealed to the hidden state $\bh_{i,t}^{s}$, respectively.  

After the recursive extraction in \eqref{lstm}, the final hidden state \(\bh_{i,t}^S\) is passed through a linear transformation to obtain the risk scores:
\begin{align}\label{lstm_fc}
    \by_{i,t} = g(\bX_{i,t-1}; \btheta) = \bW_L \bh_{i,t}^S + \bb_L, 
\end{align}
where $\bW_L \in \mathbb{R}^{2 \times D}$ and $\bb_L \in \mathbb{R}^2$ are weight matrix and bias vector, and $\btheta$ collects all parameters within the weight matrices and bias vectors in \eqref{lstm} and \eqref{lstm_fc}.

\subsubsection{GRU}

As another widely used recurrent neural network, the GRU model simplifies the LSTM model by decreasing the number of gates. Its autoregressive update for the hidden state is given in \eqref{gru}. Although the ReSGA model uses the full cross-sectional input $\mathcal{X}_{t-1}$, the GRU component itself operates independently on each asset’s temporal information $\bX_{i, t-1}$. This makes the GRU, when used alone, a purely temporal model.

Then, as in the LSTM model, the GRU model passes the final hidden state $\bh_{i,t}^S$ through a linear transformation to compute the risk scores:
\begin{align}\label{gru_fc}
    \by_{i,t} = g(\bX_{i,t-1}; \btheta) = \bW_G \bh_{i,t}^S + \bb_G,
\end{align}
where $\bW_G \in \mathbb{R}^{2 \times D}$ and $\bb_G \in \mathbb{R}^2$ are the output weight matrix and bias vector, respectively, and $\btheta$ includes all parameters within weight matrices and bias vectors in \eqref{gru} and \eqref{gru_fc}.

\subsubsection{Informer}\label{sec:informer}

While the vanilla Transformer model \citep{transformer} has achieved remarkable success in natural language processing through the attention mechanism, several variants have been developed for different tasks, such as the Vision Transformer \citep{vit} for computer vision and the Speech Transformer \citep{Dong2018speech} for audio recognition. In time-series forecasting, a milestone Transformer-type architecture is the Informer model \citep{zhou2021informer}, which follows the Transformer’s encoder–decoder framework but replaces the standard attention mechanism with a probabilistic sparse self-attention mechanism to handle long sequences more efficiently, making it particularly suitable for time-series modeling. 

Let $\bX_{i,t - 1}^\ast = [\bx_{i, t-S^\ast}, \bx_{i,t-2}, \dots, \bx_{i,t-1}] \in \mathbb{R}^{S^\ast \times P}$ be a truncated version of $\bX_{i,t-1}$ with $S^\ast$ being a constant smaller than $S$. The general architecture of Informer model is defined as:
\begin{align}
\begin{split}
    \by_{i, t} & = g(\bX_{i, t-1}; \btheta) =  \InfDec(\bX_{i,t - 1}^\ast, \bH_{i, t}; \bpsi) \in \mathbb{R}^2
    \\
    \,\,\, \text{with}\,\,\, \bH_{i, t} & = \InfEnc(\bX_{i, t-1}; \bphi) \in \mathbb{R}^{S \times D},
\end{split}
\end{align}
where $\InfEnc(\cdot; \bphi): \mathbb{R}^{S \times P} \to \mathbb{R}^{S \times D}$ represents the encoder parameterized by $\bphi$, and $\InfDec(\cdot, \cdot; \bpsi): \mathbb{R}^{S^\ast \times P} \times \mathbb{R}^{S \times D} \to \mathbb{R}^2$ denotes the decoder that depends on the encoder output $\bH_{i,t}$ and its parameters $\bpsi$. Here, the encoder extracts the past-aware feature $\bH_{i,t}$ from the full historical input $\bX_{i,t-1}$, summarizing the underlying dynamics from the global viewpoint; afterwards, the decoder then leverages both the learned feature $\bH_{i,t}$ and the shorter input $\bX_{i,t-1}^\ast$ to predict the target vector $\by_{i,t}$. 

Specifically, the encoder first applies an embedding layer to extract embeddings from $\bX_{i,t-1}$, which are then combined as the input for subsequent layers:
\begin{align}\label{encoder_embedding}
\begin{split}
    \bH_{i, t}^{(0)} & = \bE^{\mathrm{val}}_{i,t-1} + \bE^{\mathrm{pos}}_{i,t-1} + \bE^{\mathrm{temp}}_{i,t-1}
    \\
    & = \mathrm{VE}(\bX_{i,t-1}; \bzeta) + \mathrm{PE} (\bX_{i,t-1}) + \mathrm{TE}(\bX_{i,t-1})  \in \mathbb{R}^{S\times D},
\end{split}
\end{align}
where $\bE^{\mathrm{val}}_{i,t-1}$, $\bE^{\mathrm{pos}}_{i,t-1}$ and $\bE^{\mathrm{temp}}_{i,t-1}$ are three $S \times D$ matrices, referred to as the value, positional and temporal embeddings, respectively. Meanwhile, $\mathrm{VE}(\cdot; \bzeta): \mathbb{R}^{S \times P} \rightarrow \mathbb{R}^ {S \times D}$, indexed by parameters $\bzeta$, applies a one-dimensional temporal convolution to produce the value embedding, while $\mathrm{PE}(\cdot): \mathbb{R}^{S \times P} \rightarrow \mathbb{R}^ {S \times D}$ and $\mathrm{TE}(\cdot): \mathbb{R}^{S \times P} \rightarrow \mathbb{R}^ {S \times D}$ are two parameter-free mappings to provide the positional and temporal embeddings\footnote{For clarity, their precise definitions are deferred to Section \ref{sec:embedding}.}.

Given $\bH_{i, t}^{(0)}$, a multi-layer self-attention block is employed to further capture complex temporal patterns:
\begin{align}\label{encoder_attn}
    \bH_{i,t}^{(l)} = \mathrm{FFN}\big( \mathrm{SelfAttn}\big( \bH_{i,t}^{(l-1)}; \bomega^{(l)} \big) ; \bdelta^{(l)} \big) \in \mathbb{R}^{S \times D}, \,\,\, \text{for}\,\,\,  l = 1, \dots, L,
\end{align}
where $L$ is the number of self-attention layers, and $\bH_{i,t}^{(l)}$ denotes the output from the $l$-th layer. Here, two operators are involved: $\mathrm{SelfAttn}$ and $\mathrm{FFN}$. Particularly, $\mathrm{SelfAttn}(\cdot; \bomega): \mathbb{R}^{S \times D} \to \mathbb{R}^{S \times D}$ is the self-attention operator based on the probabilistic sparse technique\footnote{See \cref{sec:attn} for its details.}, indexed by parameters $\bomega^{(l)}$; meanwhile, $\mathrm{FFN}(\cdot;\bdelta): \mathbb{R}^{S \times D} \to \mathbb{R}^{S \times D}$ denotes a row-wise feed-forward network, that is, a one-hidden-layer fully connected network applied independently to each row of the input\footnote{See \cref{sec:fc} for its details.}, parameterized by $\bdelta$.

After the recursive extraction, the output of encoder is defined as 
\begin{align*}
    \bH_{i,t} \equiv \bH_{i,t}^{(L)} = \InfEnc(\bX_{i, t-1}; \bphi),
\end{align*}
where $\bphi$ encompass all parameters in \eqref{encoder_embedding}--\eqref{encoder_attn}.  The matrix $\bH_{i,t}$ is referred to as the past-aware feature in our paper, as it summarizes the general information contained in $\bX_{i,t-1}$.

So far, we have illustrated the architecture of encoder. Compared to the encoder, the decoder also first applies an embedding layer to map $\bX_{i,t-1}^\ast$ into an aggregated feature:
\begin{align}\label{decoder_embedding}
    \vF_{i, t}^{(0)} = \mathrm{VE}(\bX_{i,t-1}^\ast; \bzeta^\ast) + \mathrm{PE} (\bX_{i,t-1}^\ast) + \mathrm{TE}(\bX_{i,t-1}^\ast) \in \mathbb{R}^{S^\ast \times D},
\end{align}
where $\mathrm{VE}$, $\mathrm{PE}$, and $\mathrm{TE}$ operators are consistent with those in \eqref{encoder_embedding}, $\bzeta^\ast$ denotes the parameters of the decoder’s embedding layer, and $\vF_{i,t}^{(0)}$ is the resulting aggregated feature.

Next, a multi-layer cross-attention block is applied to compute the hidden states within the decoder:
\begin{align}\label{decoder_attn}
\begin{split}
    \vF_{i,t}^{(l)} &= 
    \mathrm{FFN} \big( \mathrm{CrossAttn} \big( \overline{\vF}_{i,t}^{(l)}, \bH_{i,t} ; \bgamma^{(l)} \big)  ; \bdelta^{(l + L)} \big) \in \mathbb{R}^{S^\ast \times D}
    \\
    \text{with}\,\,\,
    \overline{\vF}_{i,t}^{(l)} &= \mathrm{CausalAttn} \big( \vF_{i,t}^{(l-1)}; \bomega^{(l + L)} \big) \in \mathbb{R}^{S^\ast \times D}, \,\,\, l = 1, \dots, L,
\end{split}
\end{align}
where $\vF_{i,t}^{(l)}$ denotes the output from the $l$-th layer. Compared to the self-attention block in \eqref{encoder_attn} for the encoder, the cross-attention block has the same depth and applies the same $\mathrm{FFN}$ operator, but introduces two additional operators: $\mathrm{CausalAttn}$ and $\mathrm{CrossAttn}$.  

Specifically, $\mathrm{CausalAttn}(\cdot; \bomega): \mathbb{R}^{S^\ast \times D} \to \mathbb{R}^{S^\ast \times D}$ is the causal attention operator, which is identical to $\mathrm{SelfAttn}$ except that it imposes a no-forward-looking constraint, forcing the decoder to follow a strictly autoregressive structure\footnote{The autoregressive structure means that the feature for time $s$ relies only on information available up to time $s$, for any $s = 1, \dots, S$. In contrast, the encoder operates in a non-autoregressive manner, where the feature at time $s$ can also depend on information from future steps $s+1, \dots, S$.}. Meanwhile, $\mathrm{CrossAttn}(\cdot, \cdot; \bgamma): \mathbb{R}^{S^\ast \times D} \times \mathbb{R}^{S \times D} \to \mathbb{R}^{S^\ast \times D}$ denotes the cross-attention operator, which functions similarly to self-attention but also attends to the past-aware feature $\bH_{i,t}$ (i.e., the encoder’s output) rather than only its own input. This mechanism allows the decoder to selectively integrate information from the full historical context, aligning its input with the temporal dependencies captured by the encoder.

Lastly, to obtain the risk scores, the most up-to-date information extracted by the decoder is fed into a linear transformation:
\begin{align}\label{informer_fc}
    \by_{i,t} = \InfDec(\bX_{i,t - 1}^\ast, \bH_{i, t}; \bpsi) = \bW_{\mathrm{in}} \vf_{i,t} + \bb_{\mathrm{in}},
\end{align}
where $\vf_{i,t} \in \mathbb{R}^{D}$ denotes the last row of $\vF_{i,t}^{(L)}$, representing the latest information available for time $t$, $\bW_{in} \in \mathbb{R}^{2 \times D}$ and $\bb_{in} \in \mathbb{R}^2$ are the weight matrix and bias vector, respectively, and $\bpsi$ includes all parameters involved in  \eqref{decoder_embedding}--\eqref{decoder_attn}. Clearly, the Informer model can be regarded as a special case of the general temporal model, with $\btheta$ encompassing all parameters within $\bphi$ and $\bpsi$.

\subsubsection{EInformer and DInformer}

As the vanilla Informer follows the encoder–decoder framework, recent studies have highlighted the potential of encoder-only and decoder-only variants in time-series forecasting \citep{Das2024decoder,Feng2024encoder,yao2025towards}. Motivated by these works, we implement two variants, the encoder-only Informer (EInformer) and the decoder-only Informer (DInformer), for VaR and ES forecasting.  

Specifically, the EInformer model consists solely of the encoder but modifies the original input $\bX_{i,t-1}$ by appending an additional row of learnable parameters at the end:
\begin{align}
    \overline{\bX}_{i,t-1} = [\bX_{i,t-1}, \bm{\nu}] \in \mathbb{R}^{(S + 1) \times P},
\end{align}
where $\overline{\bX}_{i,t-1}$ is the augmented input, and $\bm{\nu} \in \mathbb{R}^P$ is a vector of learnable parameters. This additional vector $\bm{\nu}$, referred to as the predictive token, serves as a placeholder representing the characteristics at next (yet unseen) time step. It allows the model to generate a forecast without explicitly relying on a decoder. During training, the predictive token interacts with the historical information through the self-attention mechanism, enabling the encoder to learn how the future prediction depends on the past characteristic. In this sense, the predictive token acts as a query that asks, ``given the past, what comes next''?

Next, the augmented input $\overline{\bX}_{i,t-1}$ is passed through the encoder as in \eqref{encoder_embedding}--\eqref{encoder_attn}, and the last row of the encoder output, corresponding to the predictive token, is used to compute the risk scores:
\begin{align}\label{EInformer}
    \by_{i,t} = g(\bX_{i,t-1}; \btheta) = \bW_e \overline{\bh}_{i,t} + \bb_e \in \mathbb{R}^2,
\end{align}
where $\bW_e \in \mathbb{R}^{2 \times D}$ and $\bb_e \in \mathbb{R}^2$ are the output weight matrix and bias vector, respectively. Here, $\overline{\bh}_{i,t} \in \mathbb{R}^{D}$ denotes the last row of the encoder output
\begin{align*}
    \overline{\bH}_{i,t} = [\overline{\bh}_{i,t-S}, \dots, \overline{\bh}_{i,t}] = \InfEnc(\overline{\bX}_{i,t-1}; \bphi) \in \mathbb{R}^{(S + 1) \times D},
\end{align*}
and $\btheta$ includes all parameters in $\bm{\nu}$, $\bW_e$, $\bb_e$, and $\bphi$.

In contrast, the DInformer model relies solely on the decoder and is defined as
\begin{align}\label{DInformer}
    \by_{i,t} = g(\bX_{i,t-1}; \btheta) = \InfDec(\bX_{i,t-1}; \bpsi) \in \mathbb{R}^2.
\end{align}
Compared with the decoder in the vanilla Informer model, the DInformer uses $\bX_{i,t-1}$ rather than $\bX_{i,t-1}^\ast$ as input. Moreover, since the past-aware representation $\bH_{i,t}$ from the encoder is absent, the cross-attention operator in \eqref{decoder_attn} is omitted. 

\subsection{Spatial-Temporal Models}\label{sec:spatial-temporal}

Besides our proposed ReSGA, we consider another spatial-temporal model in (\ref{spatial_temporal_model}): SGA. The SGA model can be viewed as a simplified version of the ReSGA model. It has exactly the same encoder as that in the ReSGA model, but omits the retriever component and hence removes the retrieval layer in the decoder of ReSGA. All remaining components, especially the mechanism for learning and utilizing dynamic group structures, are retained. Therefore, comparing ReSGA with SGA isolates the role of the retrieval mechanism and highlights the additional value of incorporating long-term cross-asset information beyond only considering contemporaneous group effects.

\subsection{Econometric Models}\label{sec:econometric}

As econometric benchmarks, we consider two univariate models that are estimated separately for each asset using only its own historical returns. Let $r_{i,t}$ denote the excess return of asset $i$ at time $t$, and let $v_{i,t}$ and $e_{i,t}$ denote the corresponding conditional $\VaR$ and $\ES$ at level $\tau$.

\subsubsection{GAS}\label{sec:gas1f_model}

Following \cite{patton2019}, we adopt the one-factor GAS model by assuming that both $\VaR$ and $\ES$ are driven by a single latent factor $\kappa_{i,t}$:
\begin{align}
    v_{i,t} &= a_i \exp(\kappa_{i,t}), \qquad e_{i,t} = b_i \exp(\kappa_{i,t}), \qquad b_i < a_i < 0,
\end{align}
with dynamics
\begin{align}
    \kappa_{i,t} =
    \omega_i + \beta_i \kappa_{i,t-1} + 
    \frac{\gamma_i}{b_i \exp(\kappa_{i,t-1})}
    \left[
        \frac{1}{\tau}\mathbf{1}\!\left\{r_{i,t-1} \le a_i \exp(\kappa_{i,t-1})\right\} r_{i,t-1}
        -
        b_i \exp(\kappa_{i,t-1})
    \right].
\end{align}
where $(\omega_i, a_i, b_i, \beta_i, \gamma_i)$ is a parameter vector. We set $ \omega_i = 0 $ as in \cite{patton2019} for identification, and estimate the remaining parameters by minimizing the FZ loss. 

The one-factor GAS specification assumes that the time variation in $\VaR$ and $\ES$ is governed by a common latent factor $\kappa_{i,t}$, so that the two risk measures move proportionally over time. The recursion for $\kappa_{i,t}$ is score-driven: Under the one-factor parameterization, the updating term is obtained from the derivative of the FZ loss with respect to $\kappa_{i,t}$. The parameter $\beta_i$ controls the persistence of the latent factor, while the parameter $\gamma_i$ determines the sensitivity of tail risk forecasts to new return realizations. The parameters $a_i$ and $b_i$ scale the latent factor into the corresponding $\VaR$ and $\ES$ forecasts.

\subsubsection{GARCH}\label{sec:garch_model}
\cite{patton2019} also consider a GARCH($1, 1$) specification estimated by FZ loss minimization. This model is defined as
\begin{equation}
\begin{aligned}
    r_{i,t} &= \sigma_{i,t}\eta_{i,t}, \qquad \eta_{i,t} \sim \text{i.i.d. } F_{\eta,i}(0,1), \\
    \sigma_{i,t}^2 &= \omega_i + \beta_i \sigma_{i,t-1}^2 + \gamma_i r_{i,t-1}^2.
\end{aligned}
\end{equation}
Under this location-scale structure,
\begin{equation}
\begin{aligned}
    v_{i,t} &= a_i \sigma_{i,t}, \qquad a_i = F_{\eta,i}^{-1}(\tau), \\
    e_{i,t} &= b_i \sigma_{i,t}, \qquad b_i = E(\eta_{i,t} \mid \eta_{i,t} \le a_i),
\end{aligned}
\end{equation}
where $(\omega_i, a_i, b_i, \beta_i, \gamma_i)$ is a parameter vector, \(\sigma_{i,t}\) is the conditional volatility of asset \(i\) at time \(t\), and \(\eta_{i,t}\) is an i.i.d. innovation with distribution \(F_{\eta,i}(0,1)\). As in \cite{patton2019}, we set $ \omega_i = 1 $.

Under this GARCH specification, both $\VaR$ and $\ES$ are proportional to the conditional volatility $\sigma_{i,t}$. Therefore, their time variation is fully determined by the GARCH recursion for $\sigma_{i,t}$. The parameter $\beta_i$ captures the persistence of conditional volatility, whereas the parameter $\gamma_i$ measures the impact of lagged squared returns on current risk. The parameters $a_i$ and $b_i$ correspond to the $\tau$-quantile and the associated lower-tail conditional mean of the standardized innovation $\eta_{i,t}$, respectively. In \cite{patton2019}, this model is estimated by minimizing the FZ loss, so that the GARCH dynamics are fitted to the joint forecasting problem for $\VaR$ and $\ES$ rather than to the full conditional distribution of returns.

\section{Details of Operators in the Informer Model}

\setcounter{equation}{0}

\subsection{Embedding Layer}\label{sec:embedding}

As described in \eqref{encoder_embedding} and \eqref{decoder_embedding}, the inputs to the encoder and decoder, $\bX_{i,t-1}$ and $\bX_{i,t-1}^\ast$, are first processed by the embedding layer that consists of three embedding operators, $\mathrm{VE}$, $\mathrm{PE}$, and $\mathrm{TE}$, which generate the value, positional, and temporal embeddings, respectively. For notational simplicity, we assume a generic input matrix $\bX \in \mathbb{R}^{S \times P}$, which contains $P$ different characteristics observed over $S$ consecutive time steps. Based on this formulation, the details of each embedding operator are described below.

\subsubsection{Value Embedding Operator} The value embedding operator $\mathrm{VE}$ applies a simple one-dimensional convolution to the input matrix $\bX$, mapping it into an output matrix $\bE^{\mathrm{val}} \in \mathbb{R}^{S \times D}$. The convolution employs $D$ learnable filters (also referred to as kernels), where $D$ corresponds to the dimensionality of the hidden state in the Informer model. Each filter is a smaller matrix of size $W \times P$ that scans the input along the row axis. Specifically, each filter slides over the input, computes the dot product between its weights and the corresponding input sub-vectors, and aggregates the results to produce the convolution output. 

\begin{figure}[!h]
    \centering
    \includegraphics[width=0.9\textwidth]{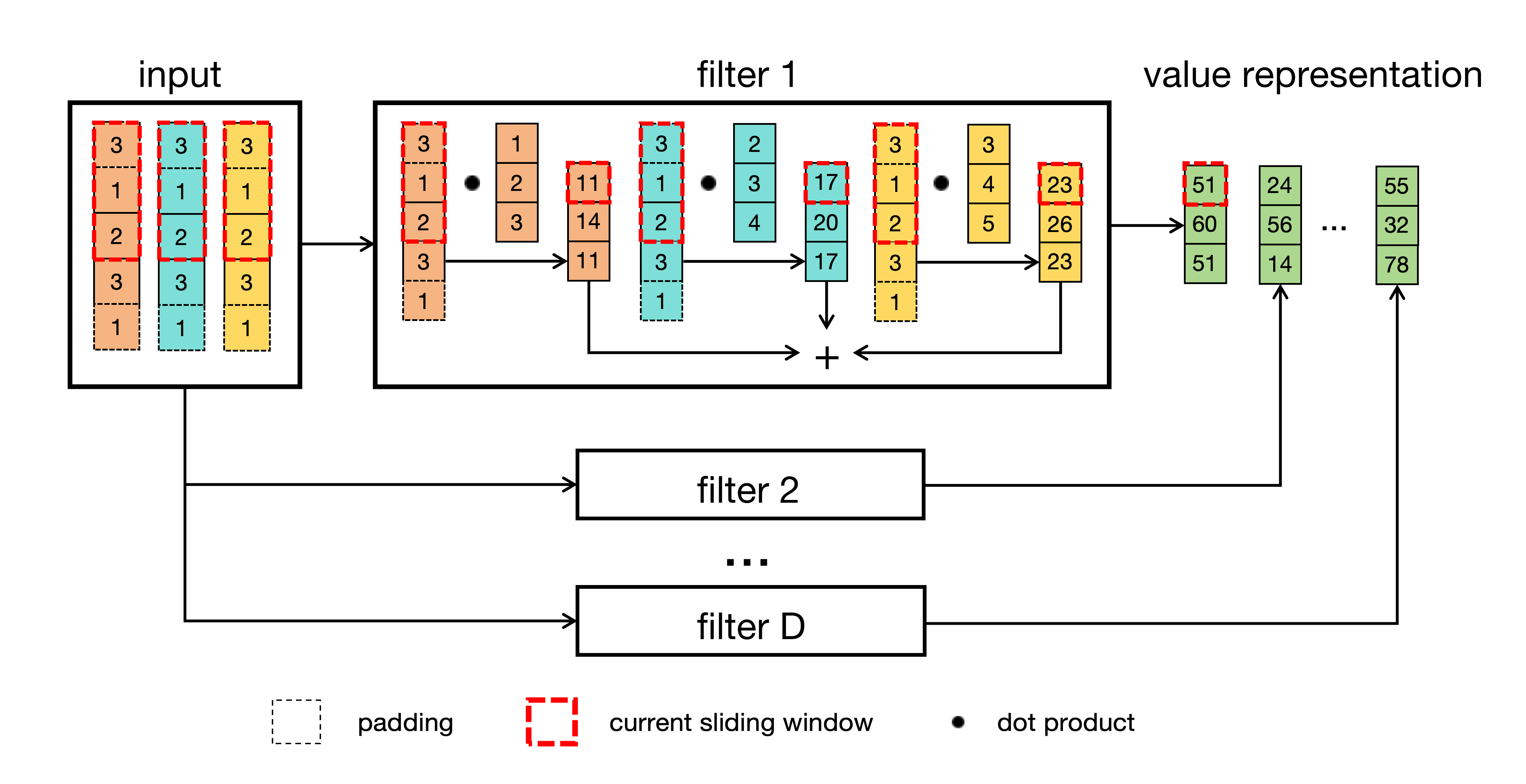}
    \caption{Illustration of the 1D convolution used for value embedding.}
    \label{fig:conv1d}
\end{figure}

The resulting matrix $\bE^{\mathrm{val}}$ effectively forms a weighted average of the input among nearby $W$ time points with data-driven weights and bias. The learnable parameters of these filters are collected in $\bzeta$. Figure \ref{fig:conv1d} provides a visual illustration of this one-dimensional convolution process.

\subsubsection{Positional Embedding Operator} 

The positional embedding operator $\mathrm{PE}$ is an essential component in Transformer-type models, as the attention mechanism itself treats the inputs as an unordered set of numbers rather than an ordered matrix. In other words, the attention mechanism does not inherently recognize that one element corresponds to a specific position or another position, unless positional information is explicitly injected\footnote{See Section~\ref{sec:attn} for a detailed discussion on this property of the attention mechanism.}.  

Following \cite{transformer}, a fixed sinusoidal mapping is adopted to construct the positional embedding representation, which encodes the positional information of each element of input $\bX$:
\[
\bE^{\mathrm{pos}} = \mathrm{PE}(\bX) \in \mathbb{R}^{S \times D},
\]
where the $(i,j)$-th element of $\bE^{\mathrm{pos}}$ is defined as
\begin{align*}
    e_{i,j}^{\mathrm{pos}} = 
    \begin{cases}
        \sin \!\left(\dfrac{i - 1}{10000^{\frac{j - 1}{2D}}} \right), & \text{if } j \text{ is odd}, \\[6pt]
        \cos \!\left(\dfrac{i - 1}{10000^{\frac{j - 2}{2D}}} \right), & \text{if } j \text{ is even}.
    \end{cases}
\end{align*}
By definition, the operator $\mathrm{PE}$ contains no learnable parameters; it depends solely on the shape of $\bX$ and provides deterministic positional representation that helps the attention mechanism preserve the sequential structure of the input.

\subsubsection{Temporal Embedding Operator} 

Motivated by the positional embedding operator, \cite{zhou2021informer} introduce the temporal embedding operator $\mathrm{TE}$, which transforms absolute time stamps into representations that capture potential seasonality and tendency. To construct the temporal embedding, we associate each row of $\bX$ with a time stamp consisting of its month, day, and corresponding weekday. Each of these calendar components can be represented by an integer: For example, months are encoded from $0$ (January) to $11$ (December), days from $0$ (1st) to $30$ (31st), and weekdays from $0$ (Monday) to $6$ (Sunday). Consequently, each time stamp is represented as a three-dimensional integer vector encoding its month, day, and weekday information.  

Let $\bM \in \mathbb{R}^{S \times 3}$ denote the matrix containing these integer-encoded time stamps for all $S$ observations. Given $\bM$, the temporal embedding representation is defined as
\begin{align*}
    \bE^{\mathrm{temp}} = \mathrm{TE}(\bX) \in \mathbb{R}^{S \times D},
\end{align*}
where the $(i,j)$-th element of $\bE^{\mathrm{temp}}$ is given by
\begin{align*}
    e_{i,j}^{\mathrm{temp}} = 
    \begin{cases}
        \displaystyle \sum_{k = 1}^3 \sin\!\left(10m_{i,k} \exp\!\left(\frac{4(j - 1)}{D}\right) \right), & \text{if } j \text{ is odd}, \\[6pt]
        \displaystyle \sum_{k = 1}^3 \cos\!\left(10m_{i,k} \exp\!\left(\frac{4(j - 2)}{D} \right) \right), & \text{if } j \text{ is even},
    \end{cases}
\end{align*}
where $m_{i,k}$ denotes the $(i,k)$-th element of $\bM$. 

Similar to $\mathrm{PE}$, $\mathrm{TE}$ contains no learnable parameters. It deterministically maps absolute time stamps into continuous temporal representations, allowing the attention mechanism to better capture seasonal patterns and potential trends inherent in the data.

\subsection{Probabilistic Sparse Attention}\label{sec:attn}

As discussed in Section \ref{sec:informer}, the Informer model employs three types of attention operators: $\mathrm{SelfAttn}$, $\mathrm{CausalAttn}$, and $\mathrm{CrossAttn}$. To efficiently handle long time input, these operators are further equipped with a probabilistic sparse (ProbSparse) technique, resulting in the so-called probabilistic sparse attention \citep{zhou2021informer}. The details of each attention operator are introduced below, while  we assume a generic matrix input $\bX \in \mathbb{R}^{S \times D}$ for these attention operators for notational simplicity.

\subsubsection{Self-Attention Operator}

First, the input $\bX$ is linearly projected into three matrices: the query $\bQ$, the key $\bK$, and the value $\bV$, defined as
\begin{align}\label{QKV}
    \bQ = \bX \bW_Q \in \mathbb{R}^{S \times D}, \quad
    \bK = \bX \bW_K \in \mathbb{R}^{S \times D}, \quad
    \bV = \bX \bW_V \in \mathbb{R}^{S \times D},
\end{align}
where $\bW_Q, \bW_K, \bW_V \in \mathbb{R}^{D \times D}$ are learnable weight matrices. In the multi-head attention setting with $M$ heads, these matrices are further projected into head-specific low-dimensional subspaces:
\begin{align}\label{head_QKV}
    \bQ^{(m)} = \bQ \bW_Q^{(m)} \in \mathbb{R}^{S \times \underline{D}}, \quad
\bK^{(m)} = \bK \bW_K^{(m)} \in \mathbb{R}^{S \times \underline{D}}, \quad
\bV^{(m)} = \bV \bW_V^{(m)} \in \mathbb{R}^{S \times \underline{D}}, 
\end{align}
for $m = 1, \dots, M$, where $\bW_Q^{(m)}, \bW_K^{(m)}, \bW_V^{(m)} \in \mathbb{R}^{D \times \underline{D}}$ are the head-specific weight matrices, and $\underline{D} = D / M$ denotes the dimensionality of each head-specific subspace.

Next, the ProbSparse technique is then applied independently within each head. Different from the standard Transformer attention where each query attends to all keys, the ProbSparse technique reduces complexity by sampling a subset of keys for each query and computing a sparsity score to identify the most informative queries. 

Specifically, for $i = 1, \dots, S$, let $\bq_i^{(m)}$, $\bk_i^{(m)}$, and $\bv_i^{(m)}$ denote the $i$-th rows of $\bQ^{(m)}$, $\bK^{(m)}$, and $\bV^{(m)}$, respectively. Moreover, let $\mathcal{S}_i$ be a uniformly random sampled subset of $\{1, \dots, S\}$ with $|\mathcal{S}_i| = \lfloor c \log S \rfloor$, where $c$ is a hyperparameter that balances computational efficiency and approximation accuracy. The sparsity score is then defined as
\begin{align}\label{attn_sparse_score}
    \gamma_i^{(m)} 
    = \max_{j \in \mathcal{S}_i} \beta^{(m)}_{i,j} - \frac{1}{|\mathcal{S}_i|} \sum_{j \in \mathcal{S}_i} \beta^{(m)}_{i,j}, 
\end{align}
where $\beta^{(m)}_{i,j}$, referred to as the attention coefficient, measures the similarity between the query vector $\bq_i^{(m)}$ and the key vector $\bk_j^{(m)}$ through their scaled inner product:
\begin{align*}
    \beta^{(m)}_{i,j} = \frac{\bq_i^{(m)\prime} \bk_j^{(m)}}{\sqrt{\underline{D}}}.
\end{align*}

Intuitively, the sparsity score $\gamma_i^{(m)}$ quantifies how much the maximum of similarity between the query vector $\bq_i^{(m)}$ and its sampled key vectors $\{\bk_j^{(m)} : j \in \mathcal{S}_i\}$ deviates from its mean. A larger sparsity score indicates that the query vector has a few highly similar key vectors, implying that these key–value pairs are more informative and thus require full attention computation. Conversely, a smaller sparsity score suggests that the query and key vectors exhibit more uniform similarity, allowing the corresponding attention computation to be pruned to reduce complexity.

To implement this idea, let $\mathcal{I}^{(m)}$ denote the set of indices corresponding to the largest $\lfloor c \log S \rfloor$ values among $\{\gamma_i^{(m)} : i = 1, \dots, S\}$. The output of the $m$-th attention head is then defined as $\bC^{(m)} \in \mathbb{R}^{S \times \underline{D}}$ with its $i$-th row satisfying 
\begin{align}\label{attn_head}
    \bc_{i}^{(m)} = 
    \begin{cases}
        \displaystyle \sum_{j=1}^S \alpha_{i,j}^{(m)} \bv_j^{(m)}, & \text{if } i \in \mathcal{I}^{(m)}, \\[8pt]
        \displaystyle \frac{1}{S} \sum_{j=1}^S \bv_j^{(m)}, & \text{otherwise},
    \end{cases}
\end{align}
where $\alpha_{i,j}^{(m)}$ denotes the normalized attention weight, defined as
\begin{align*}
    \alpha_{i,j}^{(m)} 
    = \mathrm{SoftMax}\left( \beta_{i,j}^{(m)} \right)
    = \frac{\exp(\beta_{i,j}^{(m)})}{\sum_{k = 1}^S \exp\left( \beta_{i,k}^{(m)} \right)}.
\end{align*}

Notably, by employing the sparsity score defined in \eqref{attn_sparse_score}, only $\lfloor c \log S \rfloor$ head-specific outputs $\bc_{i}^{(m)}$ require full attention computation involving $\{\alpha_{i,j}^{(m)} : j = 1, \dots, S\}$, while the remaining ones are approximated by the average of all value vectors. Consequently, the computational complexity is reduced from $\mathcal{O}(S^2)$ in the standard self-attention mechanism to $\mathcal{O}(S \log S)$ in the ProbSparse self-attention, providing a significant efficiency advantage for handling long input sequences (i.e., large value of $S$) in the Informer model.

Finally, the outputs from all attention heads are concatenated and linearly projected to obtain the final self-attention output $\bY$:
\begin{align}\label{attn_fc}
    \bY = \mathrm{SelfAttn}(\bX; \bomega) = \bC \bW_O \in \mathbb{R}^{S \times D}, 
    \,\,\, \text{with} \,\,\,
    \bC = \left[ \bC^{(1)}, \bC^{(2)}, \dots, \bC^{(M)} \right] \in \mathbb{R}^{S \times D},
\end{align}
where $\bW_O \in \mathbb{R}^{D \times D}$ is a learnable output projection matrix. The parameter set $\bomega$ includes all learnable weights involved in \eqref{QKV}, \eqref{head_QKV}, and \eqref{attn_fc}.

\subsubsection{Causal Attention Operator}

The causal-attention operator $\mathrm{CausalAttn}$ extends the self-attention operator by imposing a no-forward-looking constraint. This restriction enforces a strictly autoregressive structure within the attention computation. Specifically, following the same procedure in \eqref{QKV}--\eqref{attn_sparse_score}, we obtain $\big\{ \big( \bq_i^{(m)}, \bk_i^{(m)}, \bv_i^{(m)} \big) \big\}$ and compute the sparsity score $\overline{\gamma}_i^{(m)}$ in $\mathrm{CausalAttn}$ as
\begin{align}\label{causal_sparse_score}
    \overline{\gamma}_i^{(m)} 
    = \max_{j \in \mathcal{S}_i} \beta^{(m)}_{i,j}
    - \frac{1}{|\mathcal{S}_i|} \sum_{j \in \mathcal{S}_i} \beta^{(m)}_{i,j},
\end{align}
where the attention coefficient $\beta_{i,j}$ is defined as before, but the sampling set $\mathcal{S}_i$ is drawn from $\{1, \dots, i\}$ instead of $\{1, \dots, S\}$, with size $\lfloor c \log S \rfloor$.

Similarly, let $\overline{\mathcal{I}}^{(m)}$ denote the indices of the largest $\lfloor c \log S \rfloor$ values among $\{\overline{\gamma}_i^{(m)} : i = 1, \dots, S\}$. The head-specific output $\overline{\bC}^{(m)}$ is then obtained row-wisely as
\begin{align}\label{causal_head}
    \overline{\bc}_{i}^{(m)} =
    \begin{cases}
        \displaystyle \sum_{j=1}^i \overline{\alpha}_{i,j}^{(m)} \bv_j^{(m)}, & \text{if } i \in \overline{\mathcal{I}}^{(m)}, \\[8pt]
        \displaystyle \frac{1}{i} \sum_{j=1}^i \bv_j^{(m)}, & \text{otherwise},
    \end{cases}
\end{align}
where the normalized attention weight $\overline{\alpha}_{i,j}^{(m)}$ is given by
\begin{align*}
    \overline{\alpha}_{i,j}^{(m)} 
    = \mathrm{SoftMax}\left(\beta_{i,j}^{(m)} \right)
    = \frac{\exp(\beta_{i,j}^{(m)})}
           {\sum_{k = 1}^i \exp(\beta_{i,k}^{(m)})}.
\end{align*}

The outputs from all attention heads are concatenated and linearly projected to produce the causal-attention output $\bY$:
\begin{align}\label{causal_fc}
\begin{split}
    \bY &= \mathrm{CausalAttn}(\bX; \bomega) 
    = \overline{\bC} \bW_O \in \mathbb{R}^{S \times D}, \\[3pt]
    \text{with} \quad 
    \overline{\bC} &= \left[ \overline{\bC}^{(1)}, \overline{\bC}^{(2)}, \dots, \overline{\bC}^{(M)} \right]
    \in \mathbb{R}^{S \times D},
\end{split}
\end{align}
where $\bW_O$ is defined in \eqref{attn_fc}, and $\bomega$ includes all learnable weights involved in \eqref{QKV}, \eqref{head_QKV}, and \eqref{causal_fc}. The causal constraint guarantees that the model respects temporal order, making it suitable for autoregressive sequence modeling.

\subsubsection{Cross-Attention Operator}

The cross-attention operator $\mathrm{CrossAttn}$ is similar to the self-attention operator, but it requires two generic inputs, $\bX \in \mathbb{R}^{S \times D}$ and $\bZ \in \mathbb{R}^{S^\ast \times D}$, where $S^\ast$ is another constant and can be different from $S$. The key distinction lies in that the key and value matrices, $\bK$ and $\bV$, are computed from the external input $\bZ$ rather than from $\bX$. This design enables the model to align and integrate information between two different representations (e.g., between the encoder’s output $\bH_{i,t}$ and the decoder’s hidden state $\vF_{i,t}$ in the Informer architecture).

Specifically, the query $\bQ$, key $\bK$, and value $\bV$ for $\mathrm{CrossAttn}$ are defined as
\begin{align*}
    \bQ = \bX \bW_Q \in \mathbb{R}^{S \times D}, \,\,\,
    \bK = \bZ \bW_K \in \mathbb{R}^{S^\ast \times D}, 
    \,\,\, \text{and} \,\,\, 
    \bV = \bZ \bW_V \in \mathbb{R}^{S^\ast \times D}.
\end{align*}
Given $\bQ$, $\bK$, and $\bV$, following the same computation steps as in \eqref{head_QKV}--\eqref{attn_fc}, the output of the cross-attention operator is obtained as
\begin{align*}
    \bY = \mathrm{CrossAttn}(\bX, \bZ) = \bC \bW_O \in \mathbb{R}^{S \times D}.
\end{align*}
Through this mechanism, the cross-attention operator allows the model to selectively attend to the most relevant parts of $\bZ$ while processing $\bX$, effectively establishing an information bridge between different representation spaces.

\section{Implementation Details}\label{sec:implementation}

\subsection{Additional Missing-Value Preprocessing}

As described in \cref{sec:data}, firm characteristics are rank-normalized cross-sectionally each month to reduce the influence of outliers, and missing characteristic values are imputed using contemporaneous cross-sectional medians. For the models under the unified learning framework, we drop stock-month observations when their related $r_{i,t}$ are missing. For the econometric models, GAS and GARCH, which are estimated separately for each asset using return histories, we remove missing $r_{i,t}$ observations and concatenate the remaining non-missing observations into a valid return series.

\subsection{Regularization Learning}

As discussed in \cref{sec:competitor}, we apply a range of deep learning models to VaR and ES forecasting. While the deep learning models offer strong capacity and flexibility in extracting nonlinear relationship from data, they are also prone to overfitting. In addition, their complex architectures and high-dimensional parameter spaces may lead to unstable estimates due to sensitivity to initialization and convergence to local optima. To address the above issues, we adopt two regularization techniques for these models.

First, a standard approach to mitigate overfitting is to partition the data into three disjoint parts: training, validation, and testing samples, while preserving the temporal ordering. The training sample is used for parameter estimation, the validation sample for hyperparameter tuning and early stopping, and the testing sample exclusively for evaluating out-of-sample performance. In this study, we apply this split within each expanding window in chronological order. Details of the window construction are provided in \cref{sec:data}.

Second, following \cite{gu2020empirical}, we adopt an ensemble learning strategy for regularization. Model parameters $\btheta$ are estimated by solving the optimization problem in \eqref{eq:train_objective} using the Adam algorithm \citep{Kingma2015AdamAM}. Due to the nonlinearity of the models and the sensitivity of Adam to initial values, the resulting parameter estimates may vary across different initializations. To mitigate the impact of initialization, we repeat the optimization from multiple random initializations. For each run, we estimate the parameters and obtain forecasts $\widehat{\VaR}_{i,t}$ and $\widehat{\ES}_{i,t}$. The final forecasts are then computed as the average across these runs (five in our implementation). This ensemble procedure reduces stochasticity caused by random initialization and improves the stability of the predictions.

\subsection{Hyperparameters}

As with any deep learning model, performance depends on the choice of hyperparameters, for which optimal selection is generally infeasible in theory. We therefore adopt standard settings commonly used in the literature for the learning rate, batch size, and other non-structural hyperparameters. Our main focus is on structural hyperparameters that govern model capacity, primarily the hidden-state dimension $D$, the depth $L$ (number of layers), and, for the Informer family, the number of attention heads $M$. To reduce computational costs, hyperparameters are tuned only in the first rolling window, and based on performance on the validation sample, ensuring that no look-ahead bias is introduced. Then, they are held fixed across subsequent windows, reflecting a realistic forecasting setting in which frequent re-tuning is impractical and time-consuming. 

Moreover, these hyperparameters are also varied to study the virtue of model complexity discussed in \cref{sec:scaling}. See \cref{tab:hyperparam_settings} for a summary of the corresponding hyperparameter settings for each parameter-scale tier.

\begin{table}[!h]
\centering
\caption{Hyperparameter settings for different parameter sizes.}
\label{tab:hyperparam_settings}
\begin{threeparttable}
\setlength{\tabcolsep}{6.5pt}
\begin{tabular}{lccccc}
\toprule
Model & $\approx 10^{3}$ & $\approx 10^{4}$ & $\approx 10^{5}$ & $\approx 10^{6}$ & $\approx 10^{7}$ \\
\midrule
NN        
& $L=3,\,D=16$ 
& $L=3,\,D=64$  
& $L=3,\,D=512$ 
& $L=3,\,D=1024$ 
& $L=3,\,D=5120$  \\

LANN   
& --
& $L=3,\,D=8$  
& $L=3,\,D=64$ 
& $L=3,\,D=512$ 
& $L=3,\,D=2560$  \\

DLinear   
& $D=16$ 
& $D=64$  
& $D=1024$ 
& $D=5120$ 
& $D=64000$  \\

LSTM      
& $L=1, D=4$ 
& $L=1, D=32$ 
& $L=1, D=128$  
& $L=3, D=256$ 
& $L=3, D=800$ \\

GRU       
& $L=1, D=4$ 
& $L=1, D=32$ 
& $L=1, D=128$  
& $L=3, D=256$ 
& $L=3, D=800$  \\

\multirow{2}{*}{Informer}
& $L=1,$ 
& $L=1,$ 
& $L=1,$  
& $L=1,$ 
& $L=3,$  \\
& $D=4, M=2$ 
& $D=8, M=2$ 
& $D=64, M=4$  
& $D=256, M=8$ 
& $D=384, M=8$  \\

\multirow{2}{*}{EInformer}
& $L=1,$ 
& $L=1,$ 
& $L=1,$  
& $L=1,$ 
& $L=3,$  \\
& $D=8, M=2$ 
& $D=16, M=2$ 
& $D=128, M=4$  
& $D=512, M=8$ 
& $D=768, M=8$  \\

\multirow{2}{*}{DInformer}
& $L=1,$ 
& $L=1,$ 
& $L=1,$  
& $L=1,$ 
& $L=3,$  \\
& $D=8, M=2$ 
& $D=16, M=2$ 
& $D=128, M=4$  
& $D=512, M=8$ 
& $D=768, M=8$  \\

SGA       
& $L=1, D=8$ 
& $L=1, D=16$ 
& $L=1, D=128$  
& $L=1, D=512$ 
& $L=2, D=1024$  \\

ReSGA     
& $L=1, D=8$ 
& $L=1, D=16$ 
& $L=1, D=128$  
& $L=1, D=512$ 
& $L=2, D=1024$  \\
\bottomrule
\end{tabular}

\begin{tablenotes}[flushleft]
\footnotesize
\item \textit{Notes.} Each entity reports the hyperparameter configuration used to attain the corresponding parameter-scale tier.
For the Informer family, $L$ denotes the shared number of encoder and decoder layers, and $M$ is the number of attention heads.
For SGA and ReSGA, $L$ refers to the number of GRU layers in the encoder block.
\end{tablenotes}

\end{threeparttable}
\end{table}

\section{Statistical Testing}\label{sec:testing}

In this appendix, we introduce the details of four tests used in \cref{sec:empirical}.

\subsection{DM Test}

The DM test evaluates whether two competing models exhibit statistically different predictive accuracy under a chosen loss function. Suppose we compare two forecasting models, denoted \(a\) and \(b\), for their ability to predict \(\VaR_{i,t}\) and \(\ES_{i,t}\). Let $\left (\widehat{\VaR}^{(a)}_{i,t}, \widehat{\ES}^{(a)}_{i,t} \right) $ and $\left(\widehat{\VaR}^{(b)}_{i,t}, \widehat{\ES}^{(b)}_{i,t} \right)$ be the forecasts of $\left (\VaR_{i,t}, \ES_{i,t} \right)$ given by models $a$ and $b$, respectively. The loss differential between models $a$ and $b$ at time $t$ is defined as
\begin{align*}
    & d_{t}(a, b) = \frac{1}{N_t} \sum_{i = 1}^{N_t} d_{i,t}(a, b),
    \\
    \,\,\, \text{with} \,\,\, & d_{i,t}(a, b)= \ell_{\mathrm{FZ0}} \left(r_{i,t}, \left( \widehat{\VaR}^{(a)}_{i,t}, \widehat{\ES}^{(a)}_{i,t} \right) \right) - \ell_{\mathrm{FZ0}} \left(r_{i,t}, \left( \widehat{\VaR}^{(b)}_{i,t}, \widehat{\ES}^{(b)}_{i,t} \right) \right),
\end{align*}
where $\ell_{\mathrm{FZ0}}(\cdot)$ is the FZ loss function defined in \eqref{eq:FZ0}. 

Since the loss differential measures the difference of predictive powers of models $a$ and $b$, the DM test can examine the following null hypothesis $\mathcal{H}_{0}$ to check whether two competing models exhibit statistically different predictive accuracy:
\begin{align}
    \mathcal{H}_{0}: \mathbb{E}\left[d_t(a, b) \right] = 0.
\end{align}
To detect $\mathcal{H}_{0}$, the DM test statistic is defined as
\begin{align*}
    \mathrm{DM} = \frac{\bar{d}(a,b)}{\widehat{\sigma}(a,b)},
\end{align*}
where \(\bar{d}(a,b)\) and \(\widehat{\sigma}(a,b)\) are the mean and Newey–West standard error \citep{newey1987} of $d_t(a, b)$ over the testing sample. 

Notably, a significantly positive (or negative) DM statistic indicates that model \(a\) performs worse (or better) than model \(b\) under the FZ loss. Under some mild conditions, the critical value for the DM test statistic is the $(1 - \alpha/2)$-th quantile of standard normal distribution, where $\alpha$ is the confidence level. 

\subsection{MCS Test}

The DM test compares models pairwise, whereas the Model Confidence Set (MCS) test of \citet{Hansen2011model} evaluates a collection of competing models jointly. Starting from the full model set, the MCS procedure sequentially removes models that are significantly dominated under the chosen loss function and returns a set of models that cannot be statistically distinguished from the best-performing model(s).

Let $\mathbb{M}^0$ denote the initial set of competing models. We use the FZ loss differential $d_{i,t}(a,b)$ defined in the DM test section to compare any two models $a,b\in\mathbb{M}^0$. We aggregate over the evaluation panel and define
\begin{align*}
\bar d(a,b)
=
\frac{1}{|\mathcal{P}|}
\sum_{(i,t)\in\mathcal{P}}
d_{i,t}(a,b),
\end{align*}
where $\mathcal{P}$ is the set of stock-month observations in the test sample. Let $D(a,b)=\mathbb{E}[d_{i,t}(a,b)]$. Since lower loss indicates better predictive performance, the population superior set is
\begin{align*}
\mathbb{M}^{\ast}
=
\left\{
a\in\mathbb{M}^0:
D(a,b)\le 0
\ \text{for all }\
b\in\mathbb{M}^0
\right\}.
\end{align*}

For any candidate set $\mathbb{M}\subseteq\mathbb{M}^0$, the MCS test considers the null hypothesis
\[
\mathcal{H}_{0,\mathbb{M}}:
D(a,b)=0,
\quad
\text{for all } a,b\in\mathbb{M}.
\]
We use the range statistic
\begin{align*}
T_{\mathbb{M}}
=
\max_{a,b\in\mathbb{M}}
\left|t(a,b)\right|,
\qquad
t(a,b)
=
\frac{\bar d(a,b)}{\widehat{\sigma}(a,b)},
\end{align*}
where $\widehat{\sigma}(a,b)$ denotes the bootstrap standard error of $\bar d(a,b)$.

The critical value is obtained from a panel block bootstrap as in \cite{yang2024asset}. In each bootstrap replication, we resample stocks with replacement and resample months using contiguous time blocks, so that both cross-sectional heterogeneity and serial dependence in the loss differentials are accounted for. Denote by $\bar d^{(s)}(a,b)$ the average loss differential computed from the $s$-th bootstrap sample, $s=1,\ldots,B$. We estimate the bootstrap standard error by
\[
\widehat{\sigma}(a,b)
=
\left[
\frac{1}{B}
\sum_{s=1}^{B}
\left\{
\bar d^{(s)}(a,b)-\bar d(a,b)
\right\}^{2}
\right]^{1/2}.
\]
The bootstrap distribution of
\[
T_{\mathbb{M}}^{(s)}
=
\max_{a,b\in\mathbb{M}}
\left|
\frac{\bar d^{(s)}(a,b)-\bar d(a,b)}
{\widehat{\sigma}(a,b)}
\right|
\]
is used to obtain the critical value $CV_{\alpha,\mathbb{M}}$.

If $T_{\mathbb{M}}\le CV_{\alpha,\mathbb{M}}$, the null is not rejected and the current set $\mathbb{M}$ is retained. Otherwise, the model with the largest standardized average loss relative to the remaining models is removed. Specifically, the eliminated model is selected according to
\[
e_{\mathbb{M}}
=
\argmax_{a\in\mathbb{M}}
\frac{\bar d_{a,\cdot}}{\widehat{\sigma}_{a,\cdot}},
\qquad
\bar d_{a,\cdot}
=
\frac{1}{|\mathbb{M}|-1}
\sum_{\substack{b\in\mathbb{M}\\ b\ne a}}
\bar d(a,b),
\]
where $\widehat{\sigma}_{a,\cdot}$ is the bootstrap standard error of
$\bar d_{a,\cdot}$.
The procedure is repeated on $\mathbb{M}\setminus\{e_{\mathbb{M}}\}$ until the null of equal predictive ability is no longer rejected. The final set is reported as the model confidence set.

\subsection{CC Test}

The CC test aims to  evaluate the validity of VaR forecasts. Let $I_{i,t} = \mathbf{1}\big\{ r_{i,t} \leq \widehat{\VaR}_{i,t} \big\}$, where $\mathbf{1}(\cdot)$ denotes the indicator function. Under correct specification, the violation indicators $\{I_{i,t}\}$ should form an i.i.d. Bernoulli sequence with success probability $\tau$. Based on this fact, we examine the validity of VaR forecasts by checking the null hypothesis
\begin{align*}
    \mathcal{H}^{(v)}_{0, i}: \{I_{i,t}\} \text{ is an i.i.d. Bernoulli sequence.}
\end{align*}

Following \citet{Christoffersen1998EvaluatingIF}, let $T_{jk}$ denote the number of observations for which $I_{i,t-1}=j$ and $I_{i,t}=k$, with $j,k\in\{0,1\}$. Define
\[
T_1 = T_{01} + T_{11}, \quad 
T_0 = T_{00} + T_{10}, \quad
\hat{\pi}_0 = \frac{T_{01}}{T_{00}+T_{01}}, 
\quad
\hat{\pi}_1 = \frac{T_{11}}{T_{10}+T_{11}}, 
\quad
\hat{\pi} = \frac{T_{01}+T_{11}}{T^\ast},
\]
where $T^\ast$ is the total number of observations in the evaluation period. Then, the CC test jointly assesses unconditional coverage and independence by combining the corresponding likelihood ratio statistics:
\begin{align*}
    \mathrm{LR}_{\mathrm{cc}} = \mathrm{LR}_{\mathrm{uc}} + \mathrm{LR}_{\mathrm{ind}},
\end{align*}
where 
\[
\mathrm{LR}_{\mathrm{ind}}
= -2 \log \left(
\frac{(1-\hat{\pi})^{T_{00}+T_{10}} \hat{\pi}^{T_{01}+T_{11}}}
{(1-\hat{\pi}_0)^{T_{00}} \hat{\pi}_0^{T_{01}}
 (1-\hat{\pi}_1)^{T_{10}} \hat{\pi}_1^{T_{11}}}
\right)
\]
and 
\[
\mathrm{LR}_{\mathrm{uc}} = -2 \log \left(
\frac{(1-\tau)^{T_0}\tau^{T_1}}
{(1-\hat{\pi})^{T_0}\hat{\pi}^{T_1}}
\right),
\]
Under the null hypothesis $\mathcal{H}^{(v)}_{0, i}$, $\mathrm{LR}_{\mathrm{cc}}$ is asymptotically distributed as $\chi^2(2)$. Therefore, we reject $\mathcal{H}^{(v)}_{0, i}$ when the value of $\mathrm{LR}_{\mathrm{cc}}$ is larger than the $\alpha$-th upper quantile of $\chi^2(2)$.

\subsection{AESR test}

Following \cite{bayer2022}, we assess the validity of ES forecasts via the AESR test, which is based on two regressions, linking observed returns to the corresponding VaR and ES forecasts. Specifically, these regressions are defined as
\begin{align*}
r_{i,t} = \beta_{i,1} + \beta_{i,2}\widehat{\VaR}_{i,t} + u^{(q)}_{i,t},
\qquad
r_{i,t} = \gamma_{i,1} + \gamma_{i,2}\widehat{\ES}_{i,t} + u^{(e)}_{i,t},
\end{align*}
where $\beta_{i,1}, \beta_{i,2}, \gamma_{i,1}$, and $\gamma_{i,2}$ are unknown coefficients, and $u^{(q)}_{i,t}$ and $u^{(e)}_{i,t}$ are model errors. The AESR test makes use of the fact that, conditional on the information set $\mathcal{F}_{t-1}$, the $\tau$-th conditional VaR of $u^{(q)}_{i,t}$ and the $\tau$-th conditional ES of $u^{(e)}_{i,t}$ are equal to zero almost surely.

Consequently, for each asset $i$, the AESR test assesses the adequacy of $\widehat{\ES}_{i,t}$ by testing the null hypotheses: 
\begin{align*}
\mathcal{H}^{(e)}_{i,0}: (\gamma_{i,1}, \gamma_{i,2})' = (0, 1)'.
\end{align*}
Intuitively, under $\mathcal{H}^{(e)}_{i,0}$, the ES forecasts are correctly specified in the sense that $\widehat{\ES}_{i,t}$ behaves close to the conditional ES of $r_{i,t}$ given the information set. Hence, for each model, rejection of $\mathcal{H}^{(e)}_{i,0}$ at the significance level $\alpha$ indicates that the corresponding ES forecasts fail to capture the tail distribution of returns, suggesting deficiencies in the model’s calibration.

Notably, because ES is not elicitable on its own, $\gamma_{i,1}$ and $\gamma_{i,2}$ cannot be estimated separately. Instead, the coefficients $(\beta_{i,1}, \beta_{i,2}, \gamma_{i,1}, \gamma_{i,2})$ are estimated jointly by minimizing the FZ loss. Let $\widehat{\boldsymbol{\gamma}}_i = (\widehat{\gamma}_{i,1}, \widehat{\gamma}_{i,2})'$ denote the resulting estimates, and let $\widehat{\Sigma}_{\gamma,i}$ be a consistent estimator of the asymptotic covariance matrix of $\widehat{\boldsymbol{\gamma}}_i$. Then, we can define a Wald-type test statistic:
\begin{align*}
W^{(e)}_i
= \big(\widehat{\boldsymbol{\gamma}}_i - (0,1)'\big)^{\!\top}
\widehat{\Sigma}_{\gamma,i}^{-1}
\big(\widehat{\boldsymbol{\gamma}}_i - (0,1)'\big),
\end{align*}
which asymptotically follows a $\chi^2(2)$ distribution under $\mathcal{H}^{(e)}_{i,0}$. The null hypothesis is rejected when $W^{(e)}_i$ exceeds the corresponding critical value. Like CC test, the AESR test is applied to each asset $i$ separately.

\end{document}